\documentclass[10pt,journal,compsoc]{IEEEtran}
\usepackage{enumitem}
\usepackage{graphicx}
\usepackage{graphics}
\usepackage{tcolorbox}
\usepackage{setspace}
\usepackage{mathtools}
\usepackage{transparent}
\usepackage{xcolor}
\usepackage{bm}
\usepackage[frozencache,cachedir=minted-cache]{minted}

\newcommand{\norm}[1]{\left\lVert#1\right\rVert}

\DeclareMathOperator*{\argmax}{arg\,max}

\makeatletter
    \newcommand*{\rom}[1]{\expandafter\@slowromancap\romannumeral #1@}
\makeatother

\newcommand{\etal}{\textit{et al}. }
\newcommand{\ie}{\textit{i}.\textit{e}. }
\newcommand{\eg}{\textit{e}.\textit{g}. }

\usepackage{caption}
\usepackage{subcaption}

\ifCLASSOPTIONcompsoc
  \usepackage[nocompress]{cite}
\else
  \usepackage{cite}
\fi
\ifCLASSINFOpdf
\else
\fi
\usepackage{amssymb}
\usepackage{amsmath}
\usepackage{mathtools}
\usepackage{algorithm}
\usepackage[noend]{algpseudocode}
\ifCLASSOPTIONcompsoc
 \usepackage{subcaption}
\else
 \usepackage[caption=false,font=footnotesize]{subfig}
\fi
\newcommand\MYhyperrefoptions{bookmarks=true,bookmarksnumbered=true,
pdfpagemode={UseOutlines},plainpages=false,pdfpagelabels=true,
colorlinks=true,linkcolor={black},citecolor={blue},urlcolor={black},
pdftitle={Bare Demo of IEEEtran.cls for Computer Society Journals},
pdfsubject={Typesetting},
pdfauthor={Michael D. Shell},
pdfkeywords={Computer Society, IEEEtran, journal, LaTeX, paper,
             template}}
\ifCLASSINFOpdf
\usepackage[\MYhyperrefoptions,pdftex]{hyperref}
\else
\usepackage[\MYhyperrefoptions,breaklinks=true,dvips]{hyperref}
\usepackage{breakurl}
\fi
\begin{document}
%
\title{Learning with Capsules: A Survey}

%
%
%
%

\author{Fabio~De~Sousa~Ribeiro,
        Kevin~Duarte,
        Miles~Everett,
        Georgios~Leontidis,
        and Mubarak~Shah
\IEEEcompsocitemizethanks{\IEEEcompsocthanksitem F. De Sousa Ribeiro is with the Department of Computing, Imperial College London, London, SW7 2RH, UK\protect\\
E-mail: f.de-sousa-ribeiro@imperial.ac.uk
\IEEEcompsocthanksitem K. Duarte and M. Shah are with the Center for Research in Computer Vision, University of Central Florida, Orlando-32816, Florida, USA.
\IEEEcompsocthanksitem M. Everett and G. Leontidis are with the Department of Computing Science and the Interdisciplinary Centre for Data and AI, University of Aberdeen, Aberdeen, AB24 3FX, UK}}
\IEEEtitleabstractindextext{%
\begin{abstract}
  Capsule networks were proposed as an alternative approach to Convolutional Neural Networks (CNNs) for learning object-centric representations, which can be leveraged for improved generalization and sample complexity. Unlike CNNs, capsule networks are designed to explicitly model part-whole hierarchical relationships by using groups of neurons to encode visual entities, and learn the relationships between these entities. Promising early results achieved by capsule networks have motivated the deep learning community to continue trying to improve their performance and scalability across several application areas. However, a major hurdle for capsule network research has been the lack of a reliable point of reference for understanding their foundational ideas and motivations. The aim of this survey is to provide a comprehensive overview of the capsule network research landscape, which will serve as a valuable resource for the community going forward. To that end, we start with an introduction to the fundamental concepts and motivations behind capsule networks, such as equivariant inference in computer vision. We then cover the technical advances in the capsule routing mechanisms and the various formulations of capsule networks, e.g. generative and geometric. Additionally, we provide a detailed explanation of how capsule networks relate to the popular attention mechanism in Transformers, and highlight non-trivial conceptual similarities between them in the context of representation learning. Afterwards, we explore the extensive applications of capsule networks in video and motion (e.g., video object segmentation, regression tracking, and action video recognition), natural language processing (e.g., text classification, relation extraction, language and vision, and recommendation systems), medical imaging (e.g., semantic segmentation of lesions, brain tumour classification), fault diagnosis (e.g, bearing fault diagnosis), hyperspectral imaging and forgery detection. To conclude, we provide an in-depth discussion regarding the main hurdles in capsule network research, and highlight promising research directions for future work.
\end{abstract}

}

\maketitle

\IEEEdisplaynontitleabstractindextext

%
\IEEEpeerreviewmaketitle

\ifCLASSOPTIONcompsoc
\IEEEraisesectionheading{
\section{Introduction}\label{sec:introduction}}
\else
\section{Introduction}\label{sec:introduction}
\fi
\begin{figure*}[!t]
    \hfill
    \centering
    \begin{subfigure}{0.33\textwidth}
        \centering
        \caption{Total Publications per Year}
        \label{fig:all_caps}
        \includegraphics[trim={0 0 0 25},clip,width=\columnwidth]{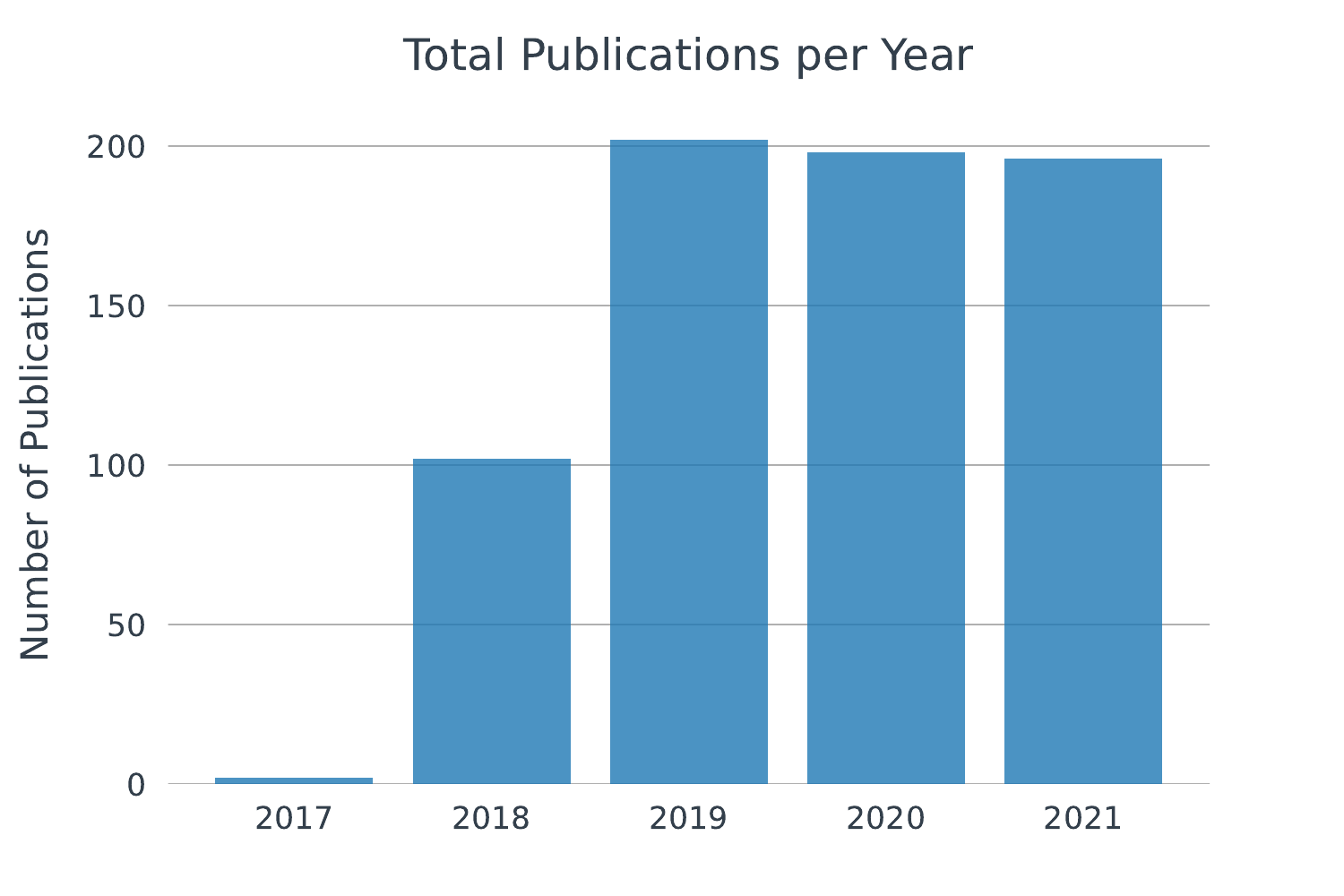}
    \end{subfigure}\hfill
    \begin{subfigure}{0.33\textwidth}
        \centering
        \caption{Top Venue Publications per Year}
        \label{fig:top_conf}
        \includegraphics[trim={0 0 0 25},clip,width=\columnwidth]{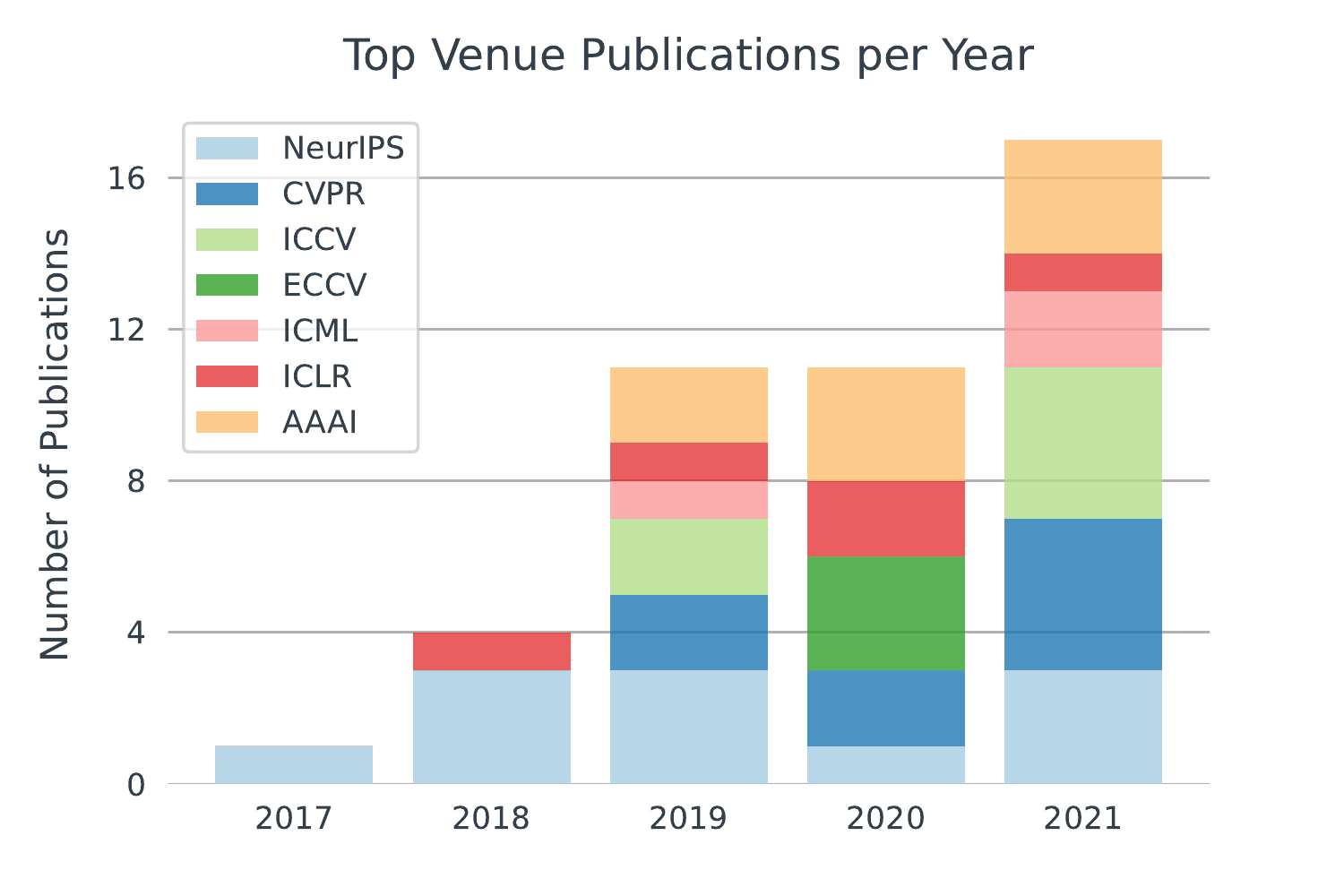}
    \end{subfigure}\hfill
    \begin{subfigure}{0.33\textwidth}
        \centering
        \caption{Total Publications per Topic}
        \label{fig:caps_categories}
        \includegraphics[trim={0 0 0 25},clip,width=\columnwidth]{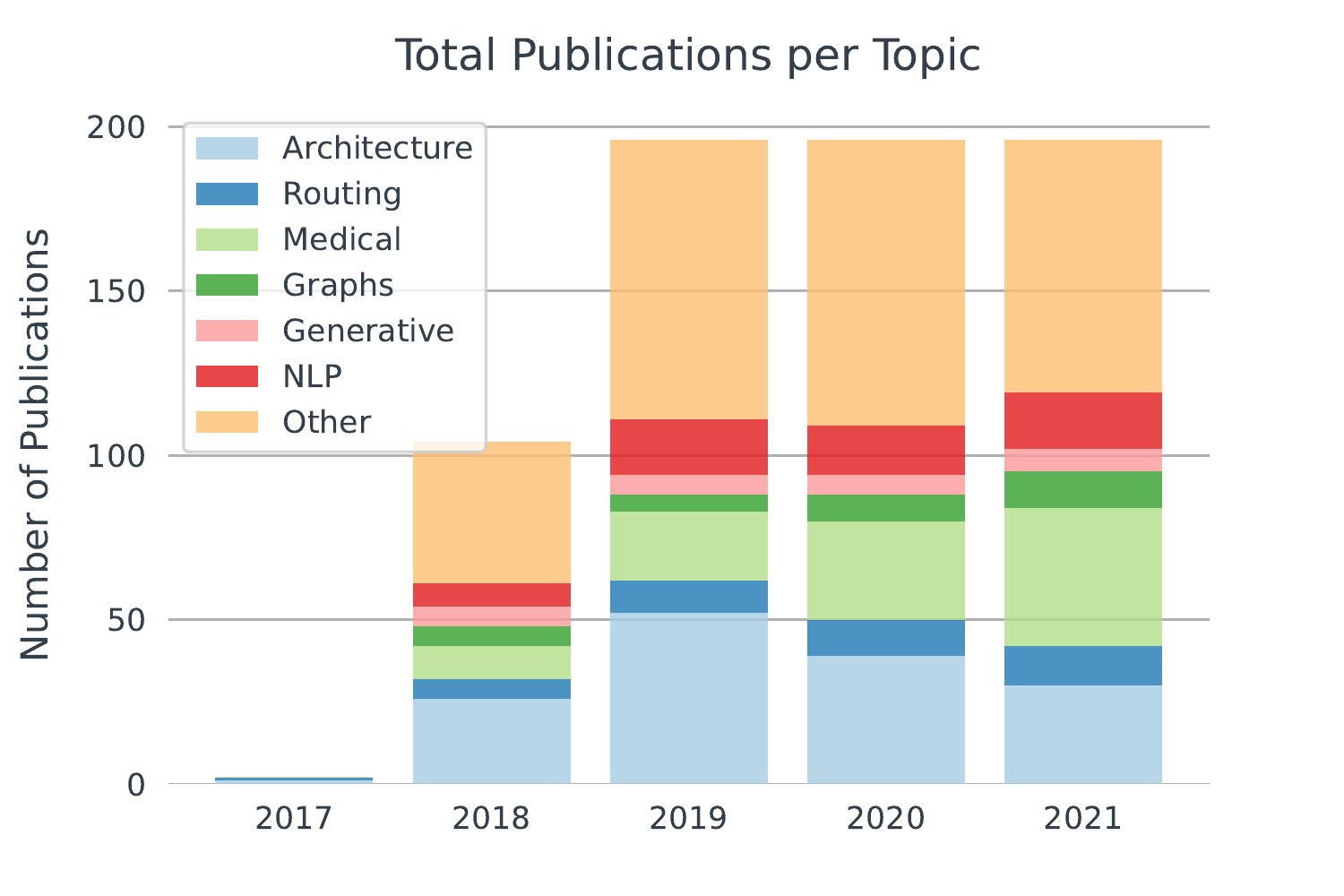}
    \end{subfigure}
    \caption{Statistics of Capsule Network related research activity. \textbf{(a)} The total number of capsule network related papers published in any scientific journal, conference or arXiv. \textbf{(b)} The total number of capsule network related papers published in top conference venues. \textbf{(c)} Categorical breakdown of the main research topics studied in capsule network literature.} 
    \label{fig:equivariance}
    \hfill
\end{figure*}
\IEEEPARstart{T}{he} quintessential task of computer vision is to classify an object from a vector of features extracted from an image, and to provide fuller descriptions such as its pose, shape, appearance etc. For decades, constructing data representations (features) that were suitable for downstream tasks involved extensive hand-engineering and expert knowledge. Representation learning~\cite{bengio2013representation} consists of a set of tools which enable a machine to automatically discover useful representations of raw data, which may then be utilised for downstream predictive tasks. The most successful representation learning method in recent years is Deep Learning (DL)~\cite{lecun2015deep}. Despite the many successes of modern DL-based vision systems~\cite{krizhevsky2012imagenet,lecun2015deep,he2016deep}, a general lack of robustness to distributional shifts remains prevalent~\cite{greff2020binding}. Indeed, unlike current systems, humans are able to quickly adapt to distributional changes using very few examples to learn from~\cite{bruce1994recognizing,cohen2017,bengio2021deep}. 
There is compelling evidence that humans parse visual scenes into part-whole hierarchies, and that we do so by modelling the viewpoint-invariant spatial relationship between a part and a whole, as the coordinate transformation between
the intrinsic coordinate frames assigned to them~\cite{rock1973orientation,hinton1979some,kahneman1992reviewing}. One way to make Neural Networks (NN) more transparent and interpretable, is to try to make them understand images in the same way humans do. However, this is difficult for standard NNs because they cannot dynamically represent a different part-whole hierarchy tree structure for each image~\cite{hinton2021represent}. This inability was the motivation for a series of models called capsule networks~\cite{hinton2011transforming,sabour2017dynamic,hinton2018matrix,kosiorek2019stacked,hinton2021represent}. A capsule network is a type of NN that is designed to model part-whole hierarchical relationships more explicitly than Convolutional Neural Networks (CNNs), by using groups of neurons to encode entities and learning the relationships between these entities~\cite{lecun1989backpropagation}. Like many other developments in machine learning~\cite{goodfellow2016deep,pashler2016attention}, capsule networks are biologically inspired, and their goal is to be able to learn more robust object-centric representations that are pose-aware and interpretable. Evidence from neuroscience suggests that groups of tightly-connected nearby neurons (i.e. hypercolumns) could represent a vector-valued unit which is able to transmit not only scalar quantities, but a set of coordinated values~\cite{bengio2021deep}. This idea of vector-valued units is at the heart of both capsule networks and soft-attention mechanisms~\cite{bengio2021deep,bahdanau2014neural,sukhbaatar2015end}, including the transformer~\cite{vaswani2017attention}. As shown later, in capsule networks these vector-valued units are known as capsules, and in transformers they are represented by query, key and value vectors. Performing operations such as the scalar product between neural activity vectors, enables powerful algorithmic concepts such as coincidence filtering and attention to be computed.

Despite the promising progress on capsule works, Barham \etal \cite{barham2019machine} explained that although their convolutional capsule model required around 4 times fewer floating point operations (FLOPS) with 16 times fewer parameters than their CNN, implementations in both TensorFlow~\cite{abadi2016tensorflow} and PyTorch~\cite{paszke2019pytorch} ran significantly slower and ran out of memory with much smaller models. Although several more efficient versions of capsule routing have since then been proposed~\cite{self-routing,ahmed2019star,Tsai2020Capsules,ribeiro2020capsule}, the underlying problem is not only caused by routing but by the capsule voting procedure as well. In their analysis, \cite{barham2019machine} conclude that current frameworks have been highly optimised for a small subset of computations used by a popular family of models, and that these frameworks have become poorly suited to research since there is a huge discrepancy in performance between standard and non-standard compute workloads. As a result, non-standard workloads like those induced by the routing and voting procedures in capsule networks are a lot slower than they could be. As pointed out by~\cite{hooker2020hardware}, while capsule network's operations can be implemented reasonably well on CPUs, performance drops drastically on accelerators like GPUs and TPUs since they have been heavily optimized for standard workloads using the building blocks found in common architectures. We hope this survey will inspire researchers to develop suitable tools for capsule networks.

In this survey, we provide a comprehensive overview of representation learning using capsule networks and related attention-based models. Although research on capsules is still at an early stage relatively speaking, Figure~\ref{fig:all_caps} shows us that despite an initial rapid growth in popularity, the total number of publications per year has somewhat stagnated. This is possibly due to the high barrier of entry to the field and lack of a reliable point of reference. Nonetheless, as shown in Figure~\ref{fig:top_conf}, the number of capsule network related publications at the top venues has continued to steadily increase. We believe that there is now sufficient material to warrant a detailed organisation of the various concepts, techniques and foundational ideas which would benefit the community and spark research interest in the area. At the time of this writing, there exist only three other Capsule Network based surveys.~\cite{vijayakumar2019comparative} was written shortly after Capsule Networks were first introduced, so it does not cover large milestones achieved more recently. Similarly \cite{shi2020brief} was written with the purpose of being brief and therefore covers a very small portion of the relevant literature. \cite{patrick2022capsule} is more recent, and covers a larger breadth of papers, but does not extensively survey the field with sufficient detail in the way we feel is necessary. Conversely, the purpose of this survey is to provide the first comprehensive and detailed breakdown of capsule networks and related research on object-centric representation learning. Specifically, we aim to: (a) Explain the foundations, motivations and fundamental concepts behind capsule networks in detail; (b) Survey the state of the art in capsule network research in various application areas; (c) Relate and compare capsules and routing-by-agreement with Transformers and self-attention. (d) Discuss open problems and provide promising future research directions. We anticipate that our survey will serve as the main point of reference on capsule networks going forward, and will help contribute towards the advancement of the field. 

This survey is organised as follows. In Section 1, we provide an introductory overview of the ideas behind capsule networks. In Section 2, we begin with a gentle introduction to \textit{invariance} and \textit{equivariance}, and explain why these concepts are fundamental in representation learning. In Section 3 we explain the foundational ideas and motivations behind capsule networks, and introduce basic concepts such as \textit{agreement} and \textit{capsule routing}. In Section 4, we delve into the most prominent capsule routing algorithms proposed in literature. In section 5, we uncover the conceptual similarities between capsule routing and the popular self-attention mechanism in Transformers. Sections 6 to 10 discuss some major applications of capsule networks for video and motion, graphs, natural language processing, and medical imaging. Section 11 focuses on other applications of capsule networks, such as fault diagnosis, hyperspectral imaging, forgery detection, and adversarial attacks. Lastly, in Section 12 we discuss open challenges and shortcoming of capsule networks, along with what we believe are promising directions for future research.
\begin{figure*}[!t]
    \centering
    \begin{subfigure}{0.495\textwidth}
        \centering
        \caption{Translation Invariance}
        \label{fig:invariance_a}
        \includegraphics[trim={6 15 52 38},clip,width=.81\columnwidth]{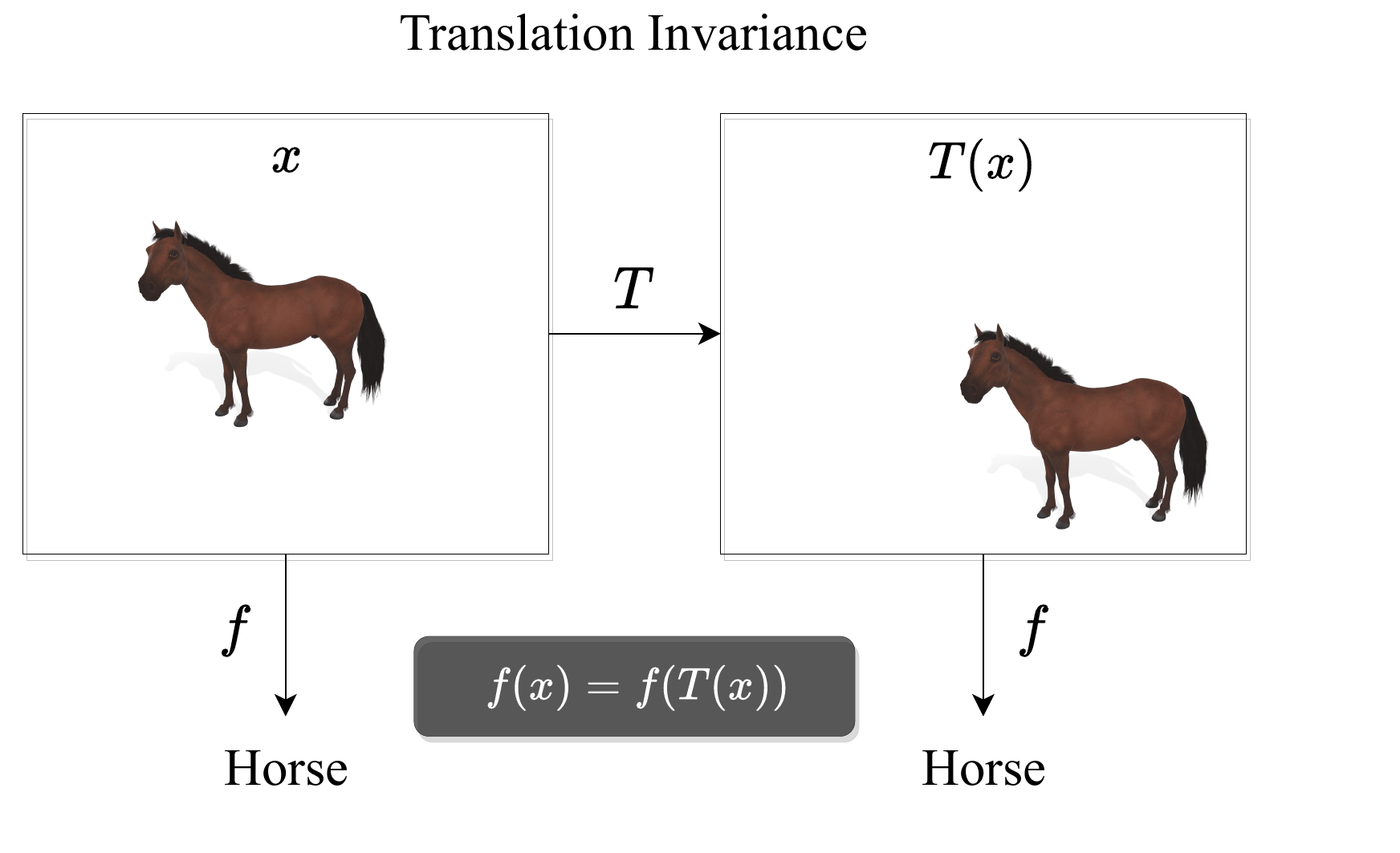}
    \end{subfigure}\hfill
    \begin{subfigure}{0.495\textwidth}
        \centering
        \caption{Viewpoint Invariance}
        \label{fig:invariance_b}
        \includegraphics[trim={6 15 7 38},clip,width=.81\columnwidth]{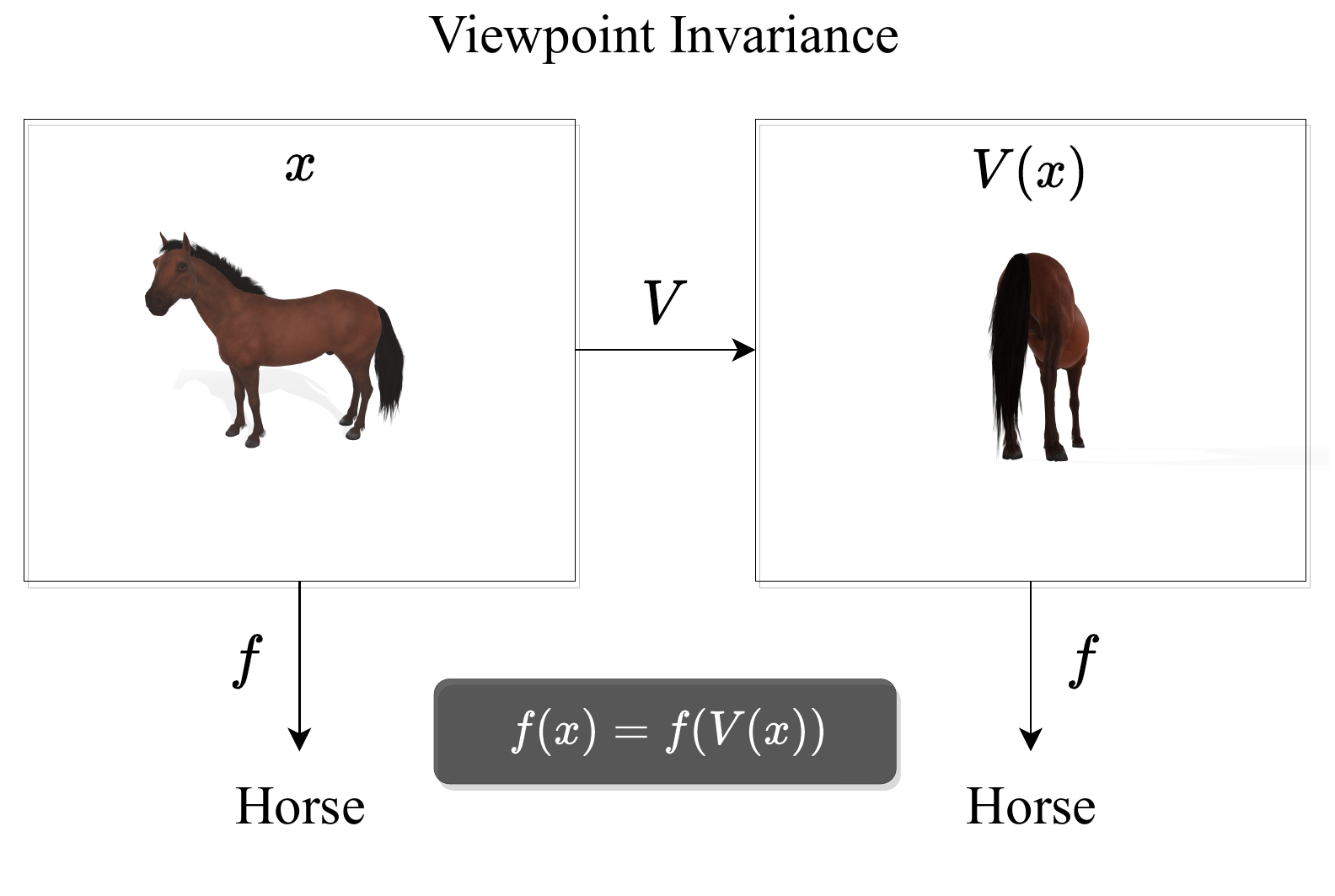}
    \end{subfigure}
    \\[5pt]
    \centering
    \begin{subfigure}{0.495\textwidth}
        \centering
        \caption{Translation Equivariance}
        \label{fig:equivariance_a}
        \includegraphics[trim={71 12.5 53 38},clip,width=.81\columnwidth]{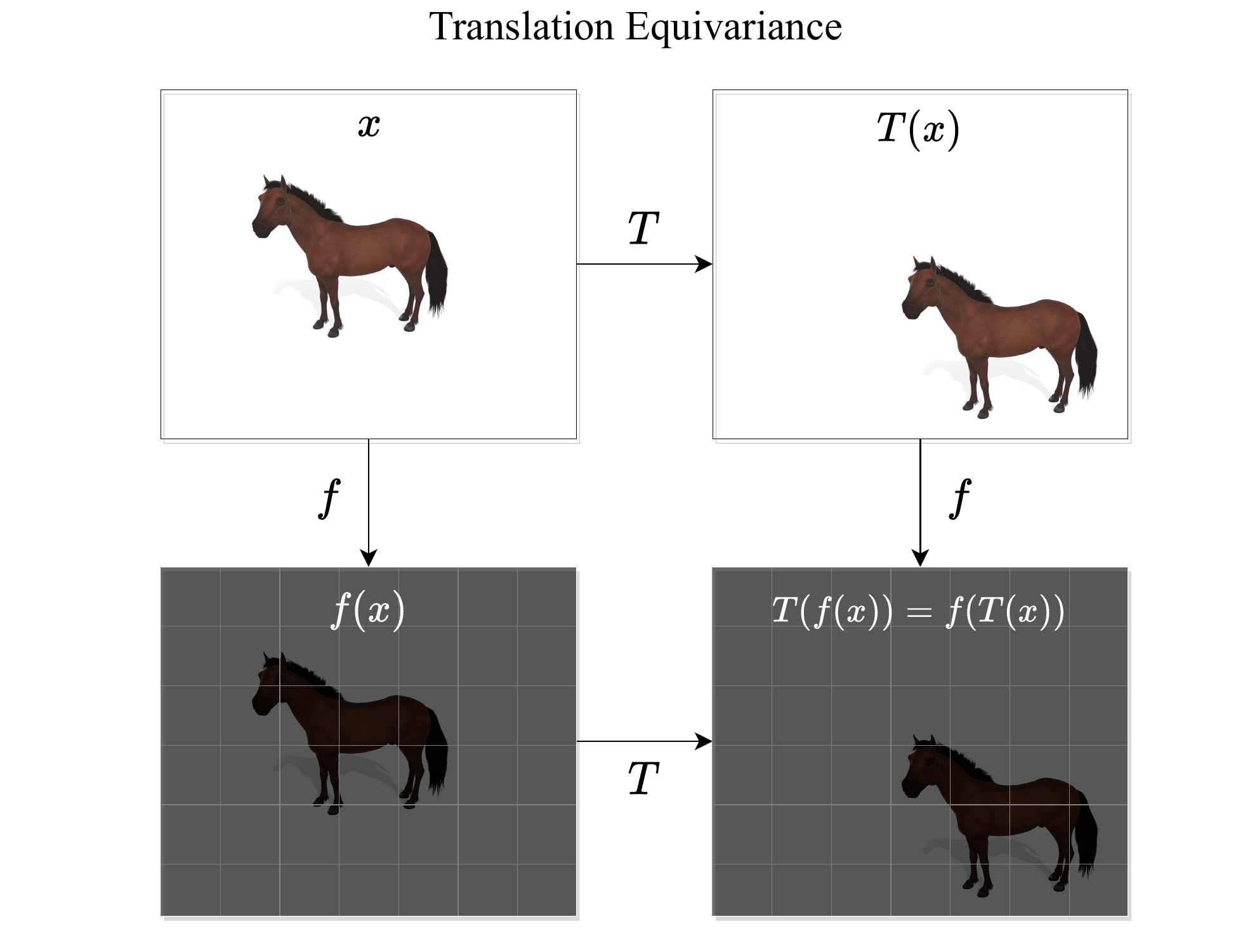}
    \end{subfigure}\hfill
    \begin{subfigure}{0.495\textwidth}
        \centering
        \caption{Viewpoint Equivariance}
        \label{fig:equivariance_b}
        \includegraphics[trim={56 0 51 38},clip,width=.81\columnwidth]{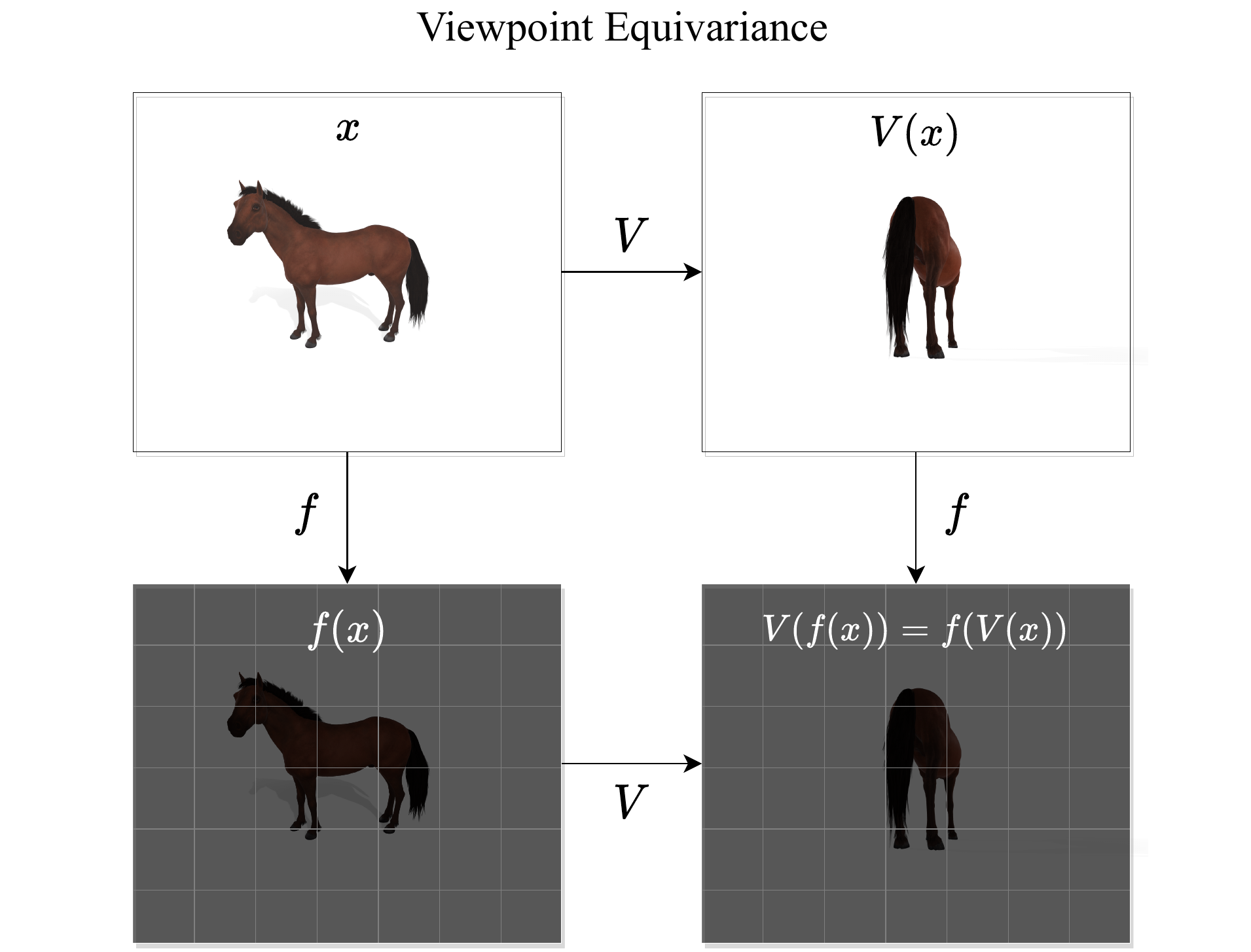}
        \makebox[6pt][r]{
            \raisebox{0em}{
            \includegraphics[trim={0 0 0 0},clip,width=.925 
            \columnwidth]{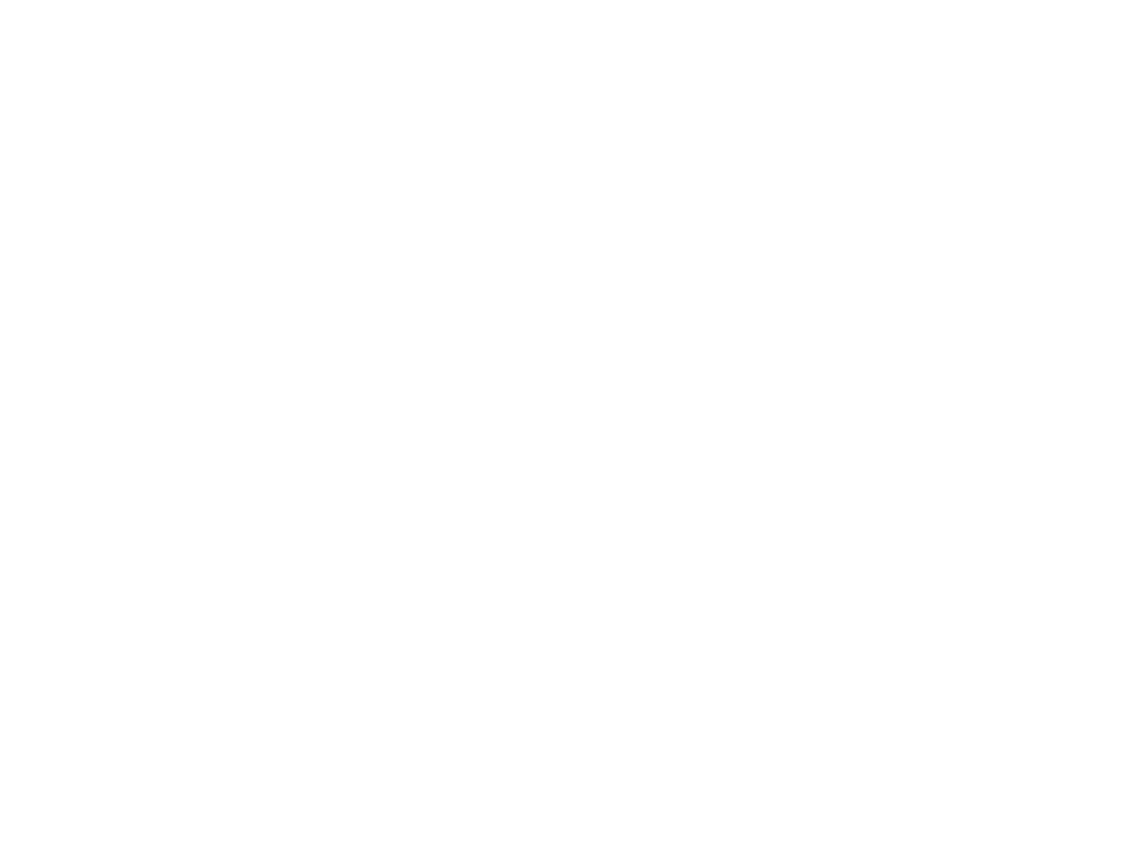}
            }\hspace*{0em}
        }
    \end{subfigure}
    \caption{Depiction of \textbf{invariance} and \textbf{equivariance} properties. Translation (or shift) is denoted by $T$, viewpoint (perspective) projection is denoted by $V$, and $f:\mathbb{R}^d \xrightarrow{} \mathbb{R}^k$ denotes a feature mapping. \textbf{(a)} E.g. a CNN with Global Average Pooling (GAP). \textbf{(b)} Capsule networks are \textit{explicitly} wired to try to produce viewpoint invariant predictions. \textbf{(c)} Convolution in CNNs, whereby changes in input translation lead to equivalent (\textit{place-coded}) changes in neural activities. \textbf{(d)} Cartoon example of capsule layers, which look to capture \textit{rate-coded} viewpoint equivariance in neural activity vectors called capsules.
    }
    \label{fig:equivariance_horse}
\end{figure*}
\section{Background \& Motivation}\label{sec:background}
To motivate capsule networks, we begin with a gentle introduction to \textit{invariance} and \textit{equivariance}, and explain why these concepts are fundamental in representation learning.

\subsection{Invariance}
Invariance is a useful property to model for a variety of recognition tasks, as we'd often like the final prediction of our model to be \textit{invariant} to transformations of the input that preserve intrinsic properties, such as relative positions and symmetries. A \textit{symmetry} of an object is a transformation that leaves it unchanged, e.g. rotating a (perfect) circle about its center (rotational invariance). For example, the intrinsic properties of the `Horse' in Figure~\ref{fig:equivariance_horse} remain unchanged relatively speaking when translating, scaling or flipping it, thus our model's prediction of `Horse' should be the same under these transformations. This notion of invariance is also linked to model generalisation and design~\cite{bengio2013representation,jaderberg2015spatial}.

\textbf{Translation Invariance.} \ To humans, both images shown in Figure~\ref{fig:invariance_a} should be classified as a `Horse' regardless of where it appears within the boundaries. This is because a positional shift (translation $T$) of the horse does not change its intrinsic properties. Concretely, \textit{translation invariance} refers to a feature mapping $f$ that produces the same output (e.g. `Horse') regardless of input translation $T(x)$. More formally, $f$ is invariant to input translations of $x$ if:
\begin{equation}
    f(x) = f(T(x)).
\end{equation}
Sub-sampling techniques typically used in CNNs, such as max-pooling, make neural activities of the next layer locally invariant to input translations~\cite{bengio2013representation,zhang2019making}. That is, the output of a pooling unit is the same irrespective of where a specific feature is located inside its pooling region. 

\textbf{Viewpoint Invariance.} \ A more challenging invariance to model is \textit{viewpoint invariance} as shown in Figure~\ref{fig:invariance_b}. By the same logic as before, a feature mapping $f$ is invariant to viewpoint projection $V$ of the input $x$ if: $f(x) = f(V(x))$. To humans, this task is relatively trivial since we are very good at extrapolating object appearance to novel viewpoints. However, this is not the case for standard CNNs~\cite{hinton2011transforming}. As explained later, unlike CNNs, capsule networks are explicitly wired to try to capture viewpoint invariance in the network's weights to produce viewpoint invariant predictions. Capsules attempt to encode explicit pose representations of parts and objects, as any change in viewpoint can be modeled by a \textit{linear} operation on these poses.
\begin{figure}[t!]
    \centering
    \includegraphics[trim={0 0 0 0},clip,width=.85\columnwidth]{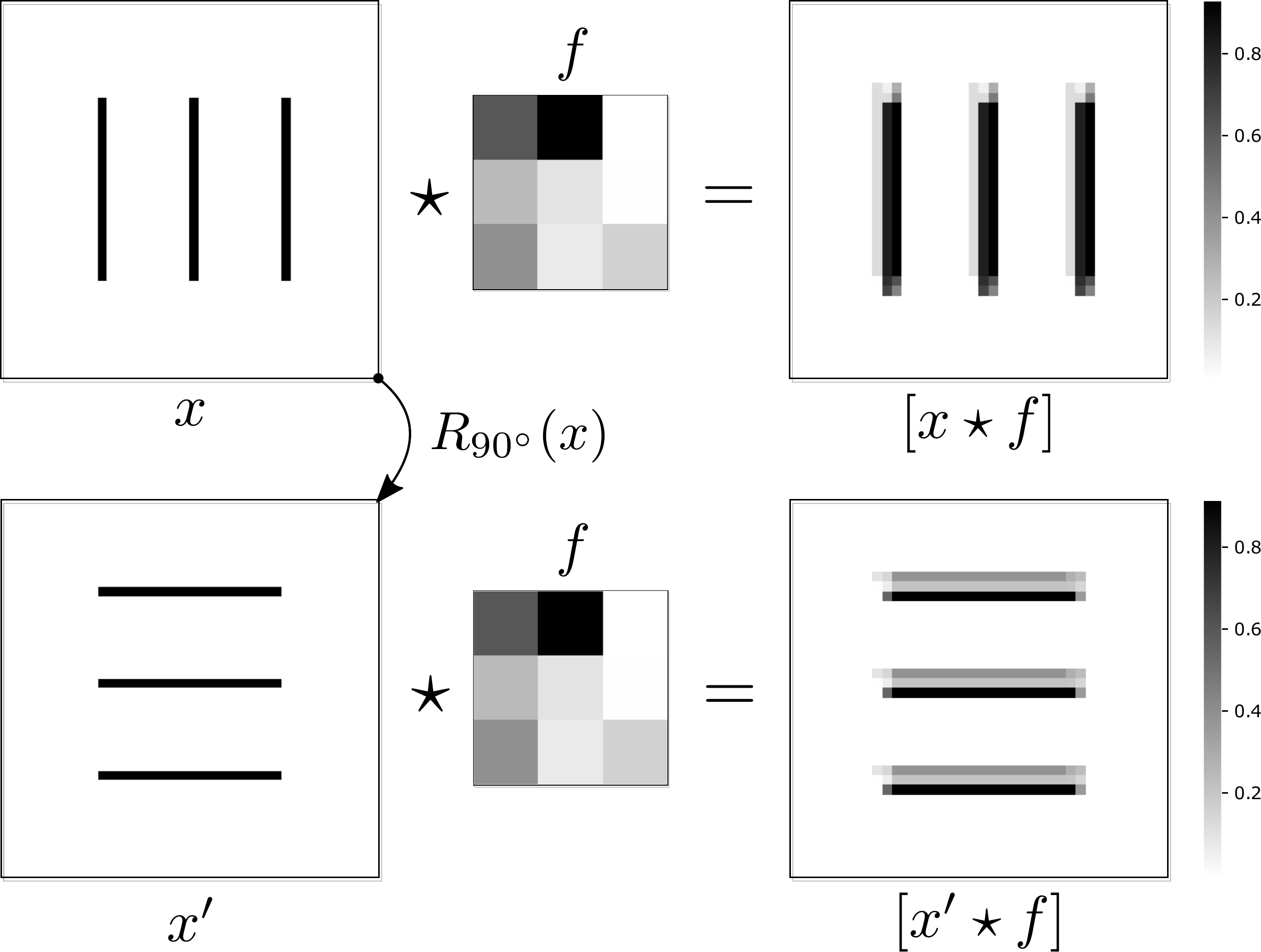}
\caption{\label{fig:rotation} Example of lack of equivariance in convolution. Each filter weight was sampled as: $f_i \sim \textrm{Uniform}(0,1)$.
Unlike the translation case, the output $[x' \star f]$ is \textit{not} simply a rotated version of $[x \star f]$ as convolution is \textit{not} rotation equivariant.}
\vspace{-3pt}
\end{figure}
\subsection{Equivariance}
The success of CNNs can be largely attributed to their ability to exploit translation \textit{symmetry} to reduce sample complexity. Indeed, the convolution operator and weight sharing provides the useful property of \textit{equivariance} under translation, enabling efficient spatial transfer of knowledge. Naturally, much research in recent years has focused on exploiting other transformations and symmetries such as rotation and scale to improve statistical efficiency~\cite{hinton2011transforming,jaderberg2015spatial,cohen2016group,worrall2017harmonic}. Two fields of particular interest are Group CNNs~\cite{cohen2016group,dieleman2016exploiting,s.2018spherical,romero2020attentive} and capsule networks~\cite{sabour2017dynamic,hinton2018matrix,kosiorek2019stacked,NEURIPS2020_47fd3c87}, which are both predicated upon the notion that intermediate NN layers should \textit{not} be fully invariant, because the relative pose of local features should be preserved for future layers to aid in generalisation to new transformations~\cite{hinton2011transforming,cohen2016group}.

\textbf{Translation Equivariance.} \ One of the simplest examples of equivariance is \textit{translation equivariance} afforded by the convolution operation in CNNs~\cite{lecun1989handwritten}. As shown in Figure~\ref{fig:equivariance_a}, a change in translation of the input leads to equivalent changes in neural activities. More formally, $f$ is equivariant with respect to (w.r.t.) translation $T$ if:
\begin{equation}
    f(T(x)) = T(f(x)),
\end{equation}
where $f$ denotes the convolution operation in this example. In other words, we can first translate $x$ then convolve, or first convolve $x$ then translate to obtain the same output. In this case, this is known as \textit{place-coded} equivariance, since a discrete change in the input $x$ results in a discrete change in which neurons are used to encode it. 

\textbf{Viewpoint Equivariance.} \ It is more challenging to capture \textit{viewpoint equivariance} in a model (see Figure~\ref{fig:equivariance_b}). That is, changes in viewpoint that lead to equivalent changes in neural activities. Like before, $f$ is said to be equivariant w.r.t. viewpoint (perspective) projection $V$ of the input $x$ if: $f(V(x)) = V(f(x))$. Convolution is \textit{not} equivariant to transformations other than translation, which makes it challenging for CNNs to deal with viewpoint changes~\cite{hinton2011transforming,cohen2016group} (see Figure~\ref{fig:rotation}). As explained later, capsule networks~\cite{sabour2017dynamic,hinton2018matrix,kosiorek2019stacked} look to move from \textit{place-coded} to \textit{rate-coded} viewpoint equivariance in the final layers, whereby a real-valued change in the input results in an equivalent real-valued change in neuronal output (capsule pose vectors).

%
\section{Capsule Network Foundations}
\label{subsec:foundations}
Although capsule networks have taken on several different forms since their inception~\cite{hinton2011transforming,sabour2017dynamic,hinton2018matrix,kosiorek2019stacked}, they are generally built upon the following core assumptions and premises~\cite{NEURIPS2020_47fd3c87}:
\begin{enumerate}[label=(\roman*)]
    \item Capturing \textbf{equivariance} w.r.t. viewpoints in neural activities, and \textbf{invariance} in the network's weights;
    \item High-dimensional coincidences are effective feature detectors, e.g. using the dot product to compute the similarity between neural activity vectors;
    \item Viewpoint changes have \textbf{non-linear} effects on pixel intensities, but \textbf{linear} effects on part-object relationships;
    \item Object parts belong to a single object, and each location contains at most a single object.
\end{enumerate}
In theory, a perfect instantiation of the above premises could yield more sample-efficient models, that leverage robust representations to better \textit{generalise} to unseen cases. Unlike current methods, humans can often extrapolate object appearance to novel viewpoints even after a single initial observation. Evidence suggests that this is because we impose coordinate frames on objects~\cite{rock1973orientation,hinton1979some}. 
Capsules imitate this concept by representing neural activities as poses of objects with respect to a coordinate frame imposed by an observer, and attempt to \textit{disentangle} salient features of objects into their composing parts. This is reminiscent of inverse graphics~\cite{kulkarni2015deep}, but is not explicitly enforced in capsule formulations since the learned pose matrices are not constrained to be interpretable geometric forms. 

Capsule networks can also be viewed as an extension of the successful inductive biases already present in CNNs, by wiring in some additional complexity to deal with viewpoint changes. 
The desired effect is to produce \textit{viewpoint invariant} predictions, and align the learned representations with those perceptually consistent to humans, such that adversarial examples become less effective~\cite{Qin2020Detecting,bengio2021deep}. 
\begin{figure}[t!]
    \centering
    \includegraphics[trim={0 0 0 0},clip,width=.99\columnwidth]{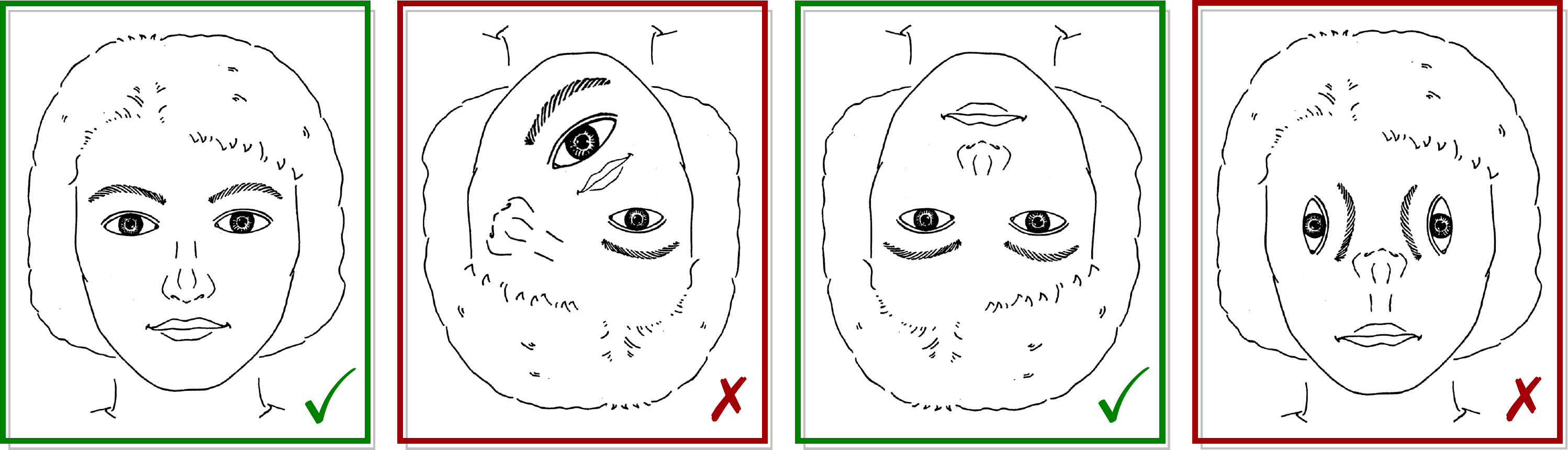}
    \caption{Example of the `Picasso problem'. Relative part-object relationships ought to be preserved if we are to label each image as a person's face. Inspired by Figure 1 from~\cite{schwarzer2000development,romero2020attentive}.}
    \label{fig:picasso}
\end{figure}
\subsection{The Picasso Problem}
In the ideal case, capsule networks address the `Picasso problem': i.e. images of an \textit{object} containing all the right \textit{parts} -- but that are not in the correct spatial relationship -- are often misclassified as said \textit{object} by typical DL-based systems. To gain some intuition, consider the example of an image of a person's face (\textit{object}), whereby the positions of the various \textit{parts} of the face (e.g. mouth, eyes and nose) are shuffled, and the image is still (wrongly) classified as a person's face. As depicted in Figure~\ref{fig:picasso}, although every image is composed of the same \textit{parts}, only the green bordered examples ought to be labelled as a person's face~\cite{schwarzer2000development}.
\begin{figure}[t!]
    \centering
    \captionsetup[subfigure]{labelformat=empty}
    \begin{subfigure}{0.33\columnwidth}
        \centering
        \frame{\includegraphics[width=.82\columnwidth]{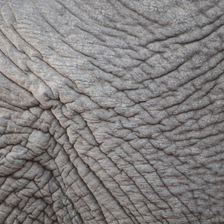}}
      \captionsetup{justification=centering,font={footnotesize}}
      \caption{Texture image\\[4pt]
      {
      \begin{tabular}{rl}
        81.4\% & \textbf{\texttt{elephant}}\\
        10.3\% & \texttt{indri}\\
         8.2\% & \texttt{black swan}
      \end{tabular}
      }
      }
      \end{subfigure}\hfill
    \begin{subfigure}{0.33\columnwidth}
        \centering
        \frame{\includegraphics[width=.82\columnwidth]{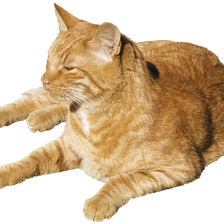}}
      \captionsetup{justification=centering,font={footnotesize}}
      \caption{Content image\\[4pt]
      {
      \begin{tabular}{rl}
        71.1\% &\textbf{\texttt{tabby cat}}\\
        17.3\% &\texttt{grey fox}\\
         3.3\% &\texttt{siamese cat}
      \end{tabular}
      }
      }
    \end{subfigure}\hfill
    \begin{subfigure}{0.33\columnwidth}
        \centering
        \frame{\includegraphics[width=.82\columnwidth]{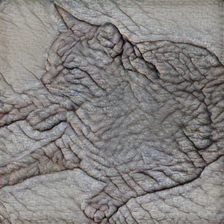}}
      \captionsetup{justification=centering,font={footnotesize}}
      \caption{Cue conflict
      \\[6pt]
      {
      \begin{tabular}{rl}
      63.9\% &\textbf{\texttt{elephant}}\\
      26.4\% &\texttt{indri}\\
       9.6\% &\texttt{black swan}
      \end{tabular}
      }
      }
      \label{subfig:catefant}
    \end{subfigure}\hfill
    \caption{Evidence of current CNNs (ResNet-50) being biased towards textures rather than using shapes like humans. Applying the `elephant' texture leads to misclassification. Figure from ~\cite{geirhos2018imagenet}.
    }
    \label{fig:intro_catefant}
\end{figure}

Classifying an image of an object viewed from a very different angle than those seen during model training also often leads to misclassification. Intuitively, this occurs because typical modern vision systems such as CNNs are not wired to explicitly model relative positions and spatial relationships between parts and objects. Instead, they tend to focus on detecting the most generally discerning properties of the input in the hope it produces the correct outcome~\cite{hinton2011transforming,geirhos2018imagenet}. In fact, as discussed later, CNNs focus much more on textures than shapes, unlike humans~\cite{geirhos2018imagenet} (see Figures~\ref{fig:intro_catefant},~\ref{fig:texture}). One inefficient way of mitigating the above issue is to use data augmentation to provide the CNN model with examples of objects from all possible angles. However, this approach is not general, and can become infeasible in real world scenarios due to lack of data.

A more efficient way to solve this problem would be to decompose the images into their constituent parts and objects, and use the linearity of part-object spatial relationships (i.e. simple pose matrix multiplication used in computer graphics) to generalise to \textit{all} viewpoints at once -- which is the goal of capsule networks. 
\subsection{Sub-sampling \& Convolution}
\begin{figure}[t!]
    \centering
    \frame{\includegraphics[trim={0 13 0 0},clip,width=.85\columnwidth]{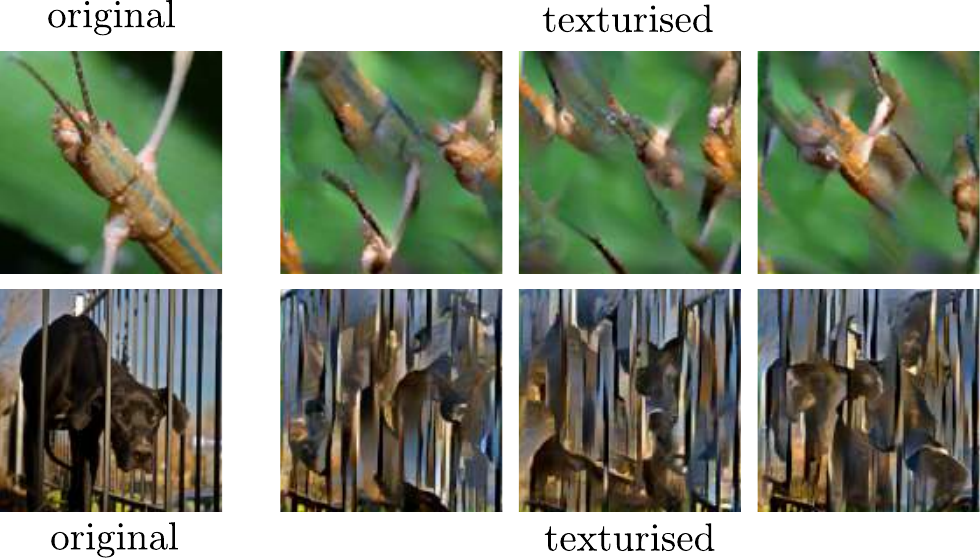}}
\caption{\label{fig:texture} Example of texture bias in CNNs. CNNs still achieve high accuracy on the texturised images, whereas humans do not due to the loss of global shape information. Figure from ~\cite{brendel2019approximating}.}
\vspace{-3.5pt}
\end{figure}

The issue of poor generalisation to novel viewpoints in CNNs is exacerbated by \textit{pooling} (sub-sampling) operators, which discard potentially pose-aware information in favour of training/inference speed and performance~\cite{sabour2017dynamic,hinton2018matrix}. Indeed, there is strong evidence that modern deep CNNs do not parse images like humans, and that they are in fact biased towards textures and other properties rather than shapes~\cite{geirhos2018imagenet,geirhos2018generalisation,brendel2019approximating}. For examples of this phenomenon see Figures~\ref{fig:intro_catefant} and~\ref{fig:texture}. These biases are also corroborated by the fact that adversarial examples are often visually indistinguishable to humans~\cite{szegedy2013intriguing,goodfellow2014explaining}. On that note, \cite{Qin2020Detecting} showed that capsule networks use features that are more aligned with human perception, and therefore have the potential to address the central issue raised by adversarial examples. Moreover, \cite{hinton2018matrix} demonstrated that capsule networks can better generalise to novel viewpoints compared to CNNs of a similar size, and \cite{sabour2017dynamic} showed that capsule networks are considerably better than CNNs at recognising overlapping digits. Although research on capsule networks is still in its infancy, there are representational reasons for believing that it is a better approach to vision, and early results on pose-aware tasks highlight their potential~\cite{sabour2017dynamic,hinton2018matrix,NEURIPS2020_47fd3c87}.
\begin{figure}[b!]
    \centering
    \includegraphics[trim={0 0 0 0},clip,width=.75\columnwidth]{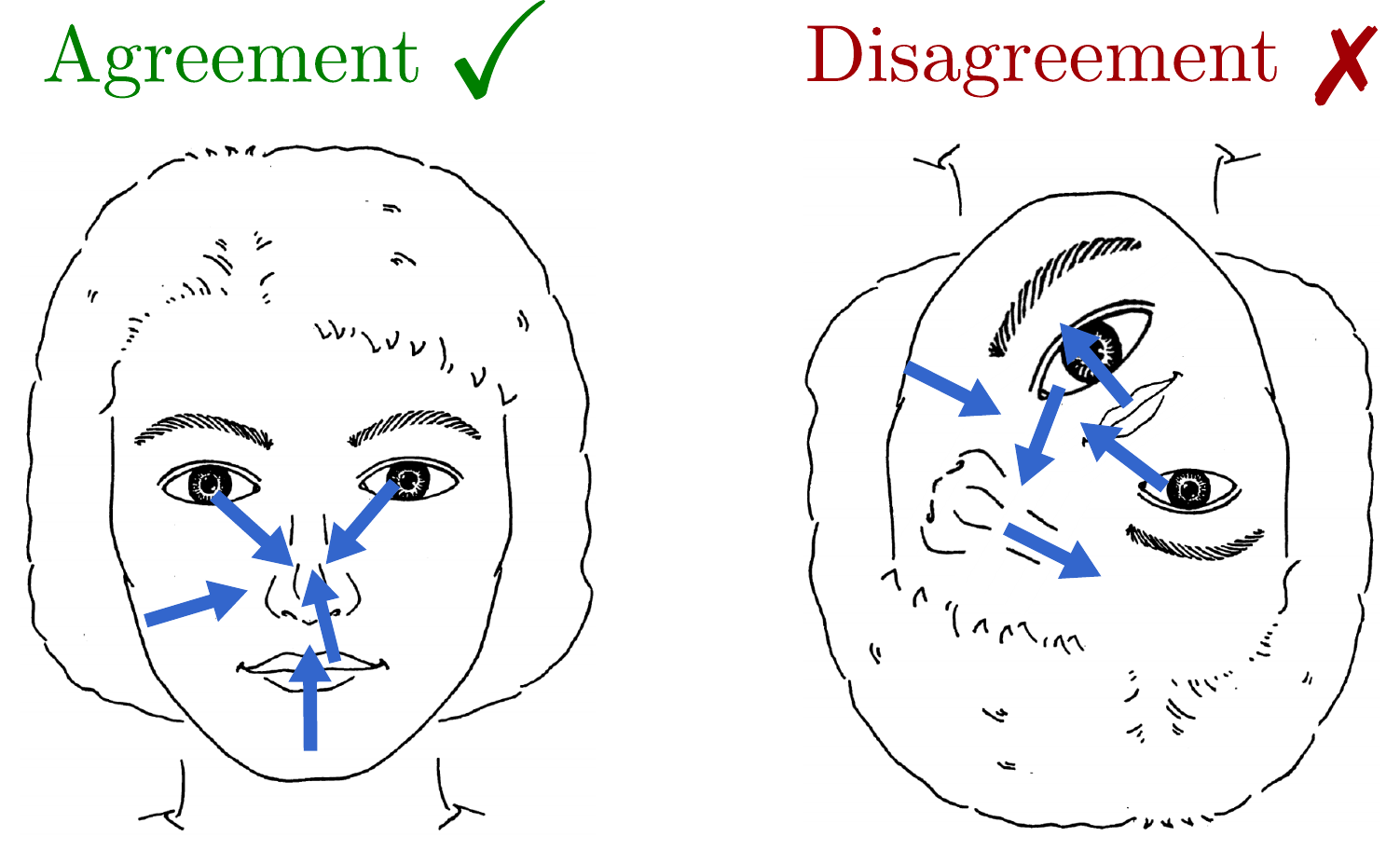}
    \caption{Example of agreement and no agreement between part poses (eyes, nose, mouth etc) with respect to the object (person), where each pose is encoded by a vector. Capsule routing aims to detect objects by looking for agreement between its parts, and thereby perform equivariant inference.}
    \label{fig:agreement_faces}
\end{figure}
\subsection{Capsule Networks} 
\begin{figure*}
        \centering
        \includegraphics[trim={0 0 0 0},clip,width=.85\textwidth]{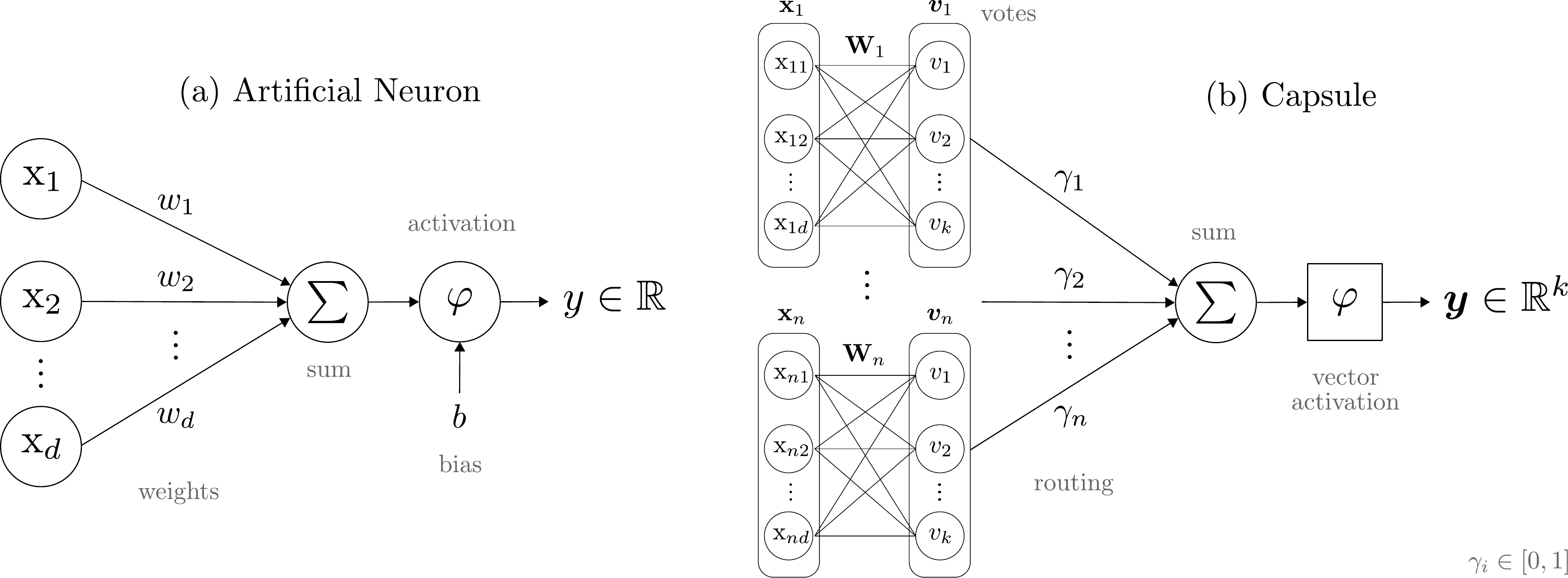}
        \caption{\textbf{(a)} Illustration of the artificial neuron as found in standard neural networks. The artificial neuron receives an input signal vector $\mathbf{x} \in \mathbb{R}^d$ and outputs a scalar quantity $y$. \textbf{(b)} Depiction of a capsule as found in capsule networks. Unlike the neuron, a capsule receives $n$ input signal vectors $\mathbf{x}_i \in \mathbb{R}^d$, and outputs a vector $\bm{y} \in \mathbb{R}^k$ of neural activities.}
        \label{fig:neuron_capsule}
\end{figure*}

In order to explain what a capsule network is, we begin by comparing the traditional \textit{artificial neuron} found in standard NNs, and the \textit{capsule} as shown in Figure~\ref{fig:neuron_capsule}. 

\textbf{Capsule.} \ A capsule is a biologically inspired structure based on \textit{hypercolumns}, wherein groups of tightly-connected nearby neurons are thought to represent vector-valued units which are able to transmit not only scalar quantities, but a set of coordinated values~\cite{bengio2021deep}. 

Formally, a capsule is a parameterised function: $\mathbf{c}(\mathcal{X};\mathcal{W}) : \mathbb{R}^{N \times D} \rightarrow \mathbb{R}^P$, where $\mathcal{X} = \{\mathbf{x}_i\}_{i=1}^{N}$ is a set of $N$ input signals $\mathbf{x}_i \in \mathbb{R}^D$, and $\mathcal{W} = \{\mathbf{W}_i\}_{i=1}^{N}$ is a set of $N$ transformation weight matrices $\mathbf{W}_i \in \mathbb{R}^{P \times D}$. A \textit{dynamic routing}
process -- akin to clustering -- gives rise to \textit{routing} coefficients: $\boldsymbol{\gamma} = \{\gamma_i\}_{i=1}^N, \ \gamma_i \in [0,1]$, which represent the affinity between each input signal vector $\mathbf{x}_i$ and the output capsule in question. As explained later, the process by which these routing coefficients are obtained takes into account the context from other capsules. Lastly, an activation function $\varphi(\cdot)$ is applied, and the output capsule $\bm{y} \in \mathbb{R}^P$ is given by
\begin{align}
    \bm{y} = \varphi \Big( \sum_{i=1}^N \gamma_i \mathbf{W}_i \cdot \mathbf{x}_i\Big).
\end{align}
Notice that unlike the artificial neuron, a capsule outputs a vector $\bm{y}\in \mathbb{R}^P$ of neural activities rather than a scalar $y \in \mathbb{R}$. Indeed, a capsule operates on a set of $N$ input signal vectors and $N$ parameter matrices, rather than a single input signal vector and weight vector. The neural activities within a capsule aim to encode the various properties of the entity it learns to represent, such as its pose, colour, texture etc.

\textbf{Capsule Layer.} \ In a capsule layer we typically have $i=1,\dots,N$ lower level \textit{part} capsules $\mathbf{x}_i \in \mathbb{R}^D$ (considered the inputs), and $j=1,\dots,M$ higher level \textit{object} capsules $\bm{y}_j \in \mathbb{R}^P$ (considered the outputs). As explained in greater detail later, the initial part capsules are extracted from the raw input (e.g. images), and the object capsules of a layer $\ell$ become the part capsules of the next layer up in a hierarchical fashion until the final layer.

\begin{figure}[!t]
    \centering
    \includegraphics[width=.85\columnwidth]{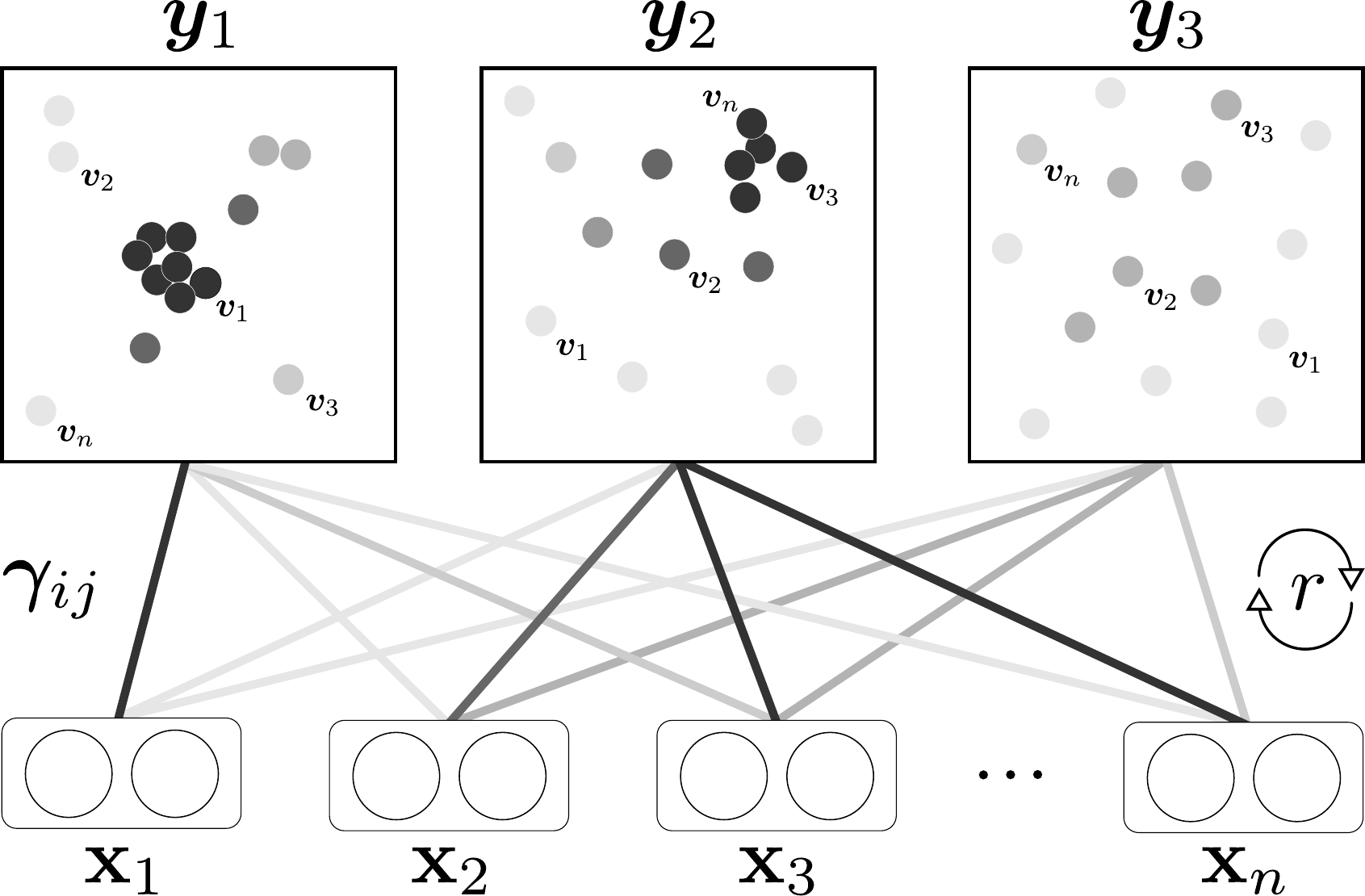}
    \caption{Simple 2D example of capsule routing. Here, all $n$ capsules from the lower layer $\mathbf{x}_i \in \mathbb{R}^2$ vote for each higher capsule $\bm{y}_j \in \mathbb{R}^2$, by: $\bm{v}_{j|i} = \mathbf{W}_{ij} \cdot \mathbf{x}_i$. If the votes for a higher capsule agree and form a cluster, then there is feedback to increase the routing weights $\boldsymbol{\gamma}_{ij}$ between the higher capsule and the lower capsules that form the cluster, whilst decreasing them for other higher capsules.}
    \label{fig:routing}
\end{figure}
\subsection{What is Capsule Routing?}
Capsule routing is a non-linear, iterative and clustering-like process that occurs between adjacent capsule layers. The goal of capsule routing is to dynamically assign \textit{part} capsules $i=1,\dots,N$ in layer $\ell$ to \textit{object} capsules $j=1,\dots,M$ in layer $\ell+1$, by iteratively adjusting the routing coefficients $\boldsymbol{\gamma} \in \mathbb{R}^{N \times M}$, where $0 \leq \gamma_{ij} \leq 1$. These routing coefficients are like an attention matrix which modulates the outputs as a weighted average of the inputs. For a simple example of capsule routing in 2D see Figure~\ref{fig:routing}. 

This type of `routing-by-agreement' is a dynamic alternative to the primitive form of routing implemented by max-pooling, whereby neurons in the upper layer ignore all but the most active feature detector in a local pool in the layer below. As shown in the example Figure~\ref{fig:agreement_faces}, rather than merely detecting whether certain parts/objects are present anywhere in an input image like pooling CNNs, capsule routing aims to detect objects by looking for coherent agreement between the pose of discovered parts. Dynamic routing has been shown to be an effective way to implement the ``explaining away” that is needed for segmenting overlapping objects and generalising to novel viewpoints~\cite{sabour2017dynamic,hinton2018matrix,NEURIPS2020_47fd3c87}.
\section{Capsule Routing Mechanisms}\label{sec:capsule_networks}
Having provided the basic necessary background and motivation behind capsule networks, in this section we focus on providing an in-depth overview of the most prominent capsule routing algorithms proposed in literature. Our exposition (loosely) follows chronological order starting with the seminal works on capsules and ending on recent work. 
%
%
%
\begin{figure}[b!]
    \centering
    \includegraphics[trim={10 10 10 10},clip,width=.8\columnwidth]{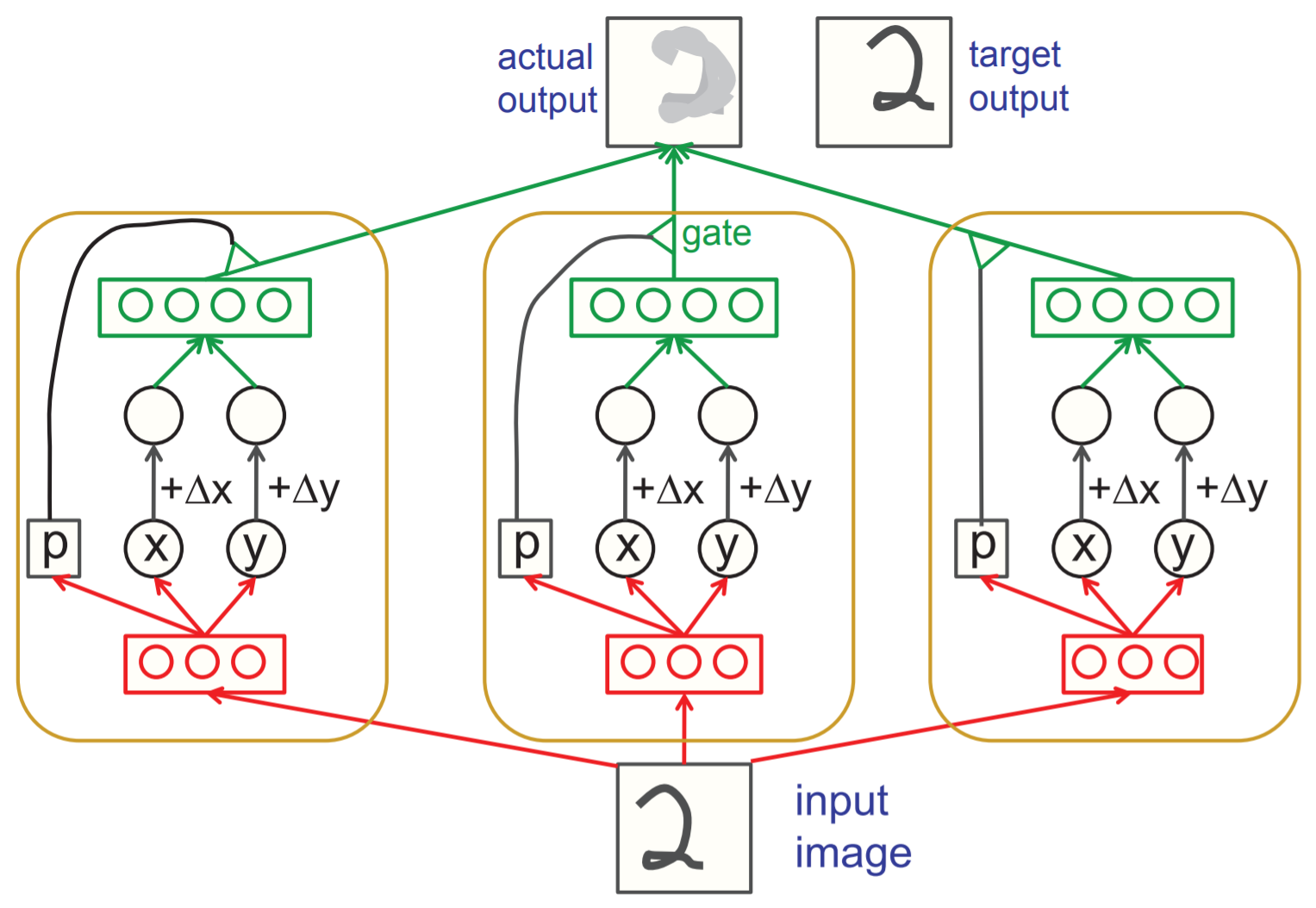}
    \caption{The transforming autoencoder for modelling translations. There are 3 capsules and each one has 3 recognition units (red) and 4 generation units (green). Figure from ~\cite{hinton2011transforming}.}
    \label{fig:transforming_autoencoder}
\end{figure}
\subsection{Transforming Autoencoders}
The idea of using capsules instead of neurons as the building blocks in a neural network was first introduced by~\cite{hinton2011transforming}. 
The authors showed how a neural network can be used to learn features that output a vector of instantiation parameters (capsule), and argued that this is a better way of dealing with tranformations of the input.

As previously mentioned, although several stages of subsampling in a CNN can afford some invariance to pose over a limited range, it ignores precises spatial relationships which are crucial for generalisation~\cite{hinton2011transforming}. The authors propose the transforming autoencoder as a way to learn the first level of capsules, whereby pixel intensities are converted to pose parameters. As shown in the example in Figure~\ref{fig:transforming_autoencoder}, transforming autoencoders can receive the input image and a desired shift, $\Delta x$ and $\Delta y$, and output the shifted input by merging information from generative capsule units. In Figure~\ref{fig:transforming_autoencoder}, $p$ is the probability that the visual entity modelled by a particular capsule is present in the input image. The authors also studied the prediction of more complex 2D transformations and changes in 3D viewpoint by using 3$\times$3 matrix representations of the desired $\Delta$'s, demonstrating the merit of the approach in their preliminary experiments.
\subsection{Dynamic Routing Between Capsules}
\begin{figure}[t]
    \centering
    \includegraphics[width=\columnwidth]{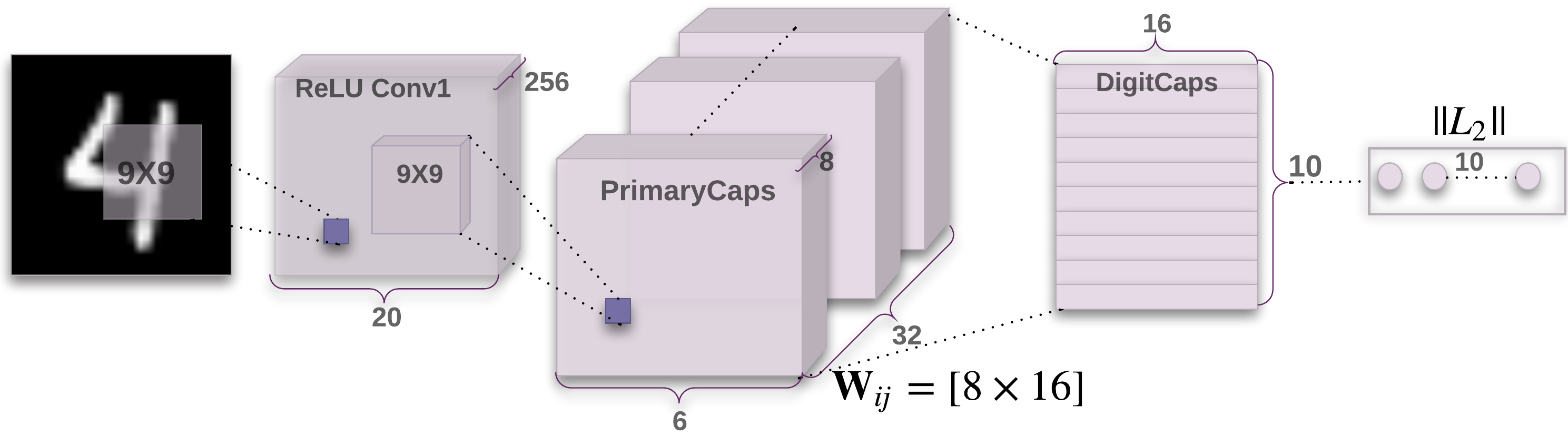}
    \caption{A simple capsule network with 3 layers. The \texttt{PrimaryCaps} are the lowest level of multi-dimensional neuronal activity (each capsule is an 8D vector). The length of the activity vector of each capsule in the \texttt{DigitCaps} layer indicates the presence of each class. Figure from ~\cite{sabour2017dynamic}.
    }
    \label{fig:capsnetArch}
\end{figure}

\begin{figure}[t]
    \centering
    \includegraphics[width=.75\columnwidth]{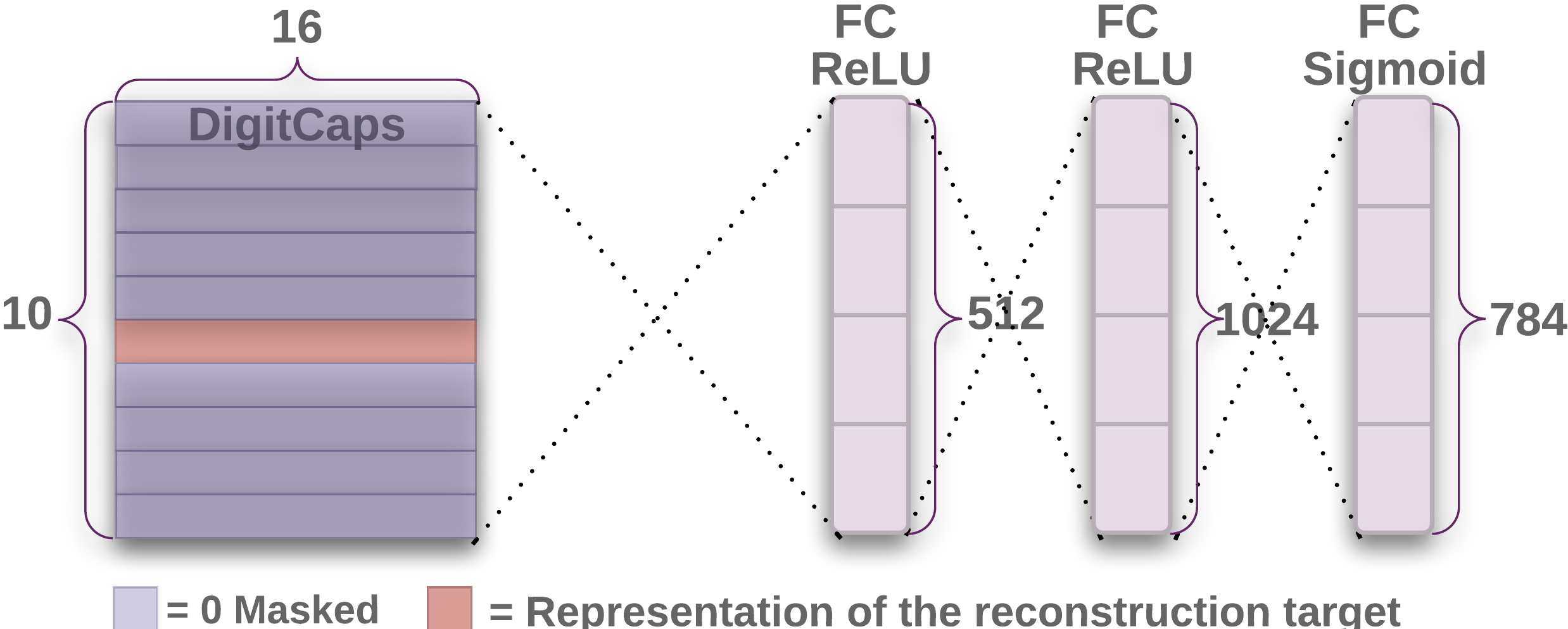}
    \caption{The decoder network used by \cite{sabour2017dynamic} to reconstruct input digits from their capsules in the \texttt{DigitCaps} layer.
    }
    \label{fig:reconsArch}
\end{figure}

The idea of routing-by-agreement was first introduced in the seminal work by~\cite{sabour2017dynamic}, and since then many other variants of capsule routing have been proposed. Their routing process is shown in Algorithm~\ref{routingalg}, whereby vectors $\mathbf{x}_i \in \mathbb{R}^D$ of lower layer capsules are transformed by weights $\mathbf{W}_{ij} \in \mathbb{R}^{D \times P}$ to make predictions for the vectors $\bm{y}_j \in \mathbb{R}^P$ of higher layer capsules. If a lower layer capsule $i$ (e.g. encoding a nose) predicts the properties of a possible parent capsule $j$ (e.g. encoding a face) with high accuracy, there is top-down feedback which increases the affinity (routing coefficient $\gamma_{ij}$) between them. The  proposed capsule network architecture in ~\cite{sabour2017dynamic} is shown in Figure~\ref{fig:capsnetArch}, and the decoder in Figure~\ref{fig:reconsArch} is used to reconstruct the input.

\begin{algorithm}[t]
\setstretch{1.2}
\caption{Dynamic Routing-by-Agreement}\label{routingalg}
\begin{algorithmic}[1]
\Function{Routing}{$\mathbf{x}, \mathbf{W}, r$}
\State {$\forall \ i, j$ capsules in layer $\ell$ and $\ell+1$ : $\gamma_{ij} \gets 0 $}
\State $\forall \ i, j$ : $\bm{v}_{j|i} \gets \mathbf{W}_{ij} \cdot \mathbf{x}_i$ \Comment{voting}
\For{$r$ iterations}
\State $\forall \ i \in \ell$ : $\boldsymbol{\gamma}_i \gets \texttt{softmax}(\boldsymbol{\gamma}_i)$ \Comment{routing weights}
\State $\forall \ j \in \ell+1$ : ${\bf s}_j \gets \sum_i{\gamma_{ij}{\bm{v}}_{j|i}}$
\State $\forall \ j \in \ell+1$ : ${\bm y}_j \gets \texttt{squash}({\bf s}_j)$ 
\Comment{Eq.~\eqref{squash}}
\State $\forall \ i, j$ : $\alpha_{ij} \gets  {\bm v}_{j|i} \cdot {\bm y}_j$ \Comment{agreement}
\State $\forall \ i, j$ : $\gamma_{ij} \gets \gamma_{ij} + \alpha_{ij}$ \Comment{update}
\EndFor
\vspace{-4pt}
\State{\textbf{return} ${\bm y}_j$}
\EndFunction
\end{algorithmic}
\end{algorithm}
\subsubsection{Capsule Vector Activation} This version of capsules uses the length of the output capsule vector to represent the probability that the \textit{entity} encoded by that capsule is present in the input~\cite{sabour2017dynamic}. To that end, the authors proposed the following non-linear ``squashing'' function to activate every $j^{\mathrm{th}}$ capsule:
\begin{align}
    \label{squash}
    &\bm{y}_j = \frac{\norm{\mathbf{s}_j}^2}{1+ \norm{\mathbf{s}_j}^2} \frac{\mathbf{s}_j}{\norm{\mathbf{s}_j}}, & \mathbf{s}_j = \sum_{i} \gamma_{ij} \mathbf{W}_{ij} \cdot \mathbf{x}_i.
\end{align}
The routing coefficients $\gamma_{ij}$ are iteratively updated based on the \textit{agreement} between the output $\bm{y}_j$ of each higher layer capsule $j$, and the prediction (votes) $\bm{v}_{j|i}$ made by each lower layer capsule $i$. 
The agreement is measured by the scalar product: 
\begin{equation}
    \alpha_{ij} = \bm{v}_{j|i} \cdot \bm{y}_j.     
\end{equation}
To gain some intuition, imagine $\bm{y}_j$ and $\bm{v}_{j|i}$ are both unit vectors, i.e. their magnitudes are: $|\bm{y}_j| = 1$ and $|\bm{v}_{j|i}| = 1$. Then, the dot product between them is equal to $\cos{\theta}$, where $\theta$ is the angle between the two vectors. The agreement $\alpha_{ij}$ is added to $\gamma_{ij}$, updating part-object affinities based on how well the vectors $\bm{y}_j$ and $\bm{v}_{j|i}$ point in the same direction. 
\subsubsection{Margin Loss Function}
The objective function presented in~\cite{sabour2017dynamic} for learning capsule network parameters leverages the length of the capsule vectors to represent the probability that a capsule's entity is present in the input. Note that the norm of the last layer capsule vector $||{\bm y}_k||$, representing class $k$ must be (long) close to 1, if and only if (iff) an image belonging to class $k$ is present in the input. With that in mind, the (multiple) margin loss used by~\cite{sabour2017dynamic} is defined as follows:
\begin{align}
\mathcal{L}_{\mathrm{margin}} = \sum_k &T_k \max(0, m^{+} - ||{\bm y}_k||)^2 \\& + \lambda (1-T_k) \max(0, ||{\bm y}_k|| - m^{-})^2,
\label{digit-loss}
\end{align}
where $T_k = 1$ iff a (digit of class) $k$ is present, $m^{+} = 0.9$, $m^{-} = 0.1$ and $\lambda = 0.5$. The $\lambda$ down-weighting of the loss for absent (digit) classes stops the initial learning from shrinking the lengths of the activity vectors of all the digit capsules. The authors opt for this multi-margin loss function over standard cross-entropy (CE) used in CNNs to more easily accommodate for multi-label classification tasks. 
\begin{figure}[t]
    \centering
    \includegraphics[trim={0 0 0 0},clip,width=\columnwidth]{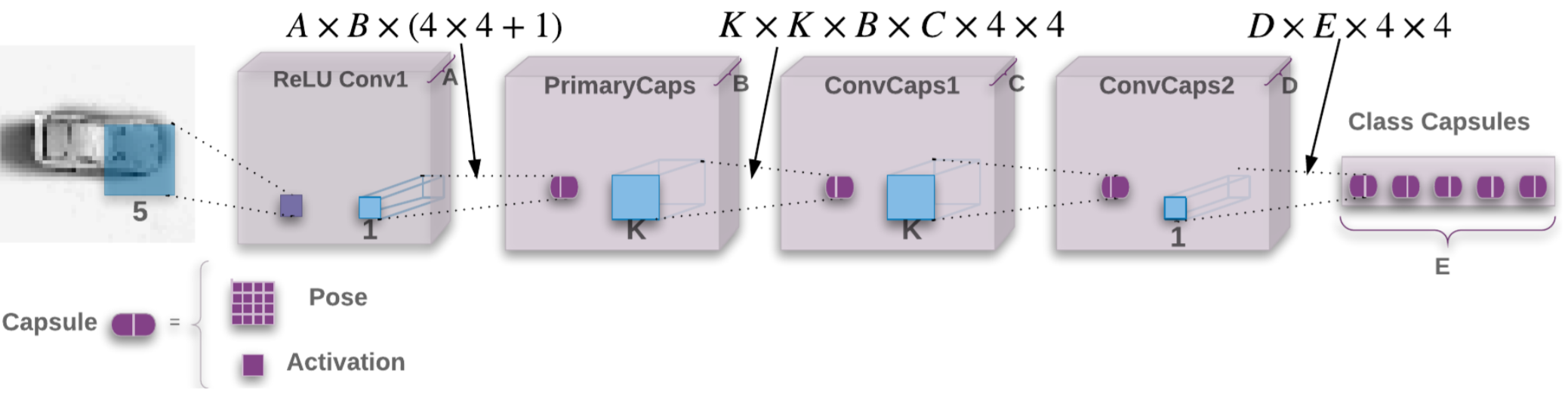}
    \caption{The capsule network architecture with 3 convolutional capsule layers (\texttt{ConvCaps}) used in~\cite{hinton2018matrix}.
    The number of weights in each layer is shown above.}
    \label{fig:em_caps_arch}
\end{figure}
\subsection{Matrix Capsules with EM-Routing}
More recently, a new version of capsules was proposed~\cite{hinton2018matrix}, which overcomes the following deficiencies of~\cite{sabour2017dynamic}:
\begin{enumerate}[label=(\roman*)]
    \item Using the length of the pose vector to represent the probability that an entity is present, requires an unprincipled non-linearity (``squashing") that prevents use of typical objective functions~\cite{hinton2018matrix}. Instead, they propose to separate probability of existence from the pose vector.
    \item Using the cosine of the angle between vectors to measure agreement makes the system insensitive to small differences between good and very good agreements. 
    \item Using a vector of length $D$ to represent poses (unnecessarily) increases the number of transformation weights. They propose to use a matrix with $D$ elements instead, which reduces parameters from $D^2$ to $D$.
\end{enumerate}
Rather than using vector capsules $\mathbf{x}_i \in \mathbb{R}^{D}$ as before, the authors use $\mathbf{M}_i \in \mathbb{R}^{\sqrt{D} \times \sqrt{D}}$ capsule pose matrices, and a separate activation probability $a \in \mathbb{R}$ to represent presence of the entity modelled by each capsule~\cite{hinton2018matrix}. They also presented a new capsule network architecture (see Figure~\ref{fig:em_caps_arch}) featuring convolutional capsules, and proposed an alternative routing-by-agreement procedure based on the Expectation-Maximisation (EM) algorithm~\cite{dempster1977maximum}. 

\begin{figure*}[!t]
    \centering
    \includegraphics[trim={0 0 0 0},clip,width=\textwidth]{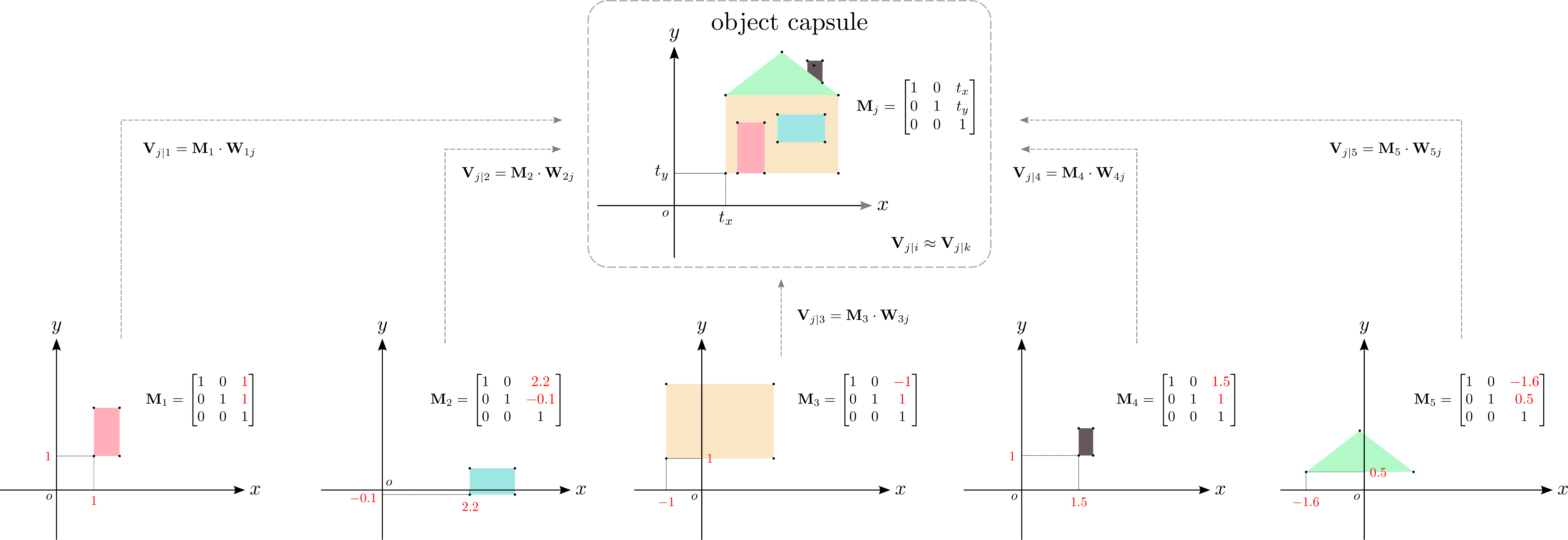}
    \caption{How capsules deal with viewpoint changes (2D translation example). An object can be detected by looking for agreement between votes $\mathbf{V}_{j|i}$ for its pose matrix $\mathbf{M}_j$.
    Lower level (part) capsules produce a vote by multiplying their pose matrix $\mathbf{M}_i$ by a learned viewpoint-invariant transformation matrix $\mathbf{W}_{ij}$. Agreement between multiple parts means that their votes are similar: $\mathbf{V}_{j|i} \approx \mathbf{V}_{j|k}$, for $i \neq k$, which is unlikely to occur by chance in high dimensions~\cite{sabour2017dynamic,hinton2018matrix}. As the viewpoint changes, all the pose matrices change in a coordinated manner, so that any vote agreement will persist~\cite{hinton2018matrix}.}
    \label{fig:house}
\end{figure*}
\subsubsection{Matrix Capsule Voting}
In \cite{hinton2018matrix}, the voting procedure from capsules $i$ in a lower layer $\ell_i$ for the pose matrices of capsules $j$ in a higher layer $\ell_j$ is:
\begin{align}
\label{el_votes}
&\mathbf{V}_{j|i} = \mathbf{M}_{i}\cdot\mathbf{W}_{ij}, &\mathbf{W}_{ij} \in \mathbb{R}^{4\times4}.
\end{align}
where $\mathbf{V}_{j|i}$ denotes the vote from the $i^{\mathrm{th}}$ part capsule for the $j^{\mathrm{th}}$ object capsule, and $\mathbf{W}_{ij}$ is the transformation weight matrix. Since both the pose matrices and the transformations weights are 4$\times$4, each capsule's vote is $\mathbf{V}_{j|i} \in \mathbb{R}^{4\times4}$. See Figure~\ref{fig:house} for a simple 2D translation example.

\textbf{Geometric Interpretation.} We can motivate the use of 4$\times$4 matrices through a geometric interpretation: i.e. 4$\times$4 transformations matrices are commonly used in 3D computer graphics under homogeneous coordinates for perspective projection~\cite{andrew2001multiple}. A 4$\times$4 matrix can represent the following transformations among others: translation, rotation, reflection, glides, scale, contraction, expansion, shear, dilation etc. Assuming a capsule network is able to extract sensible entities from the input, 4$\times$4 pose and transformation weight matrices are theoretically sufficient.

It is worth noting the distinction between the voting procedures in eq.~\eqref{squash} and eq.~\eqref{el_votes}: i.e in the former the vote is calculated via the matrix-vector product: $\bm{v}_{j|i} = \mathbf{W}_{ij}\mathbf{x}_i$, whereas in the latter we have a matrix-matrix product: $\mathbf{V}_{j|i} = \mathbf{M}_i \mathbf{W}_{ij}$. Once again we can provide a geometric interpretation that justifies the order of the product, since due to the non-commutativity of square matrices: $\mathbf{M}_i \mathbf{W}_{ij} \neq \mathbf{W}_{ij} \mathbf{M}_i$. In geometric terms, $\mathbf{M}_i \mathbf{W}_{ij}$ applies a transformation on the pose matrix $\mathbf{M}_i$ defined by matrix $\mathbf{W}_{ij}$, whereas $\mathbf{W}_{ij} \mathbf{M}_{i}$ would wrongly imply that $\mathbf{W}_{ij}$ is the pose matrix and $\mathbf{M}_i$ is the transformation matrix. 


%
\begin{algorithm}[!t]
\setstretch{1.3}
\caption{Expectation-Maximisation Routing}\label{em_routing_algo}
\begin{algorithmic}[1]
\Function{EM-Routing}{$\bm{a}, \mathbf{M}, \mathbf{W}, r$}
\State {$\forall \ i, j$ capsules in layer $\ell$ and $\ell+1$: $\gamma_{ij} \gets M^{-1} $}
\State $\forall \ i, j$ : $\mathbf{V}_{j|i} \gets \mathbf{M}_{i} \cdot \mathbf{W}_{ij}$ \Comment{voting}
\For{$r$ iterations}
\State{$\forall \ i \in \ell$ : $\gamma_{ij} \gets \gamma_{ij} \odot \bm{a}_i$}\Comment{routing weights}
\State $\forall \ j \in \ell+1$ : $\boldsymbol{\mu}_j, \boldsymbol{\sigma}_j \gets$ \Call{M-step}{$\boldsymbol{\gamma}, \mathbf{V}$}
\State{\quad $\mathrm{cost}^h \gets (\beta_u + \log\boldsymbol{\sigma}^h_j) \sum_i \gamma_{ij}$}
\State{\quad $a_j' \gets \mathrm{sigmoid} (\lambda (\beta_a - \sum_h\mathrm{cost}^h))$}\Comment{activation}\label{mdl_activation}
\State $\forall \ i \in \ell$ : $\boldsymbol{\gamma}_i \gets$ \Call{E-step}{$\boldsymbol{\mu}, \boldsymbol{\sigma}, \bm{a}', \mathbf{V}$}\Comment{update}
\EndFor
\vspace{-4pt}
\State{\textbf{return} $\bm{a}, \mathbf{M}$}
\EndFunction
\end{algorithmic}
\end{algorithm}
\subsubsection{Convolutional Capsules}
In \cite{hinton2018matrix}, the authors also introduced the idea of convolutional capsules, whereby the connectivity between capsules in adjacent layers, follows that of a CNN. That is, rather than performing a regular convolution (sharing scalar feature detector kernels across the input), convolutional capsule layers share transformation weight matrices $\mathbf{W}_{ij}$ spatially across input capsules. Multiple convolutional capsule layers are then stacked to build a capsule network (see Figure~\ref{fig:em_caps_arch}). This extension makes intuitive sense as ideally we would like to retain the ability to generalise knowledge across all spatial locations in the image like CNNs, whilst replacing pooling operations in favour of routing-by-agreement.

\subsubsection{Expectation-Maximisation Routing}
Similar to~\cite{sabour2017dynamic}, a non-linear procedure to route between adjacent capsule layers they call EM-Routing (see Algorithm~\ref{em_routing_algo}) is proposed in \cite{hinton2018matrix}. Concretely, the procedure is a version of the Expectation-Maximisation (EM) algorithm~\cite{dempster1977maximum}, that iteratively adjusts the means, variances and activation probalitites of the capsules in layer $\ell+1$, and the assignment probabilities between all capsules (routing weights $\boldsymbol{\gamma}$). 

Unlike Dynamic routing~\cite{sabour2017dynamic}, EM-Routing fits Gaussian distributions on the votes coming from part capsules $i$ to object capsule $j$ poses: $\mathcal{N}(\mathbf{M}_{j} \ | \  \boldsymbol{\mu}_{j},\boldsymbol{\sigma}^2_{j}) \ \forall j$, where each capsule $j$ (of $N_j$ total object capsules in layer $\ell+1$) has a diagonal covariance matrix with $h$ components: $\boldsymbol{\sigma}^2_{j} \in \mathbb{R}^h$. Following Algorithm~\ref{em_routing_algo} closely, EM-Routing iterates between updating the means $\boldsymbol{\mu}_j$, variances $\boldsymbol{\sigma}^2_j$ and activations $\bm{a}_j$ of capsules $j$ whilst holding the routing coefficients $\gamma_{ij}$ fixed (M-step), and updating $\gamma_{ij}$ holding $\boldsymbol{\mu}_j$, $\boldsymbol{\sigma}^2_j$ and $\bm{a}_j$ fixed (E-step). For more detailed explanation please refer to~\cite{hinton2018matrix}.

\subsubsection{Capsule Activation}
As shown in Algorithm~\ref{em_routing_algo}, vote agreement is measured using the variance $\boldsymbol{\sigma}_j$ of each object capsule's Gaussian, which is then weighted by its \textit{support}: $\sum_i \gamma_{ij}$, i.e. amount of part capsules $i$ assigned to object capsule $j$. To set the activation probability $a_j$ for a particular object capsule $j$,~\cite{hinton2018matrix} compare the description lengths (energies $-\beta_u$ and $-\beta_a$ that are learned discriminatively) of two different ways of coding the poses of the activated capsules $i$ assigned to $j$ by the routing procedure. The difference in the two energies is put through a logistic function to determine the activation probability of each object capsule $j$, noting that the logistic function computes the distribution: $\mathrm{Bernoulli}(p)$, that minimises free-energy when the difference in the energies is its argument (see line~\ref{mdl_activation} in Algorithm~\ref{em_routing_algo}). 
\begin{figure}[!t]
    \centering
    \includegraphics[width=\columnwidth]{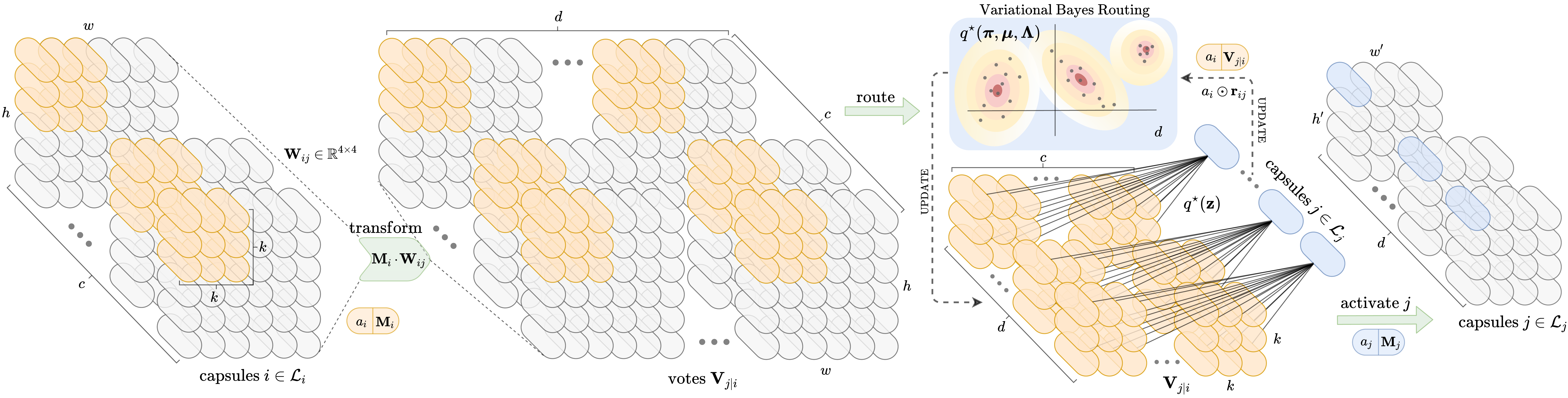}
 \caption{Depiction of VB-Routing between convolutional capsule layers. Each capsule has an activation  $a$ and a pose matrix $\mathbf{M}\in\mathbb{R}^{4\times4}$. Parent capsules $j \in \mathcal{L}_j$ (blue) only receive votes from child capsules $i \in \mathcal{L}_i$ (orange) within their receptive field. $c$ and $d$ denote the number of child and parent capsule types (channels) respectively. Figure from ~\cite{ribeiro2020capsule}.
 }
\label{fig:vb_routing}
\end{figure} 
\subsubsection{Spread Loss function}
The ``spread'' loss function used in~\cite{hinton2018matrix} directly maximises the gap between the activation of the (final layer) capsule representing the target class $a_t$, and the other class capsules:
\begin{align}
    &\mathcal{L}_{\mathrm{spread}} = \sum_{i\neq t} \mathcal{L}_i, &\mathcal{L}_i = \max(0, m - (a_t - a_i))^2, 
\end{align}
where the margin $m$ is linearly increased during training from 0.2 to 0.9, avoiding dead capsules in the earlier layers. It is reported in~\cite{hinton2018matrix}  that EM-Routing matrix capsules outperform~\cite{sabour2017dynamic}, and significantly outperform comparable size CNNs on viewpoint-invariance and adversarial robustness tasks.
\begin{algorithm}[t]
\setstretch{1.3}
\small
\caption{Variational Bayes Routing}\label{vb_routing_algo}
\begin{algorithmic}[1]
\Function{VB-Routing}{$\bm{a}, \mathbf{M}, \mathbf{W}, r$}
\State {$\forall \ i, j$ capsules in layer $\mathcal{L}_i$ and $\mathcal{L}_j$: $\gamma_{ij} \gets N_j^{-1} $}
\State $\forall \ i, j$ : $\mathbf{V}_{j|i} \gets \mathbf{M}_{i} \cdot \mathbf{W}_{ij}$ \Comment{voting}
\State{$\forall \ j \in \mathcal{L}_j$ : $\alpha_0, \mathbf{m}_0, \kappa_0, \boldsymbol{\Psi}_0, \nu_0$} \Comment{initialise priors}
\For{$r$ iterations}
\State{$\forall \ i \in \mathcal{L}_i$ : $\gamma_{ij} \gets \gamma_{ij} \odot \bm{a}_i$}\Comment{routing weights}
\State $\forall \ j \in \mathcal{L}_j$ : 
\Call{Update $q^\star(\boldsymbol{\pi}, \boldsymbol{\mu}, \boldsymbol{\Lambda})$}{}
\State $\forall \ i \in \mathcal{L}_i$ : 
\Call{Update $q^\star(\bm{z})$}{}
\EndFor
\State{$\forall \ j \in \mathcal{L}_j$ : $a'_j = \sigma(\beta_a - \big(\beta_u + \mathbb{E}[\ln \boldsymbol{\pi}_j] + \mathbb{E}[\ln \mathrm{det}(\boldsymbol{\Lambda}_j)]))$}
\State{\textbf{return} $\bm{a}', \mathbf{M}$}
\EndFunction
\end{algorithmic}
\end{algorithm}
\subsection{Capsule Routing via Variational Bayes}
In~\cite{ribeiro2020capsule}, the authors propose Variational Bayes (VB) routing as a way to address some of the inherent drawbacks of EM-Routing~\cite{hinton2018matrix} encountered by various other authors such as training instability and reproducibility~\cite{gritzman2019avoiding,NIPS2019_8982,Tsai2020Capsules}. In EM, variance-collapse~\cite{murphy2012machine} singularities occur when an object capsule (Gaussian cluster) claims sole custody of a part capsule (datapoint), yielding infinite likelihood and zero variance. 
To address this, Bayesian learning is brought to capsule networks by placing priors and modelling uncertainty over capsule parameters between capsule layers~\cite{ribeiro2020capsule}. The advantages of VB-Routing include: (i) Flexible control over capsule complexity by tuning priors; (ii) Number of effective object capsules is determined automatically; (iii) Known pathological solutions of the EM algorithm (variance-collapse) are addressed in a principled manner.
\subsubsection{Variational Bayes Routing}
See Figure~\ref{fig:vb_routing} for an illustration of VB-Routing. The authors place conjugate priors over $\boldsymbol{\mu}$ and $\boldsymbol{\Lambda}$, which are the mean and inverse covariance (precision) matrix
of each object capsule's Gaussian distribution, and over $\boldsymbol{\pi}$ which are the mixing coefficients of the mixture model. The latent variables $\bm{z} = \{\bm{z}_i,\dots,\bm{z}_{|\mathcal{L}_i|}\}$ are a set of one-hot vectors describing the cluster assignments of each of the lower capsules' votes $\mathbf{V}_{j|i}$ to higher capsules' Gaussian distributions. Variational inference (VI)~\cite{hinton1993keeping,jordan1999introduction} of the above latent variables is performed~\cite{ribeiro2020capsule} -- analogously to Bayesian Gaussian mixture models~\cite{10.5555/2073796.2073799,bishop2006pattern} -- between all adjacent capsule layers. Unlike standard mixture models, here every cluster (object capsule) has its own learnable matrix $\mathbf{W}_{ij}$, with which its datapoints (votes) are transformed, so every cluster sees a different view of the data~\cite{hinton2018matrix,ribeiro2020capsule}.

The \textit{generative story} for each part capsule $i$ is that of a mixture model with priors over object capsule $j$ parameters $\boldsymbol{\pi}, \boldsymbol{\mu}, \boldsymbol{\Lambda}$. 
%
%
The joint distribution of the model factorises as:
$p(\mathbf{V},\bm{z}, \boldsymbol{\pi}, \boldsymbol{\mu}, \boldsymbol{\Lambda}) =  p(\mathbf{V}|\bm{z},\boldsymbol{\mu}, \boldsymbol{\Lambda}) p(\bm{z}|\boldsymbol{\pi}) p(\boldsymbol{\pi}) p(\boldsymbol{\mu}|\boldsymbol{\Lambda}) p(\boldsymbol{\Lambda})$, and the posterior is approximated with factorised variational distribution over all the latent variables:
\begin{align}
    \underbrace{p(\bm{z},\boldsymbol{\pi},\boldsymbol{\mu},\boldsymbol{\Lambda}|\mathbf{V})}_{\mathrm{posterior}} \approx q(\bm{z})q(\boldsymbol{\pi}) \prod_{j \in \mathcal{L}_j} q(\boldsymbol{\mu}_j,\boldsymbol{\Lambda}_j).
\end{align}

\textbf{Coordinate Ascent Updates.} \ To perform routing between adjacent capsule layers, the authors~\cite{ribeiro2020capsule} iteratively optimise parent capsule parameter distributions: 
$q(\boldsymbol{\pi},\boldsymbol{\mu},\boldsymbol{\Lambda})$, with the responsibilities $\bm{z}$ over child capsules fixed, and re-evaluate the new expected responsibilities $q^{\star}(\bm{z})$ with the distributions over parent capsule parameters fixed.
%
%
This approach leads to an EM-like algorithm with variational EM updates as outlined in Algorithm~\ref{vb_routing_algo}. For further details on the standard closed-form update equations of Bayesian mixture models refer to~\cite{ribeiro2020capsule,10.5555/2073796.2073799,bishop2006pattern}.
\begin{figure}[t]
    \centering
    \includegraphics[trim={0 0 0 0},clip,width=.968\columnwidth]{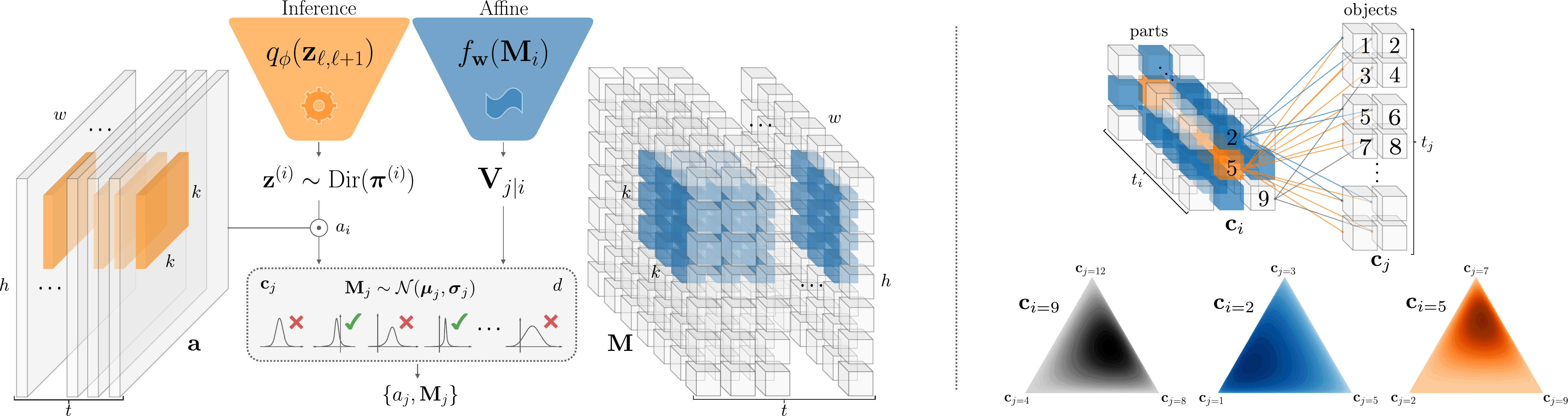}
 \caption{The inference procedure in a capsule layer~\cite{NIPS2019_8982}. (Left) Feature maps of part capsule activations $\mathbf{a}$ and pose matrices $\mathbf{M}$ give rise to votes $\mathbf{V}_{j|i}$, and part-object connections $\mathbf{z}$ which are sampled from Dirichlet distributions during training. (Right) Example of part-object connectivity in convolutional capsule voting for kernel size $k=2$. Figure from ~\cite{NIPS2019_8982}.
 }
 \label{fig:main_diagram}
\end{figure}
\subsubsection{Agreement \& Capsule Activation}
To measure agreement,~\cite{ribeiro2020capsule} propose to use the differential entropy of higher capsule $j$'s Gaussian-Wishart variational posterior distribution $q^\star(\boldsymbol{\mu}_j, \boldsymbol{\Lambda}_j)$. In simple terms, if the entropy of a capsule $j$'s Gaussian is low that means the votes received from capsules $i$ agree (i.e. form a tight cluster), and vice-versa. In practice, the authors approximate the entropy up to constant factors with:  
\begin{align}
    \mathbb{H}[q^{\star}(\boldsymbol{\mu}_j, \boldsymbol{\Lambda}_j)]
    \approx \mathbb{E}[\ln \mathrm{det}(\boldsymbol{\Lambda}_j)].
\end{align}
As outlined in Algorithm~\ref{vb_routing_algo}, this agreement measure is then weighted by the amount of \textit{support} for each object capsule (mixing coefficient) and activated
via the logistic function $\sigma(\cdot)$, where $\beta_a, \beta_u$ are (optional) parameters learned discriminatively, similar to EM-Routing~\cite{hinton2018matrix}. Unlike EM~\cite{hinton2018matrix} or Dynamic routing~\cite{sabour2017dynamic} however, capsules are only activated after the routing iterations.
\begin{algorithm}[t!]
\setstretch{1.3}
\small
\caption[Routing uncertainty algorithm.]{Capsule Layer with Routing Uncertainty 
}\label{uncertainty_algo}
\begin{algorithmic}[1]
\Function{ConvCaps2D }{$\mathbf{a}$, $\mathbf{M}, \mathbf{W}$} 
\State $\forall \ i, j$ capsules in $\ell,\ell+1$ : $\mathbf{V}_{j|i} \gets \mathbf{M}_{i} \cdot \mathbf{W}_{ij}$ \Comment{voting}
\State {$\forall \ i \in \ell $ : $\boldsymbol{\pi}_0^{(i)} \in \mathbb{R}^{N_{i\rightarrow j}}$}{} \Comment{set Dirichlet priors}
\State {$\forall \ i \in \ell $ : $\mathbf{z}^{(i)} \sim q_\phi(\mathbf{z}_{\ell,\ell+1})$} 
\Comment{posterior sample}
\State $\forall \ j \in \ell+1$ : $\boldsymbol{\mu}_j, \boldsymbol{\sigma}_j \gets$ \Call{Route }{$\mathbf{z}_{\ell,\ell+1}$, $\mathbf{V}_{j|i}$}
\State $\forall \ j \in \ell+1$ : $a'_j = \sigma \big(
    -\boldsymbol{\eta}_j\mathbb{E}(S_j)^{-1}\mathcal{H}(\mathbf{M}_{j})\big)$ \Comment{activation}
\State {\textbf{return} $\mathbf{a}', \mathbf{M}$}{} 
\EndFunction
\end{algorithmic}
\end{algorithm}
\subsection{Uncertainty in Capsule Routing}
Sources of uncertainty in assembling objects via a composition of parts can arise from: (i) feature occlusions due to observed viewpoints; (ii) sensory noise in captured data; (iii) object symmetries for which
poses may be ambiguous such as spherical objects and/or parts.  Recently,~\cite{NEURIPS2020_47fd3c87} proposed a \textit{global} (locally non-iterative) view of capsule routing based on representing the inherent \textit{uncertainty} in part-object relationships, by approximating a posterior distribution over part-object connections.  

In simple terms, the \textit{local} routing iterations are replaced with Variational Inference (VI) of part-object connections in a \textit{probabilistic} capsule network, leading to increased efficiency and improved performance on pose-aware benchmarks. In this way, they encourage global context to be taken into account when routing information, by introducing \textit{global} latent variables which have direct influence on their end-to-end variational free-energy objective~\cite{NEURIPS2020_47fd3c87}.
\subsubsection{Inference \& Model Assumptions}
To represent uncertainty about part-object relationships in a capsule network, the posterior distribution $p(\mathbf{z}|\mathcal{D},\mathbf{W})$
over part-object latent connections $\mathbf{z}$ given the data $\mathcal{D}$ is needed. However, exact inference is intractable for complex models~\cite{NEURIPS2020_47fd3c87}. To circumvent this, the authors~\cite{NEURIPS2020_47fd3c87} use stochastic VI tools to find the best approximation $q_{\phi}^{\star}(\mathbf{z})$ that minimises $D_{\mathrm{KL}}(q_{\phi}(\mathbf{z}) \ || \ p(\mathbf{z}|\mathcal{D},\mathbf{W}))$, where $\mathbf{z}$ are global latent part-object connection variables, and $\mathbf{W}$ are viewpoint-invariant transformation parameters, in a CapsNet with $L$ layers.

\textbf{Priors.} \ The authors~\cite{NEURIPS2020_47fd3c87} place a prior distribution $p(\mathbf{z}^{(i)})$ over each part capsule's $\mathbf{c}_i \in \ell$ connections to the object capsules they vote for $\mathbf{c}_j \in \ell+1$, and assume fully factorised independence across layers:
\begin{equation}
    p(\mathbf{z}) = \prod_{\ell=1}^{L-1}\prod_{i=1}^{N_i}  p(\mathbf{z}^{(i)}_{\ell}),
\end{equation}
the connections vector $\mathbf{z}^{(i)} \in \mathbb{R}^{N_{i \rightarrow j}}$, of each $\mathbf{c}_i$ is:
\begin{align}
    &\mathbf{z}^{(i)} = (z_{1},\dots, z_{N_{i\rightarrow{j}}}) \sim  p(\mathbf{z}^{(i)}), &\forall \mathbf{c}_i \in \ell,
\end{align}
where $N_{i \rightarrow j}$ denotes the number of object capsules $j$ that each part capsule $i$ votes for in a particular layer (see Figure~\ref{fig:main_diagram} for convolutional capsule voting example). Different choices of prior are considered, and the authors opt for Dirichlet priors due to a reduced parameter count. A mean-field variational approximation $q_\phi(\mathbf{z}_{\ell,\ell+1})$ to the (intractable) posterior on part-object connection variables is made between all adjacent capsule layers.
%
%
The model is defined hierarchically where the object capsules in layer $\ell$ are the part capsules of $\ell +1$ and so on.

\textbf{Free Energy Objective.} \ The model is fit end-to-end by maximising a lower bound on the conditional marginal log likelihood $\log p(\mathbf{y}|\mathbf{x})$.
%
%
In practice, VI of latent variables $\mathbf{z}$, and maximum a posteriori (MAP) inference of $\mathbf{W}$ is performed~\cite{NEURIPS2020_47fd3c87}.
\begin{figure}[!t]
    \centering
    \includegraphics[trim={0 0 0 0},clip,width=1\columnwidth]{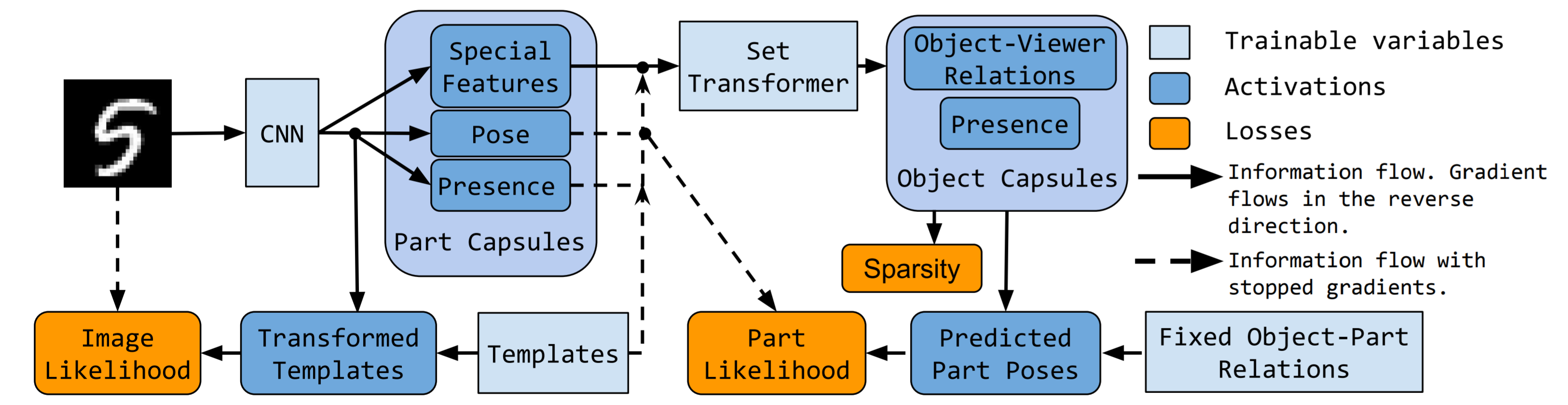}
 \caption{Stacked Capsule Autoencoder architecture. Figure from ~\cite{kosiorek2019stacked}.
 }
 \label{fig:scae}
\end{figure} 
\subsubsection{Routing \& Activating Capsules}
As in previous work~\cite{hinton2018matrix,ribeiro2020capsule}, matrix capsules $\mathbf{M} \in \mathbb{R}^{4\times4}$, and convolutional capsule voting are used (see Figure~\ref{fig:main_diagram}). During training, the authors~\cite{NEURIPS2020_47fd3c87} fit multivariate Gaussians $\mathbf{M}_j \sim \mathcal{N}(\boldsymbol{\mu}_j, \boldsymbol{\sigma}_j)$, on each object capsule's $D{=}16$ dimensional poses, and randomly sample part-object connections from the approximate posterior at each capsule layer $\ell$:
\begin{align}
    &\mathbf{z}^{(i)} = (z_{1},\dots, z_{N_{i\rightarrow{j}}}) \sim  q_{\phi}(\mathbf{z}_{\ell,\ell+1}), &\forall \mathbf{c}_i \in \ell,
\end{align}
then calculate the parameters of each capsule's Gaussian.
%
%
The procedure (see Algorithm~\ref{uncertainty_algo}) can be interpreted as \textit{global} (locally non-iterative) routing, since the posterior: $q_{\phi}^{\star}(\mathbf{z}|\mathcal{D}) \approx p(\mathbf{z}|\mathcal{D},\mathbf{W})$ is inferred for all layers at once, rather than performing \textit{local} (iterative) inference of $\mathbf{z}$ in the E-step of EM-Routing~\cite{hinton2018matrix} between each each pair of adjacent capsule layers.

\textbf{Agreement \& Activation.} \ To activate capsules,~\cite{NEURIPS2020_47fd3c87} follow the general concept of~\cite{ribeiro2020capsule}, and measure vote agreement via the average negative entropy of each capsule's Gaussian:
\begin{align}
    -\mathcal{H}\big[\mathcal{N}(\mathbf{M}_j \ | \ \boldsymbol{\mu}_{j},\boldsymbol{\sigma}_{j})\big] &\triangleq  -\frac{1}{D}\sum_{i=1}^D \log \left(\boldsymbol{\sigma}_j^{(i)}\sqrt{2 \pi e} \right) 
\end{align}
The \textit{agreement} is weighted by the (normalised) \textit{support} (\# parts assigned to an object) for each capsule, and activated using the logistic function $\sigma(\cdot)$. As shown in Algorithm~\ref{uncertainty_algo}, $\mathbb{E}(S_j)$ is the average support each object capsule receives in a given layer, $S_j \sim \mathrm{Binomial}(N_i, {N_j}^{-1})$.
%
%
\subsubsection{Uncertainty Quantification}
With their method, the authors~\cite{NEURIPS2020_47fd3c87} unlock \textit{uncertainty} representation in capsule networks. To that end, they draw $T$ Monte Carlo samples of part-object connections $\mathbf{z}$ from the approximate posterior, and calculate the predictive entropy $\mathcal{H}(\widehat{\mathbf{y}}|\mathbf{x},\mathbf{z},\mathbf{W})$ of the model's output distribution with sampled $\mathbf{z}^t \sim q^{\star}_{\boldsymbol{\phi}}(\mathbf{z}|\mathcal{D})$. Under full posterior learning: $q_\phi(\mathbf{z},\mathbf{W})$, the pose transformation matrices $\mathbf{W}$ are also randomly sampled.
%
%

%
\begin{figure}[t]
    \centering
    \includegraphics[trim={0 0 0 0},clip,width=1\columnwidth]{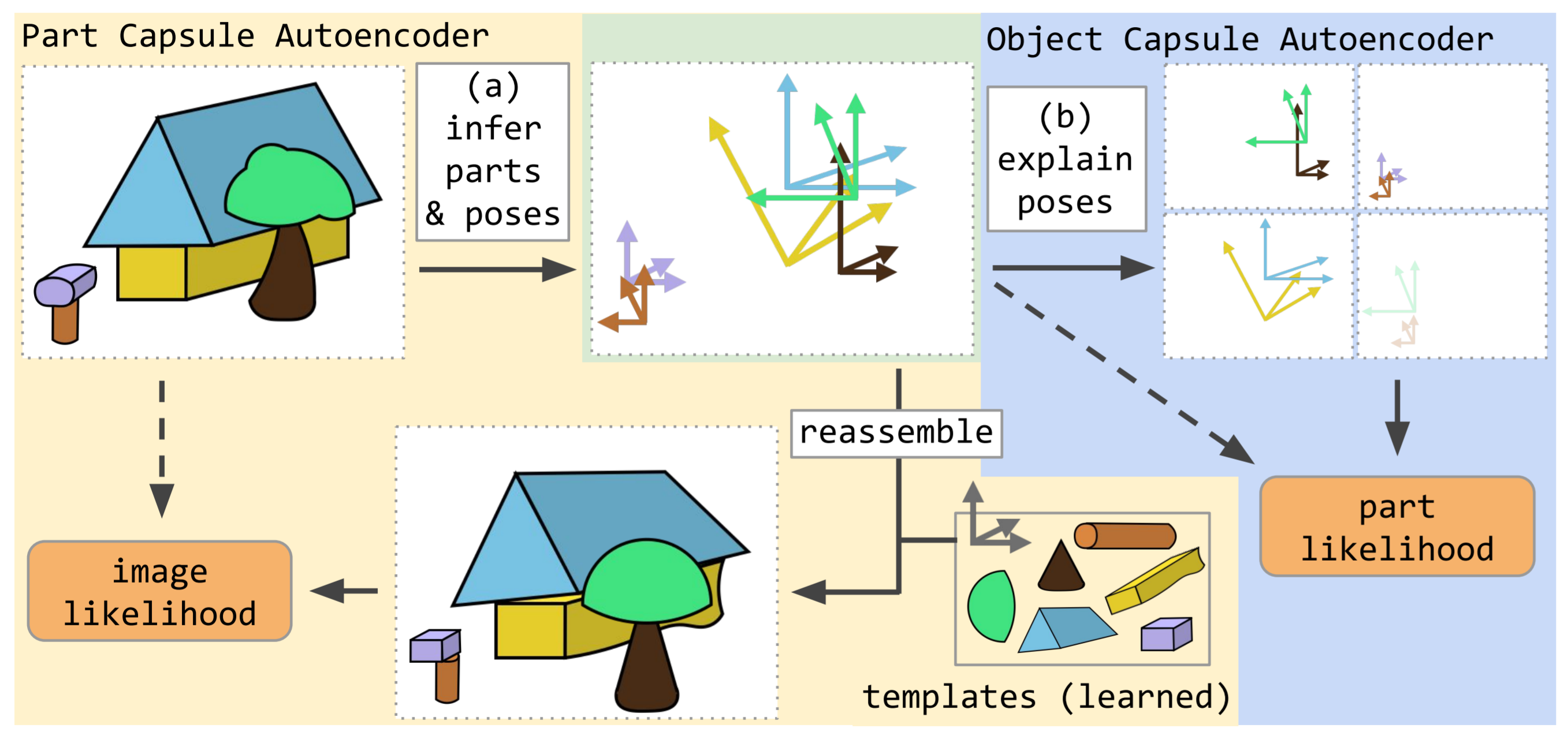}
 \caption{A schematic showing the two components of SCAE: a) part capsule autoencoder segment the input images into individual parts and their corresponding poses. The extracted poses are used to reconstruct the input by affine-transforming learned templates; and b) the object capsule autoencoder use the inferred poses and try to rearrange them so that they better learn the underlying structure of the objects. Finally the SCAE is trained by maximizing image and part log-likelihoods subject to sparsity constraints. Figure from ~\cite{kosiorek2019stacked}.}
  \label{fig:scae_sch}
\end{figure} 
\subsection{Stacked Capsule Autoencoders} \label{sec:scae}
Kosiorek et al. ~\cite{kosiorek2019stacked} introduced the Stacked Capsule Autoencoder (SCAE), which is a seminal work on formulating CapsNets as an unsupervised capsule autoencoder which explicitly uses geometric
relationships to reason about objects. This work combines ideas from Transforming Autoencoders~\cite{hinton2011transforming} and EM routing capsules~\cite{hinton2018matrix} -- but unlike previous methods -- inference in this model is amortized and performed by off-the-shelf neural encoders. The authors also used discovered objects to predict parts rather that using parts to predict objects as in previous capsule networks. Even though the training objective used in SCAE is not concerned with classification or clustering, it is the only method that achieves competitive results in unsupervised object classification without relying on mutual information. The proposed SCAE architecture can be seen in Figure~\ref{fig:scae}, and it consists of two main parts: i) the Part Capsule Autoencoder (PCAE); ii) the Object Capsule Autoencoder (OCAE). A more detailed schematic of what each of these parts does can be seen in Figure~\ref{fig:scae_sch}.

In addition to the two stages mentioned above that are presented in the paper as being the main parts, there is also an earlier step that deals with abstracting away pixels and the part-discovery stage, called Constellation Capsule Autoencoder (CCAE). CCAE uses two-dimensional points as parts, whose coordinates are given as input to the system. CCAE then learns to model the sets of points as arrangements of familiar constellations, each of which has been transformed by an independent similarity transform.
 
\textbf{Part Capsule Autoencoder (PCAE).} \ Although CCAE considers a part as a 2D point (x and y coordinates), for PCAE each part capsule $m$ has a six-dimensional pose $x_m$ (two rotations, two translations, scale and shear), a presence variance $d_m \in [0, 1]$, and a unique identity. Discovering part is formulated as an auto-encoding exercise: the \textit{encoder} learns to infer the poses and presences of different part capsules, while the decoder learns an image template $T_m$ for each part. In the case where a part exists, the corresponding template is affine-transformed with the inferred pose giving $\hat{T}_m$. Finally, all the transformed templates are arranged into the image.

\begin{figure}[!t]
    \centering
    \includegraphics[trim={0 0 0 0},clip,width=.95\columnwidth]{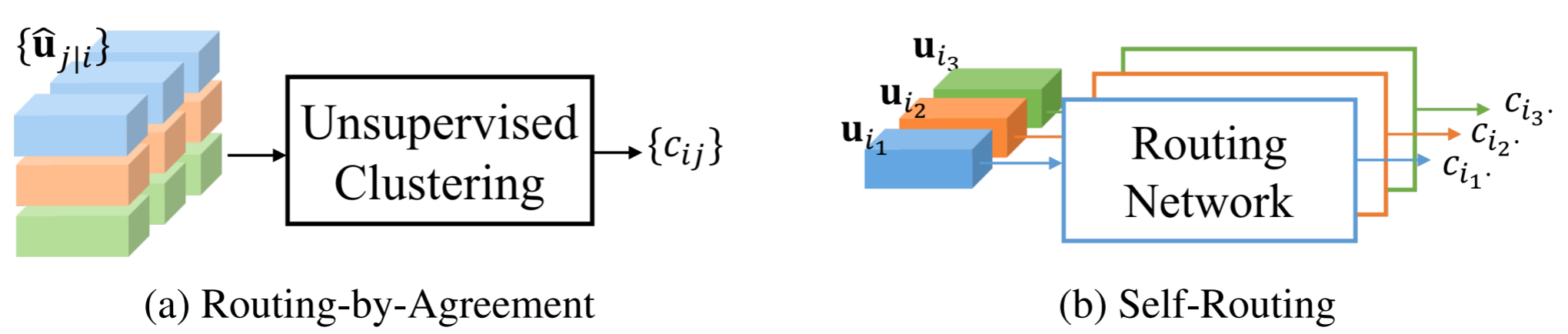}
    \caption{Self-routing capsule networks. \textbf{(a)} typical routing-by-agreement; \textbf{(b)} the self-routing mechanism proposed by~\cite{self-routing}. In self-routing, subordinate routing networks with parameters $\mathbf{W}^{route}_{i}$ are fed capsule pose vectors $\mathbf{u}_i$ and learn to output the routing coefficients $c_{ij}$ directly. Figure from ~\cite{self-routing}.}
    \label{fig:self-routing2}
\end{figure}
\textbf{Object Capsule Autoencoder (OCAE).} \ This stage proceeds PCAE and resembles the processes involved in the CCAE. All the parameters that have been extracted and identified from PCAE need to be composed in a way that form objects. This is achieved by providing concatenated poses $\mathbf{x}_m$, special features $\mathbf{z}_m$, and flattened templates $T_m$ as input to the OCAE. This differs from the CCAE in that the part capsules presence probabilities $d_m$ are fed into the OCAE's encoder to add bias to the attention mechanism of the Set Transformer~\cite{lee2019set} not to consider absent points. In addition, $d_m$s are also used to weight the part-capsules' log-likelihood, so that we do not take absent points into account. This is achieved by raising the likelihood of the $m^th$ part capsule to the power of $d_m$. Parts discovered by the PCAE have independent identities, therefore every part-pose is explained as an independent mixture of predictions from object capsules. The OCAE is trained by maximising the likelihood of the detected parts, and it learns to discover further structure in previously identified parts, leading to sparsely-activated object capsules. For more detailed information on the mathematical formulations of SCAE please refer to~\cite{kosiorek2019stacked}.

\begin{figure}[!t]
    \centering
    \includegraphics[width=1.05\columnwidth]{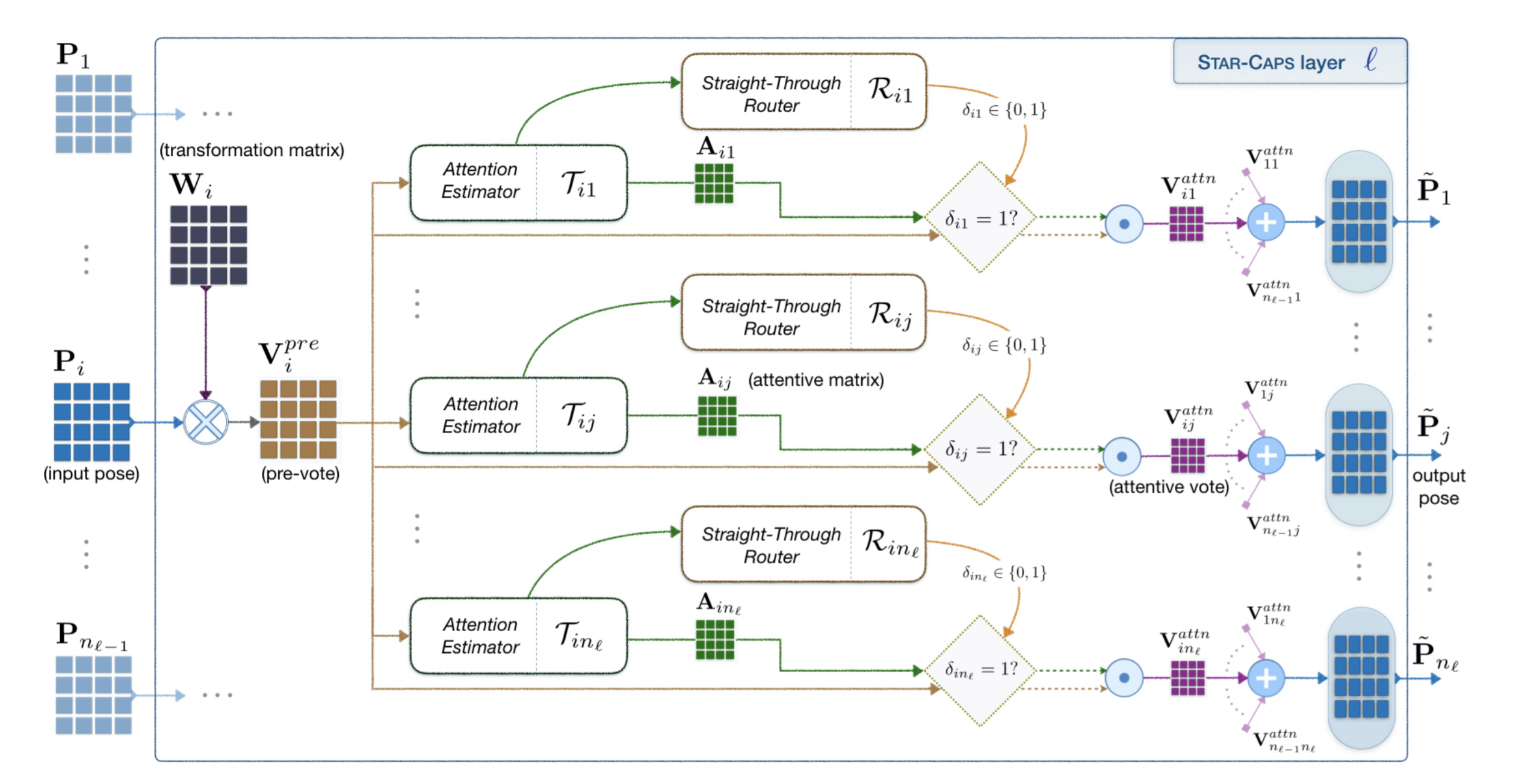}
    \caption{STAR-CAPS layer architecture~\cite{ahmed2019star}. Given the pose features from the lower-level capsules, the pose is transformed through shared trainable weight matrices (pre-vote). The routing between the lower-level and higher-level capsules takes place through two components: the Attention Estimator and the Straight-through Router. This router estimates a binary signal that decides whether to connect or disconnect the current route between the lower-level capsule and the higher-level capsule. Figure from ~\cite{ahmed2019star}.}
    \label{fig:star-caps}
\end{figure}
\subsection{Inference for Generative Capsule Models}
In most previous works on capsules~\cite{sabour2017dynamic,hinton2018matrix} an inference algorithm is typically presented without specification of a corresponding generative model for the data (with the exception of~\cite{kosiorek2019stacked} who use an autoencoder for generation). In~\cite{nazabal2021inference} the authors argue for a generative approach to model the relationships of objects and their parts in capsule networks. They state that it is more natural to describe the generative process by which an object gives rise to its parts, rather than the other way around as is typical. To that end, they present a principled generative capsules model that leads naturally to a variational algorithm for inferring the transformation of each object and the assignments of observed parts to the objects.

Their work is built on the premise that the input to a capsule should be a set of parts. For example, say we have an object $k$ with instantiation parameters $y_k$, and each object has parts $p_n$, where $n = 1,...,N_k$. These parts are then matched against observed parts $x_m$. Under a probabilistic framework, the authors get posterior distributions for both the $y_k$'s and the match variables $z_{mnk}$ that match $x_m$ to part $n$ of object $k$. Such a setup leads directly to a principled routing-by-agreement algorithm via variational inference, which can be derived similarly to the classical Gaussian Mixture Model~\cite{bishop2006pattern}. They therefore avoid having to devise a custom inference algorithm with an ad hoc objective function as previously proposed in EM-Routing capsules~\cite{hinton2018matrix}. 

The authors demonstrated that their approach outperforms the CCAE part of the SCAE method~\cite{kosiorek2019stacked} in the  constellations data generated from multiple geometric objects, e.g., triangles, squares, etc. that they used, as well as data from a parts-based model of faces.
They also demonstrated that random sample consensus (RANSAC)~\cite{fischler1981random}---where a minimal
number of parts are used in order to instantiate an object---is often an effective alternative to variational inference routing-by-agreement, especially when the basis in RANSAC is highly informative about the object.

\subsection{Self-Routing Capsule Networks}
The fact that capsules specialize in disjoint regions of the feature space leads to them making multiple predictions based on the information that is made available to them for each region. At a layer level, this means that we have an ensemble of submodules that are activated differently per example -- similar to a mixture of experts, where each expert specializes in different regions of input space. Motivated by this observation and the fact that routing-by-agreement is computationally costly, the authors in~\cite{self-routing} proposed a simpler self-routing strategy inspired by Mixture-of-Experts. 

In Self-Routing Capsule Networks (SR-CapsNet) proposed by~\cite{self-routing}, each capsule independently defines its routing coefficients without coordinating the agreement with other capsules. Instead, each capsule is empowered by higher modeling capabilities in the form of a subordinate routing network that learns to predict the routing coefficients directly (Figure~\ref{fig:self-routing2}). In self-routing, computing the routing coefficients $c_{ij}$ and predictions $\widehat{\mathbf{u}}_{j|i}$ involves two learnable weight matrices $\mathbf{W}^{route}$ and $\mathbf{W}^{pose}$ respectively. For each layer of the routing network, each pose vector $\mathbf{u}_i$ is multiplied by a trainable weight matrix $\mathbf{W}^{route}$ to output the routing coefficients directly. After softmax normalization, the calculated routing coefficients $c_{ij}$ are then multiplied by the capsule's activation scalar $a_i$ to generate weighted votes. The activation $a_j$ of an upper-layer capsule is the summation of the weighted votes of lower-level capsules over spatial dimensions $H \times W$, or $K \times K$ when using convolutions. The authors observed competitive performances on standard capsule network benchmarks, as well as improved robustness to adversarial attacks. Although this method is simple, it does somewhat limit the capsule network's capability to dynamically adjust the routing weights based on the input, since these are now fully determined by the learned parameters of the routing subnetworks. This approach is also reminiscent of the Synthesizer in Transformer literature~\cite{tay2021synthesizer}.

\begin{figure}[t]
    \centering
    \includegraphics[scale=0.47,trim=110 493 110 22, clip]{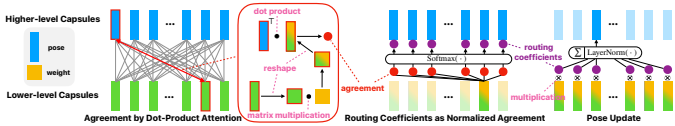}
    \caption{The Inverted Dot-Product Attention routing mechanism, showing the two-step process, i.e., the agreement between lower-level capsules and higher-level capsules, and the update of the pose of the higher-level capsules. Figure from ~\cite{Tsai2020Capsules}.
    }
    \label{fig:inverted}
\end{figure}
\subsection{Straight-Through Attentive Routing}
One of the main drawbacks of CapsNets is computational complexity that stems from the complex mechanisms of the voting and routing processes. Even if the CapsNet architecture has a fixed number of parameters, the number of routing iterations can increase the training and inference time greatly. In~\cite{ahmed2019star} a non-recursive attention-based routing mechanism is proposed, inspired by the non-recurrent self-attention approach found in Transformers~\cite{vaswani2017attention}. The proposed STAR-Caps layer architecture can be seen in Figure~\ref{fig:star-caps}, and utilizes a straight-through attentive routing mechanism, formulating each capsule as a matrix rather than a vector like in EM routing~\cite{hinton2018matrix}.

STAR-Caps~\cite{ahmed2019star} employs the following two-mechanism process for routing capsules: i) the attention estimator; ii) the straight-through router. The role of the attention estimator $\mathcal{T}_{ij}$, is to estimate the attention matrix $\mathbf{A}_{ij}$  $\in$   $\mathbb{R}^{p \times p}$ between lower and higher level capsules. The straight-through router $\mathcal{R}_{ij}$ decides which capsules to connect/disconnect. As shown in Figure~\ref{fig:star-caps}, given the attentive matrix $\mathbf{A}_{ij}$, the straight-through router $\mathcal{R}_{ij}$ acts as a gate that estimates a binary decision value $\delta_{ij} \in {0,1}$, indicating whether to disconnect ($\delta_{ij} = 0$) or connect ($\delta_{ij} = 1$) the route between capsules $i$ and $j$. This process is akin to hard attention, where each $\mathcal{R}_{ij}$ sends its hard attention signal to the higher-level capsules. To make this process differentiable they authors employ a straight-through estimator~\cite{bengio2013estimating,jang2016categorical,maddison2016concrete}.

As usual, the ClassCaps layer outputs the final predictions, where each capsule vector represents a single class. Like~\cite{sabour2017dynamic}, the authors encode activations implicitly in the capsule, and the final probability is given by a global average pooling operation on the poses followed by a logistic transformation. Given the activations, they then calculate the spread loss as in EM routing~\cite{hinton2018matrix} for training.
\subsection{Inverted Dot-Product Attention Routing}
In~\cite{Tsai2020Capsules}, the authors proposed a routing algorithm for capsule networks inspired by the attention mechanism commonly found in transformers~\cite{vaswani2017attention}(see Fig.~\ref{fig:inverted}). They design their routing algorithm via an inverted dot product attention mechanism that includes layer normalization when updating the poses of higher-level capsules. This approach is most similar to Dynamic routing~\cite{sabour2017dynamic} since the capsule voting scheme and agreement is computed in much the same way. Unlike Dynamic routing however, inverted dot-product attention routing introduces the concept of \textit{concurrent routing}, whereby multiple layers of capsules are routed concurrently rather than routing capsules in each each layer sequentially. 

\begin{figure}[!t]
        \centering
        \includegraphics[trim={0 0 0 0},clip,width=\columnwidth]{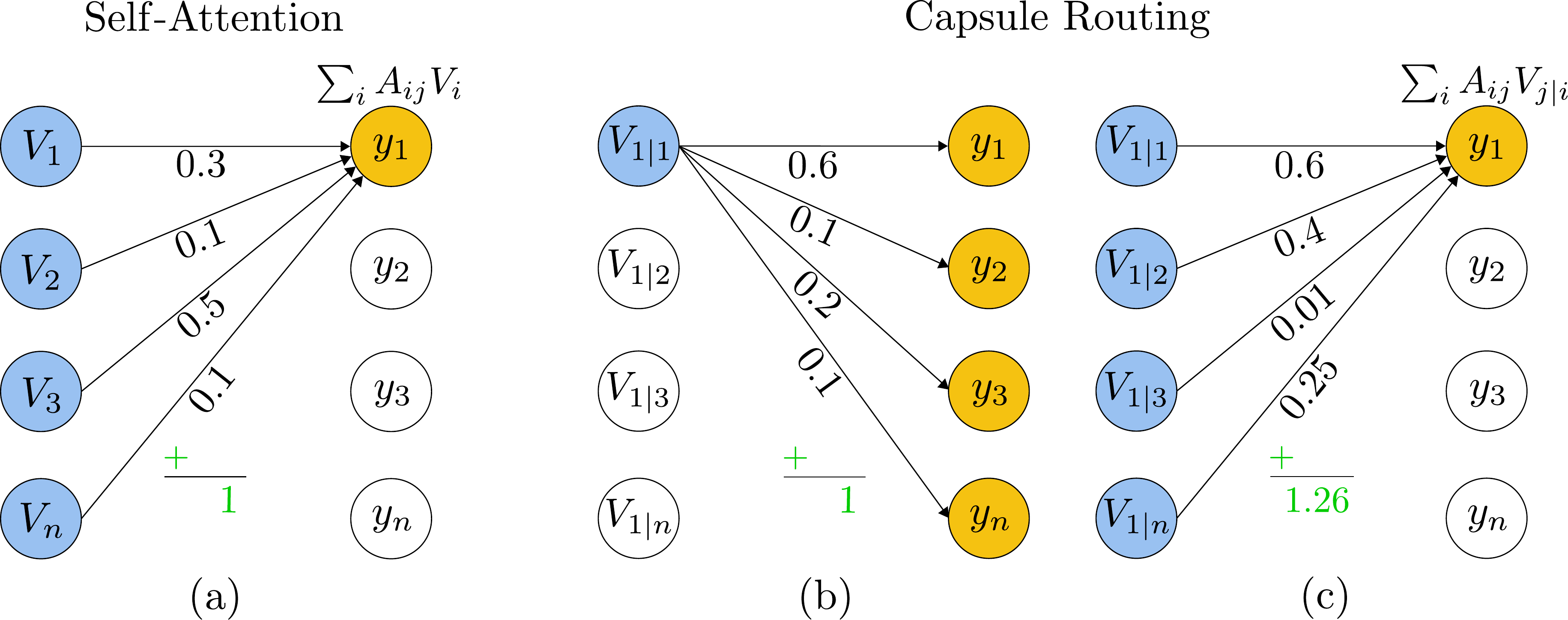}
        \caption{Comparing self-attention and capsule routing. \textbf{(a)} In self-attention, an output token (e.g. $y_1$) is a weighted average of the input values, where the weights sum to 1 over $i=1,\dots,n$. \textbf{(b)} In capsules, the weights of each input sum to 1 over the outputs instead. \textbf{(c)} A single output capsule (e.g. $y_1$) is a weighted average of the input votes with weights that do not necessarily sum to 1 over $i=1,\dots,n$.}
        \label{fig:caps_sa}
\end{figure}
\begin{figure*}
\centering
\begin{subfigure}{.45\textwidth}
\centering
\begin{minted}
[
% frame=lines,
% framesep=2mm,
% baselinestretch=1,
% bgcolor=LightGray,
fontsize=\scriptsize,
% linenos
]
{python}
def capsule_routing(X, iters=3):
    N, D = X.shape  # N-by-D inputs
    M, P = 3, 4     # M-by-P outputs

    W = torch.randn(N, M, D, P)             
    V = torch.einsum('ijdp,id->ijp', W, X)  
    Y = (1./M) * torch.sum(V, dim=0)            
    
    agreement = torch.zeros(n, m)
    for _ in range(iters): 
        agreement += torch.einsum(
            'jp,ipj->ij', Y, V.transpose(2,1)) 
        A = F.softmax(agreement, dim=-1)
        Y = torch.einsum('ij,ijp->jp', A, V)
    return Y
\end{minted}
\end{subfigure}
\begin{subfigure}{.45\textwidth}
\centering
\begin{minted}
[
% frame=lines,
% framesep=2mm,
% baselinestretch=1,
% bgcolor=LightGray,
fontsize=\scriptsize,
% linenos
]
{python}
def self_attention(X):
    N, D = X.shape  # N-by-D inputs
    M, P = n, 4     # M-by-P outputs, M=N

    Wq = torch.randn(D, P)  
    Wk = torch.randn(D, P)  
    Wv = torch.randn(D, P)  
    
    Q = torch.einsum('id,dp->ip', X, Wq)
    K = torch.einsum('id,dp->ip', X, Wk)
    V = torch.einsum('id,dp->ip', X, Wv)
    
    agreement = torch.einsum(
        'ip,pj->ij', Q, K.transpose(1,0))
    A = F.softmax(agreement / np.sqrt(d), dim=-1)
    Y = torch.einsum('ij,jp->ip', A, V)
    return Y
\end{minted}
\end{subfigure}
\caption{Comparing capsule routing and self-attention in Pytorch code~\cite{paszke2019pytorch}. Both operations take $i=1,\dots,N$ input vectors $\mathbf{x}_i \in \mathbb{R}^{D}$, and produce $j=1,\dots,M$ output vectors $\bm{y}_j \in \mathbb{R}^{P}$, where $M=N$ in self-attention. The outputs are attention weighted averages of the inputs, with attention weights $A \in \mathbb{R}^{N \times M}$ computed using the \textit{agreement} (similarity) between activity vectors. Notice how the final steps are similar, and that in capsule routing the output capsules $\mathbf{Y}$ act as the queries $\mathbf{Q}$ in self-attention, and the capsule votes $\mathbf{V}_{j|i}$ act as both the keys $\mathbf{K}$ and the values $\mathbf{V}$.}
\label{fig:caps_sa_code}
\end{figure*}
\section{Attention \& Capsules}
As previously alluded to in Section~\ref{sec:introduction}, there are notable conceptual similarities between capsule routing and the self-attention mechanism popularised by transformers~\cite{vaswani2017attention}. In this section, we first provide a detailed breakdown of the relationship between them, and show how we can think of each method from a unified perspective using similar notation. There is also significant conceptual overlap between capsule networks and other object-centric representation learning techniques~\cite{greff2020binding,locatello2020object}, thus we discuss previous research on using the attention mechanism for object-centric learning, and highlight conceptual similarities to the capsule formulation along the way. When appropriate, we adopt Einstein index notation to make it easier to highlight the similarities between the two methods, and provide respective code examples in Figure~\ref{fig:caps_sa_code}.

\subsection{On Self-Attention \& Capsule Routing}
\label{subsec:sa_caps}
In this section, we introduce both self-attention and capsule routing as ``agreement machines", consisting of dynamic weighted averaging layers that operate on $D$ dimensional vector-valued units $\mathbf{x}_i \in \mathbb{R}^D$. These vector-valued units are known as token embedding vectors in transformers and capsules in capsule networks. To illustrate this, let $\mathbf{X} = (\mathbf{x}_1,\dots,\mathbf{x}_N) \in \mathbb{R}^{N \times D}$ denote a matrix of input token embedding vectors or capsules $\mathbf{x}_i = (x_1,\dots,x_D)$. Consider the computation of a single output token $\bm{y}_j \in \mathbb{R}^P$ in simple self-attention, given a sequence of input tokens $\{\mathbf{x}_i\}^N_{i=1}$:
\begin{equation}
    \bm{y}_j = \sum_{i=1}^N A_{ij} \mathbf{x}_{i},
\end{equation}
where $A \in \mathbb{R}^{N \times M}$ is an attention weight matrix between all input/output pairs, with $N=M$ and $0 \leq A_{ij} \leq 1$. Similarly, to compute a single output capsule $\bm{y}_j \in \mathbb{R}^P$, given input capsules $\{\mathbf{x}_i\}^N_{i=1}$, we have the same expression: $\bm{y}_j = \sum_{i=1}^N A_{ij} \mathbf{x}_{i}$. 

As we describe in greater detail next, the main differences lie in the precise introduction of parameters and how the attention weights are normalised. Moreover, we often have fewer outputs than input capsules ($M < N$), making $A$ no longer a square matrix like in self-attention.

\textbf{Self-Attention.} \ In scaled dot product self-attention~\cite{vaswani2017attention}, the $N$ input vectors $\mathbf{X} \in \mathbb{R}^{N \times D}$ are transformed into respective query, key and value matrices as follows:
%
\begin{align}
    \mathbf{Q} = \mathbf{X} \mathbf{W}^Q, &&\mathbf{K} = \mathbf{X} \mathbf{W}^K, &&\mathbf{V} = \mathbf{X} \mathbf{W}^V,
\end{align}
where $\mathbf{W}^Q$, $\mathbf{W}^K$ and $\mathbf{W}^V$ are $D \times P$ dimensional parameter matrices, and $\mathbf{Q}$, $\mathbf{K}$ and $\mathbf{V}$ are therefore $N \times P$ dimensional. Let's now consider the calculation of the attention weights for a single input token $\mathbf{x}_i$. Given $\mathbf{x}_i$'s corresponding query row vector $\mathbf{Q}_{i,:} \in \mathbb{R}^{1 \times P}$, its attention weights $A_{i,:} \in \mathbb{R}^{1 \times N}$ are given by the normalised dot product with each of the key vectors:
\begin{equation}
    \label{eq:self-attn}
    A_{i,:} = \mathrm{softmax}\Big(\underbrace{\frac{\mathbf{Q}_{i,:}\mathbf{K}^\top}{\sqrt{D}}}_{\mathrm{agreement}}\Big).
\end{equation}
The dot product \textit{agreement} between token $\mathbf{x}_i$'s query $\mathbf{Q}_{i,:}$ and all the keys $\mathbf{K}$ dictates how much ``value'' $\mathbf{V} = (V_1,\dots,V_N) \in \mathbb{R}^{N \times P}$ from each other token should be represented in token $\mathbf{x}_i$'s revised representation. That is, a single output token $\mathbf{Y}_{i,:} \in \mathbb{R}^{1\times P}$ is simply a weighted average of the input token's values: 
\begin{align}
    \mathbf{Y}_{ip} = (A\mathbf{V})_{ip} = \sum_{j=1}^N A_{ij} V_{jp},     
\end{align}
for $i=1,\dots,N$ and $p=1,\dots,P$. Each output token $\bm{y}_i \coloneqq \mathbf{Y}_{i,:} \in \mathbb{R}^{1\times P}$ constitutes the revised representation for $\mathbf{x}_i$.

\textbf{Capsule Routing.} \ Comparatively, a single output capsule is given by first calculating the capsule votes, which are each input capsule's prediction of what the output capsule should be. Using tensor contraction notation as above, we start by multiplying input capsules $\mathbf{X} \in \mathbb{R}^{N \times D}$ by a 3D learned parameter matrix (3-tensor) $\mathbf{W} \in \mathbb{R}^{N \times D \times P}$:
\begin{equation}
    \mathbf{V}_{ip} = \sum_{d=1}^D W_{idp} X_{id},
\end{equation}
for $i=1,\dots,N$ and $p=1,\dots,P$. The resulting votes are $\mathbf{V} = (V_1,\dots,V_N) \in \mathbb{R}^{N \times P}$. Note that unlike in self-attention, here we have $N$ separate weight matrices, one for each input vector-valued unit $\{\mathbf{x}_i\}_{i=1}^N$, i.e. capsule. In the first routing iteration, the output of the $j^{\text{th}}$ capsule is an average of its votes: $\bm{y}_j = \frac{1}{M} \sum_{i=1}^N V_{i}$, since the attention weights are uniform over outputs. The \textit{agreement} (also measured by the dot product) between the votes and the output is then used to iteratively revise both the attention weights and the output: $\bm{y}_j = \sum_{i=1}^N A_{ij} V_{i}$. However, unlike in self-attention, to compute the attention weights for a single output capsule we require the context of all the other output capsules in the same layer. Indeed, there is no direct equivalent equation to Eq.~\eqref{eq:self-attn} we can use here without breaking the softmax normalisation. Moreover, since each output capsule receives $i=1,\dots,N$ votes, in order to compute $j=1,\dots,M$ output capsules we require the transformation weights to be a 4-tensor: $\mathbf{W} \in \mathbb{R}^{N \times M \times D \times P}$. Thus there is a learned ${D \times P}$ matrix between every input/output capsule pair.

With that in mind, carefully consider the following full procedure for computing a single output capsule in a capsule layer, in conjuction with the accompanying code implementation in Figure~\ref{fig:caps_sa_code}. The steps in Eqs.~\eqref{eq:caps1} to~\eqref{eq:caps3} can be repeated to constitute \textit{routing}, by iteratively refining the initial uniform attention weights in Eq.~\eqref{eq:caps1} with new estimates from Eq.~\eqref{eq:caps_norm}:
\begin{align}
    \mathbf{V}_{ijp} &= \sum_{d=1}^D W_{ijdp} X_{id},  &(\mathrm{voting}) \\
    \bm{y}_{jp} &= \frac{1}{M} \sum_{i=1}^N V_{ijp}, &(\mathrm{avg. \ votes}) \label{eq:caps1} \\
    A_{ij} &= \mathrm{softmax}\Big(\underbrace{\sum_{p=1}^{P} V_{ijp} \bm{y}_{jp}}_{\mathrm{agreement}}\Big)_j, & (\mathrm{attention}) \label{eq:caps_norm} \\
    \bm{y}_{jp} &= \sum_{i=1}^N A_{ij} V_{ijp}, &(\mathrm{output \ capsule})
    \label{eq:caps3} 
\end{align}
for each $j=1,\dots,M$ output capsule. Notably, Eq.~\eqref{eq:caps_norm} entails a softmax normalisation over outputs, rather than inputs as in self-attention (Eq.~\eqref{eq:self-attn}). As depicted in Figure~\ref{fig:caps_sa}, the input capsules spread their value among output capsules, causing output capsules to compete with each other for input capsule's values---whereas in self-attention the competition is between the input tokens instead.

In the language of transformers, we can think of the output capsules as the query, and the capsule votes act as both the keys and the values in self-attention. To calculate the attention weights in capsule routing we compute the \textit{agreement} between outputs and votes via the dot product, just like we would in self-attention.

\begin{figure}[!t]
        \centering
        \includegraphics[trim={0 0 0 0},clip,width=\columnwidth]{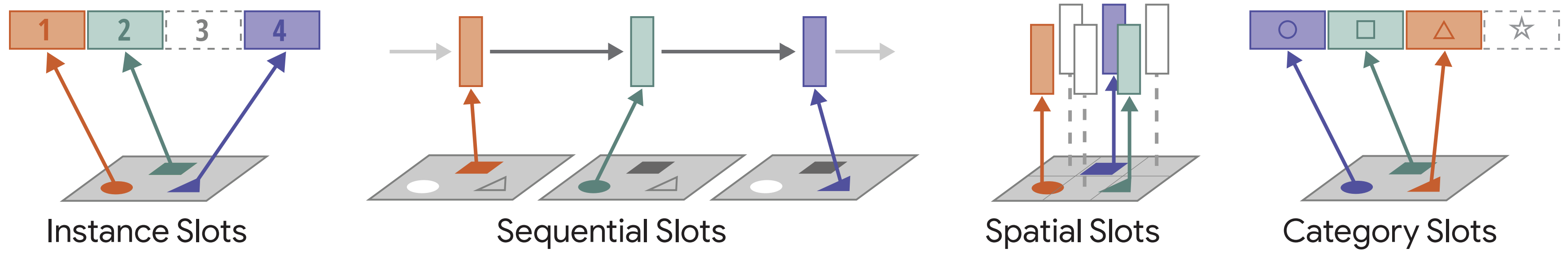}
        \caption{Depicting the four different types of slot-based representations, from~\cite{greff2020binding}. Capsules are typically instantiated as \textit{category slots} since they \textit{bind} to objects in the input based on some categorical criteria like class identity. Convolutional capsules~\cite{hinton2018matrix} can also be thought of as \textit{spatial slots}. Figure from ~\cite{greff2020binding}}
        \label{fig:slots}
\end{figure}
\subsection{Comparing Inductive Biases} 
Having established a formal relationship between self-attention and capsule routing, in this section we compare and contrast the inductive biases inherent to both methodologies. Recall that an inductive bias of a learning algorithm is a modelling assumption which induces a preference for certain solutions. Inductive biases often consist of encoding useful prior assumptions about the target function mapping inputs to outputs, that can aid in generalisation to unseen cases and reduce sample complexity. 

\textbf{Capsule.} The exposition in Section~\ref{subsec:sa_caps} highlights the inductive biases induced by the capsule formulation are such that each part (input) capsule belongs to a single object (output capsule), and each object must compete with other objects for parts. This is also known as the ``single parent'' assumption, commonly found in mixture models and clustering algorithms. Moreover, as previously outlined in Section~\ref{subsec:foundations}, the (per-capsule) vote transformation matrix $\mathbf{W}$ is biased towards encoding invariance to viewpoint transformations, and the capsule vectors are biased towards capturing equivariance of neural activities. 

\textbf{Self-Attention.} On the other hand, the self-attention mechanism induces weaker inductive biases since there are no equivariance or single parent assumptions like in CNNs or capsule networks for example. This relaxed inductive bias makes transformers with self-attention very flexible models, but it also means that more data is typically required to match the performance of models with more explicit inductive biases~\cite{dosovitskiy2020image}, since a portion of the modelling capacity has to be spent on learning to encode any useful biases. With that said, one inductive bias we can interpret from self-attention is that each output token is best explained by a single input token (due to the softmax normalisation over inputs) which can be thought of as a ``single child'' assumption. Subsequently, input tokens compete with each other to be included in each output token's revised representation through pairwise interactions.

\textbf{Common Ground.} Lastly, it is important to note that both capsules and attention share some key inductive biases, such as: using vectors of neural activity to represent a collection/hierarchy of concepts, and taking the \textit{agreement} between these high dimensional vectors as a feature detection mechanism. The relationships between these concepts is then dynamically adjustable based on the input, and the concepts themselves are refined based on global context. 
\begin{figure}[!t]
        \centering
        \includegraphics[trim={0 0 0 0},clip,width=\columnwidth]{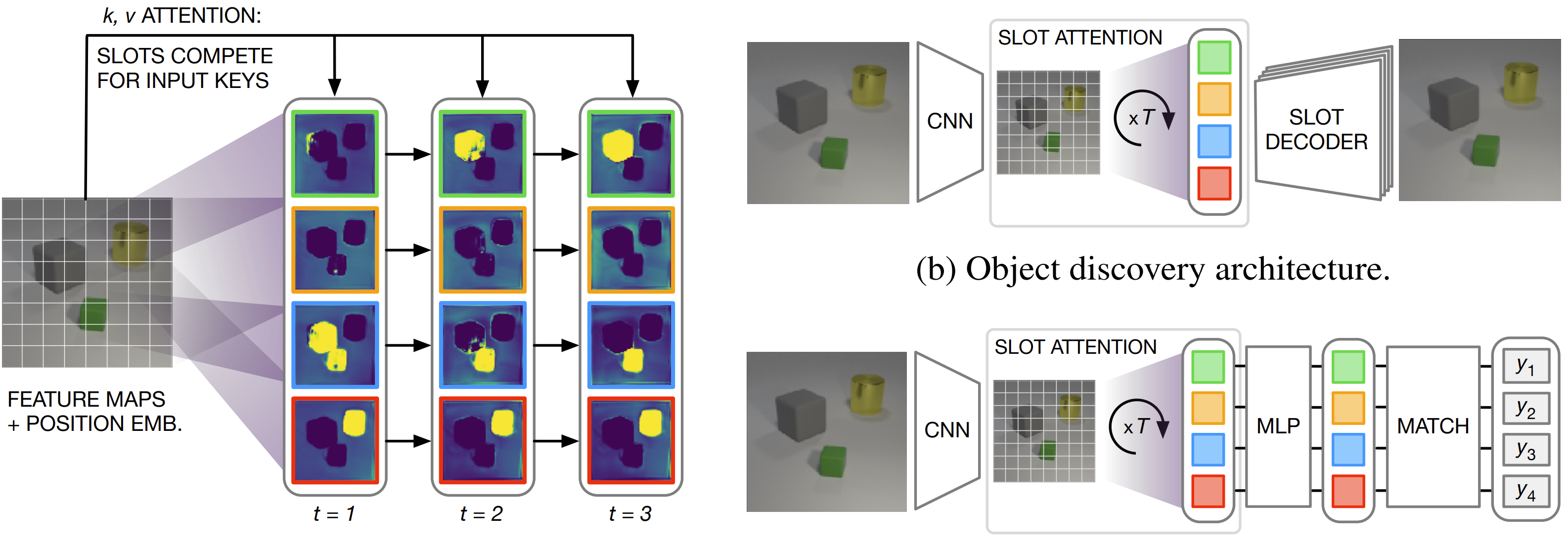}
        \caption{Slot attention module (left), object discovery and set prediction architecture (bottom right). Slots are ``universal'' capsules that can bind to any object in the input. Figure from ~\cite{locatello2020object}.}
        \label{fig:slot_attention}
\end{figure}
\begin{figure}[!t]
        \centering
        \includegraphics[trim={0 0 0 0},clip,width=.95\columnwidth]{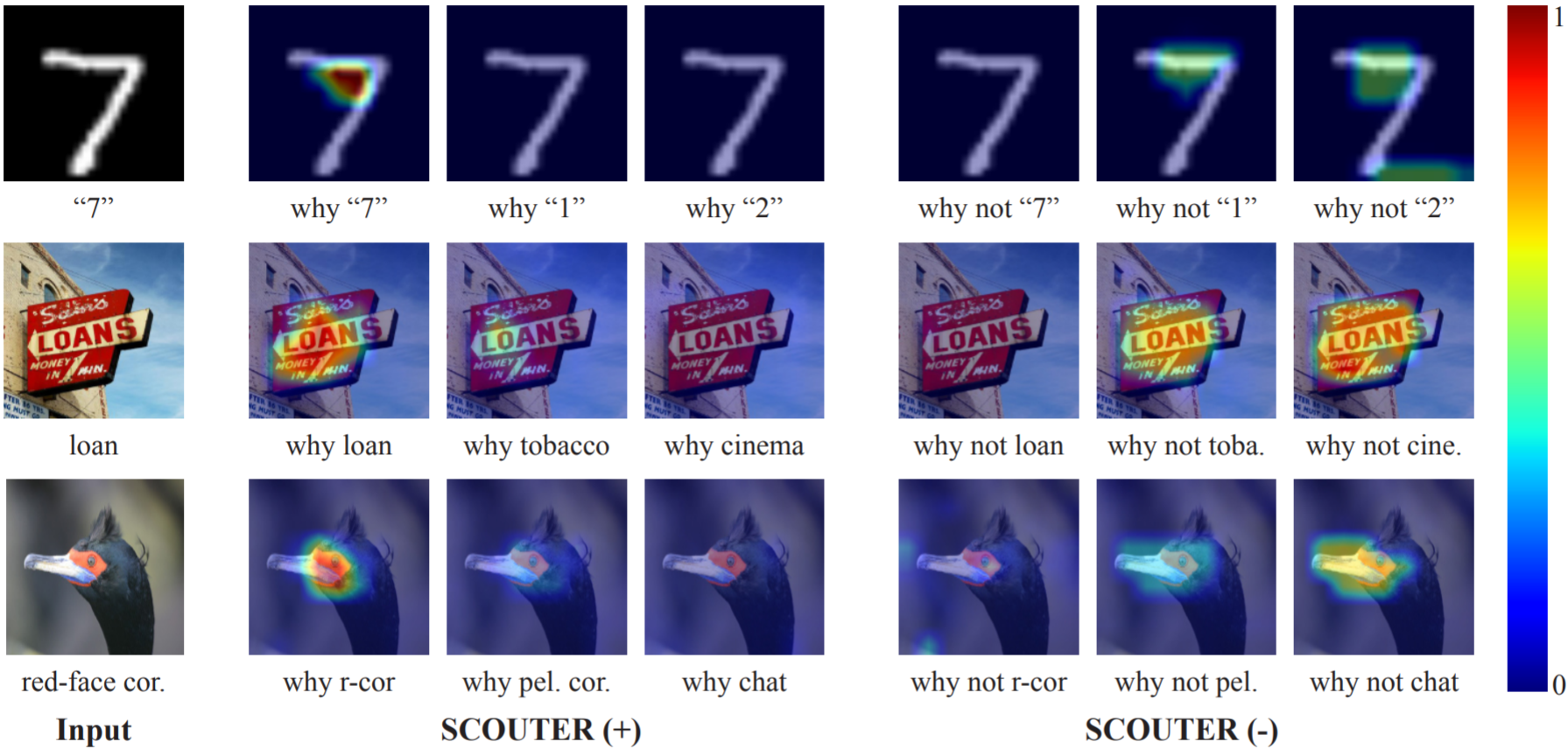}
        \caption{Positive and negative explanations given by SCOUTER~\cite{li2021scouter}: a variant of slot-attention with specialised slots that bind to class categories like in capsule networks. The \textit{support} for SCOUTER slots can provide explanations.Figure from ~\cite{li2021scouter}.}
        \label{fig:scouter}
\end{figure}
\subsection{Slots \& Attention}
Many object-centric representation learning approaches using neural networks can be categorised as being \textit{slot} based~\cite{greff2020binding}. As shown in Figure~\ref{fig:slots}, slots constitute a general representational format used for separating object-based representations. They provide a sort of working memory with fixed capacity which can be used to access independent object representations simultaneously. There are four main types of slots as outlined by~\cite{greff2020binding}, but to remain within the scope of this survey, we focus mainly on: (i) \textit{category slots}, which are the most commonly used representational format in capsule networks; (ii) \textit{instance slots}, which can be classed as ``universal'' capsules that can bind to multiple objects rather than a specific category.

\textbf{Slot-Attention.} Recent work by~\cite{locatello2020object} showed how attention can be used to extract object-centric representations that enable generalization to unseen compositions, which is the motivation behind capsule networks. Indeed, the proposed method they call \textit{Slot Attention} is reminiscent of recent developments in both capsule networks and self-attention. As shown in Figure~\ref{fig:slot_attention}, the authors introduce the slot attention module, a differentiable interface between the outputs of a CNN and a set of variables they call \textit{slots}. They employ an iterative attention mechanism, much like capsule routing, wherein the slots play the role of the capsules. However, unlike capsules, the slots produced by slot attention do not specialise to one particular type of class or object, instead they can store/bind to any object in the input, making them more flexible. Because of this, slots have been referred to as ``universal'' capsules~\cite{hinton2021represent}, as they can contain enough knowledge to model more than one type of object/part.

Nonetheless, the \textit{slots} still compete with each other at each iteration for explaining parts of the input via a softmax-based attention mechanism, just like in capsule networks and transformers. In fact, the slot attention iterations in their method can be thought of as equivalent to unrolled transformer layers that share parameters. This is reminiscent of capsule routing, whereby the routing iterations can be thought of as being equivalent to unrolled layers of attention that share parameters. In their experiments, the authors demonstrate that slot attention is competitive with previous approaches on unsupervised visual scene decomposition tasks, whilst being more efficient. 

More recently, the authors in \cite{li2021scouter} proposed SCOUTER, a slot-attention based classifier for transparent, explainable and accurate classification (see Figure~\ref{fig:scouter}). The main difference between SCOUTER and vanilla slot attention is that the slots are now associated to single categories like in capsule networks. Indeed, the \textit{evidence} for a certain category in SCOUTER can be thought of as its \textit{support} in capsule networks, i.e. using an attention mechanism to find support in the image that directly correlates to a certain output category. The authors in ~\cite{li2021scouter} also employ an iterative attention mechanism to update the slots, where the number of iterations $T{=}3$ just like in capsule routing and vanilla slot attention. We can think of SCOUTER as a capsule network with restricted inductive biases and fewer parameters, thus the explainability insights from SCOUTER are applicable to capsule networks.
\begin{figure}[t!]
        \centering
        \includegraphics[trim={0 0 0 0},clip,width=.95\columnwidth]{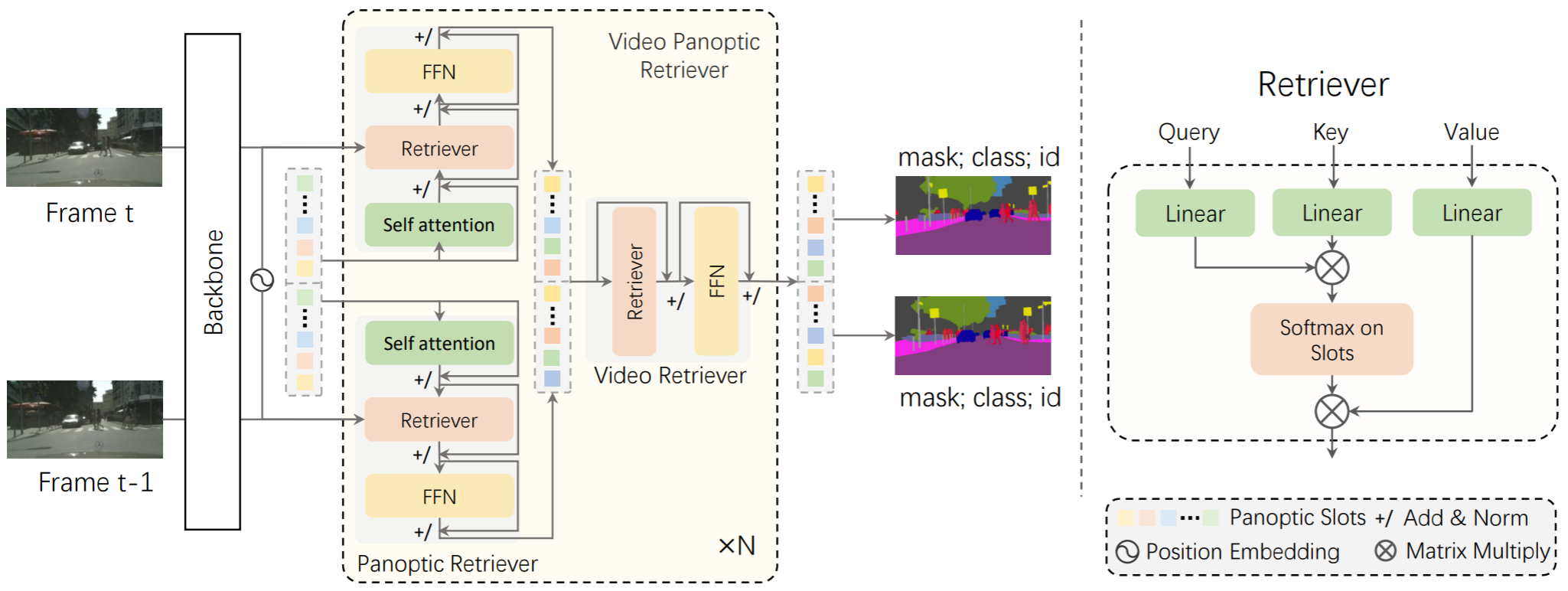}
        \caption{The Slot-VPS system~\cite{zhou2021slot} based on slot attention. Position embeddings and slots are fed into the retriever modules to generate spatio-temporal coherent panoptic slots. The slots are then directly converted into objects' masks, classes and IDs. Figure from ~\cite{zhou2021slot}.}
        \label{fig:vpr}
\end{figure}
The authors in~\cite{zhou2021slot} bring the slot attention ideas to real world data, and achieve state-of-the-art performance on video panoptic segmentation tasks. The proposed Video Panoptic Retriever (VPR) (see Figure~\ref{fig:vpr}) retrieves and encodes all panoptic entities in a video, including both foreground instances and background semantics, with a unified object-centric representation called panoptic slots. The output panoptic slots can be directly converted into the class, mask, and object ID of panoptic objects in videos.
\subsection{Transformers for Routing \& Sets}
Sun et al. \cite{sun2021visual} propose a visual parser that attempt to learn part-whole hierarchies through attention operations. The visual parser learns a two level hierarchy iteratively refining the part and whole representations. At each iteration the part encoder uses a set of learned part prototypes and performs an attention operation on the previous whole representations to obtain a set of $N$ part representations. Then the whole decoder refine the previous whole representation with the global information in this set of parts. Using this iterative encoder-decoder structure, the visual parser learn robust representations which can be applied to several tasks including image classification, object detection, and instance segmentation.

Carion et al. \cite{carion2020end} present a transformer-based framework for object detection. After extracting a spatio-temporal grid of features from an image, the DEtection TRansformer (DETR) uses an encoder-decoder architecture to generate a set of $N$ object predictions. This is accomplished by using $N$ learned object queries in the transformer decoder whose output features are given to a feed forward network to generate the class and bounding box dimensions. The network is trained end-to-end with a bipartite graph matching loss which attempts to minimise the difference between the predicted and ground-truth objects. 

Self-attention has a quadratic $O(N^2)$ space and time complexity for a $N$ inputs. Wu et al. \cite{wu2021centroid} alleviate this by proposing the centroid transformer which clusters the $N$ inputs into a set of $M$ centroids which are then passed to the self-attention operation resulting in an $O(NM)$ complexity. This clustering operation can be viewed as a means to "route" the information from $N$ inputs (parts) to $M$ higher-level outputs (wholes).
Similarly, Roy \etal \cite{roy2021efficient} reduce the computational cost of the self-attention operation by only computing attention between a subset of keys and queries. Given an input sequence of length $N$, a clustering operation (k-means) is performed on the keys and queries to obtain $k$ centroids. Then, for each of the $N$ inputs, attention is computed on the set of keys which belongs to the same centroid as its given query. This proposed ``Routing Transformer" reduces the computational cost from $O(N^2)$ to $O(N^{1.5})$, and outputs a sequence length of $N$.

Contrary to standard self-attention, the set transformer \cite{lee2019set} performs the attention operation on a fixed set of $m$ learned query vectors (also known as inducing points). This allows for reduced computational cost when the number of input vectors ($n$) becomes large (\ie operation becomes $O(mn)$ as opposed to $O(n^2)$ in self-attention). Furthermore, set transformers have been utilised in as a capsule routing procedure in the Stacked Capsule Auto-encoder \cite{kosiorek2019stacked}: the part capsules are passed through multiple set transformer layers to obtain a set of object capsules.
Although the inducing points are learned in the set transformer, they are static and do not change based on the given input. Zare \etal \cite{zare2021picaso} remedy this by introducing a ``PICASO" block to update the learned inducing points based on information from a given input. By passing the inducing points through multi-head attention blocks, PICASO leads to improved representations for down-stream tasks including classification, clustering, and anomaly detection.

\begin{figure}[t!]
    \centering
    \includegraphics[width=1.0\columnwidth]{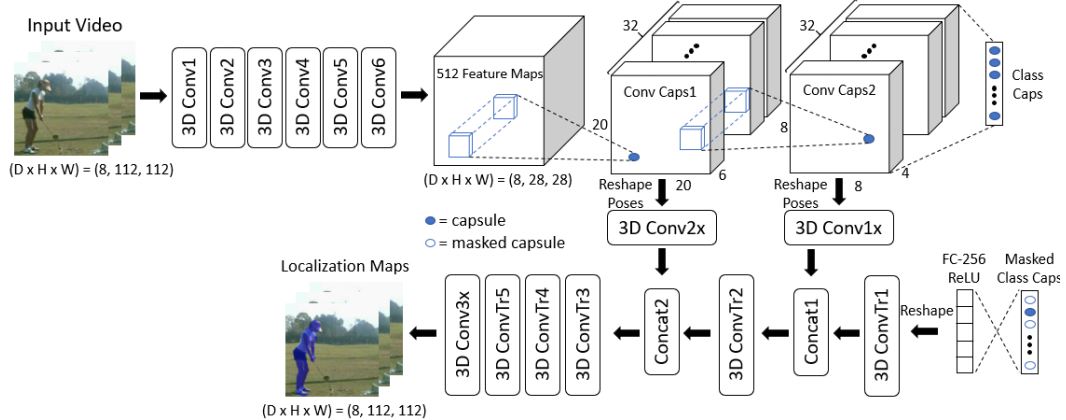}
    \caption{The VideoCapsuleNet architecture proposed by~\cite{duarte2018videocapsulenet}, which includes 3D convolutions and capsule pooling for processing video inputs.}
\end{figure}
\subsection{Relational Neural Expectation Maximization} The learning of object centric representations through Neural Expectation Maximization (N-EM) was studied by \cite{greff2017neural,van2018relational}. N-EM is a probabilistic model which attempts to group pixels within an image into $K$ entities whose properties are described by a vector $\theta_k$. A differentiable Expectation Maximization (EM) algorithm is used to find these groupings by computing the Maximum Likelihood Estimate for each $\theta_k$. Van \etal \cite{van2018relational} extended this work by proposing a Relational N-EM (R-NEM) approach to learn interactions between different entities (parts/objects) over time. By replacing the M-step of the EM algorithm with a recurrent neural network, R-NEM is able to model the temporal dynamics of a given scene. Although these methods were primarily evaluated on primitive objects and shapes (\eg triangles, squares, and circles), they have similar goals to capsule networks and constitute a promising probabilistic alternative for learning object centric representations of visual data.



\begin{figure}[t!]
    \centering
    \includegraphics[width=1.0\columnwidth]{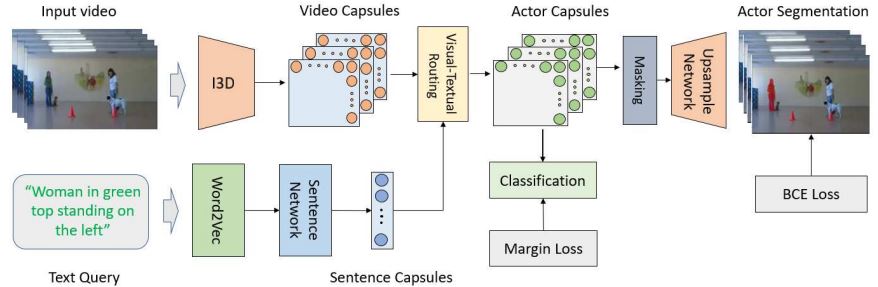}
    \caption{
        Depiction of Visual-Textual Capsule Routing as proposed by~\cite{mcintosh2020visual}. The framework produces a segmentation of an actor/object from video and text description pairs.
    }
\end{figure}
\section{Capsules for Video and Motion}

Although the majority of foundational capsule approaches tend to be applied to image data, there have been several works that focus on the video domain. Generalizing 2-dimensional image-based capsule networks to the 3-dimensional video domain is non-trivial. Applying  capsule networks to video data raises several questions. First, with the addition of a temporal dimension, how can capsule networks successfully capture the motion information from multiple frames, or time-steps, in a video sequence? Second, how can the iterative and computational costly routing operations scale to deal with video inputs, which tend to be much larger than images conventionally processed by previous capsule-based approaches? There is no current capsule work which completely answers these questions, but there have been several works which apply capsule networks to various video and motion problems.

Similar to how 2D image-based convolutions were generalized to 3D convolutions \cite{tran2015learning} to process a sequence of video frames, traditional 2D convolutional capsule routing was extended to 3D convolutional routing in \cite{duarte2018videocapsulenet}. In 3D convolutional routing, capsules which are both spatially and temporally nearby are routed together to obtain the higher layers' capsule outputs. Since the number of capsules being routed increases drastically as the size of the receptive field increases, conventional iterative routing operations, without modification, are unsuited for 3D capsule networks. To this end, a capsule-pooling procedure is proposed, which averages each capsule types' poses and activations within the receptive field. Capsule-pooling ensures the number of capsules routed is only proportional to the number of capsule types within each layer, rather than the size of the receptive field. Duarte etal. \cite{duarte2018videocapsulenet} present a video capsule network, VideoCapsuleNet, which performs end-to-end action detection. The network consists of 3D convolutional layers to transform the input RGB video sequence (8 frames) into the initial video capsules. Then, this is followed by a 3D convolutional capsule layer with capsule pooling, followed by a fully connected capsule layer to produce class capsules. VideoCapsuleNet is not only able to classify the action being performed within the video, but also spatio-temporally localizes the video by the use of a convolutional decoder. 
\begin{figure}[t!]
    \centering
    \includegraphics[width=1.0\columnwidth]{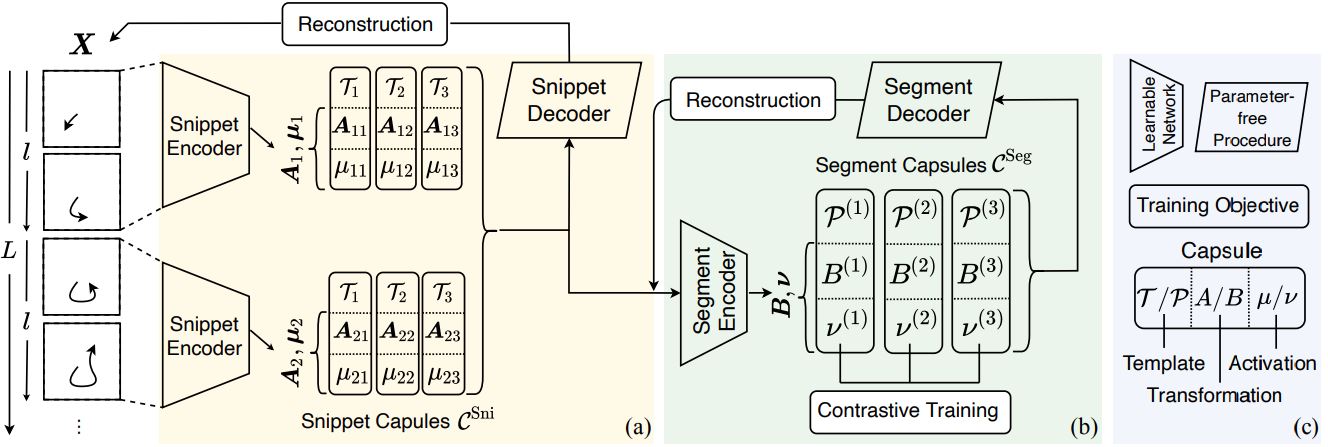}
    \caption{The Motion Capsule Autoencoder (MCAE), figure from~\cite{xu2021unsupervised}. The MCAE presents an elegant capsule-based approach to directly model motion in input sequences. ``Snippet'' (part) and ``segment'' (object) capsules contain semantic-agnostic information over shorter and longer time frames respectively.}
    \label{fig:videocapsules}
\end{figure}

This idea is extended in \cite{duarte2019capsulevos}, where the authors propose CapsuleVOS, a network which can perform video object segmentation. Given a video clip and the segmentation of the object of interest in the first frame, CapsuleVOS propagates the segmentation through all frames of the video. The network consists of two branches which generate capsules for the video clip (video capsules) and capsules for the first frame and segmentation (frame capsules). Then an attention routing algorithm is proposed to condition the video capsules based on their agreement with the frame capsules. This routing procedure first performs EM-routing on the frame capsules to obtain some higher-level capsule representation for the object in the first frame. Then, the routing coefficients are obtained for the video capsules by measuring their similarity to the higher-level frame capsules. The resultant video capsule representations are used by a convolutional decoder to segment the object of interest throughout the input video clip.

A capsule-based approach for regression tracking has been proposed in \cite{ma2021capsulerrt}. Instead of obtaining a single set of video capsules from 3D convolutions, two sets of capsules S-Caps and T-Caps are obtained which learn the spatial and temporal relationships within the video. These two sets of capsules are then combined and passed through a series of convolutional routing layers to obtain regression capsules (RegCaps) which classify the target and background. Finally, the pose matrices of the RegCaps are compressed using knowledge distillation to reduce computational cost and obtain more discriminative capsule representations.

Video capsule networks have also been applied to the multimodal domain of actor and action video segmentation from a sentence. Given a video and a natural language description \cite{mcintosh2020visual} propose an end-to-end capsule network that segments the object/actor described by the description. The network first extracts a spatial grid of video capsules which represent the various entities or objects within the video. From the input sentence, a set of sentence capsules are obtained. Then a visual-textual routing algorithm combines both capsule modalities at each location on the spatial grid. These capsules are then sent through a convolutional decoder network to obtain the output segmentation mask for the actor described in the sentence.

Previous video capsule networks implicitly learn temporal and motion information through 3D convolutions and routing. Recently, a capsule autoencoder architecture has been proposed to explicitly learn robust motion representations \cite{xu2021unsupervised}. This work takes concepts from the stacked capsule autoencoders (see Section \ref{sec:scae}), but the Motion Capsule Autoencoder (MCAE) replaces the part and object capsules with ``snippet" and ``segment" capsules. Here, a snippet capsule contains a semantic-agnostic representation for a short time-frame and a segment capsule contains a semantic-aware representation for a longer time-frame. Segment capsules are obtained by aggregating the snippet capsules and reconstructing their parameters. Although this work shows impressive results in unsupervised motion representation learning, the MCAE operates on individual points (2-dimensional coordinates) and not directly on video pixels. Nonetheless, MCAE presents an elegant capsule-based approach to directly model motion in input sequences, and extending such a approach to RGB videos is an interesting avenue for future work.

\section{Geometric and Graph based Capsules}

\begin{figure}[t]
        \centering
        \includegraphics[trim={20 0 0 40},clip,width=\columnwidth]{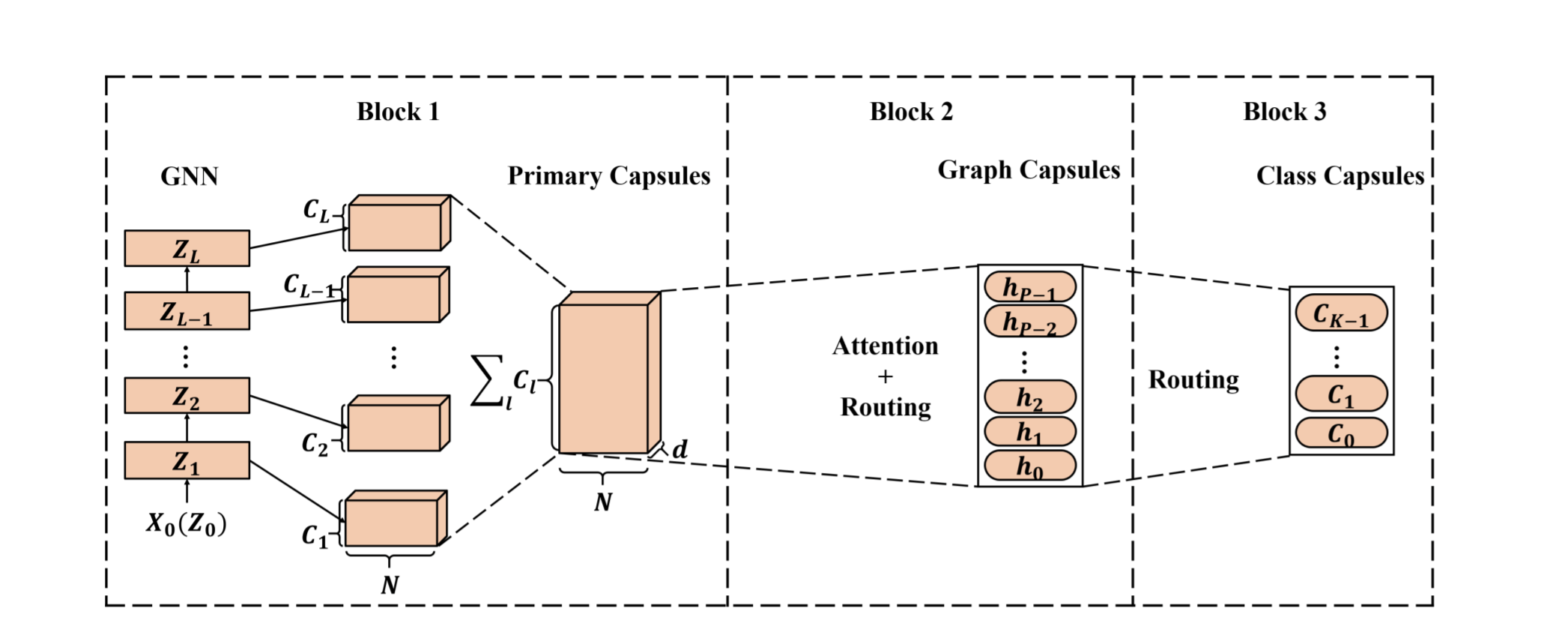}
        \caption{Framework of CapsGNN taken from ~\cite{xinyi2018capsule}. GNN is used to extract node embeddings and form primary capsules which are then scaled via an attention module and dynamically routed to generate graph capsules. Lastly, dynamic routing is applied again to perform graph classification.
        }
        \label{fig:capsgraph}
\end{figure}

In the past couple of years there have been several studies published that proposed CapsNets variations based on graphs that can better model the topological information of structured data and other types of data, such as social networks. Similarly, CapsNet models have been proposed for processing 3D point clouds that are equivariant to 3D rotations and translations, as well as invariant to permutations of the input points. One of the earliest works is  by Xinyi \etal ~\cite{xinyi2018capsule}, who  proposed CapsGNN, which is a framework that combines graph neural networks (GNN) and capsules.
GNN is used to extract node embeddings which are fed onto primary capsules (Block 1). At the second stage the node embeddings are scaled via an attention module that together with dynamic routing generate the graph capsules (Block 2). At the last stage, graph classification takes place via dynamic routing (Block 3). The whole framework can be seen in Figure~\ref{fig:capsgraph}

Li \etal ~\cite{li2021graph} proposed a graph-based capsule routing mechanism that focuses on learning intra-relationships between capsules in each layer, which is relevant to text classification problems. Intra-relationships that are found in text data need to be taken into account, along with hierarchical relationships, in order to improve sentiment analysis. The proposed method treats capsules in each layer as nodes in a graph and applies a new routing mechanism that combines bottom-up routing and top-down attention to learn hierarchical- and intra- relationships. Finally, the relationship between different capsules is evaluated by the Wasserstein distance, and a normalization trick is used to approximate the adjacency matrix. 

\begin{figure}[t]
        \centering
        \includegraphics[width=\columnwidth]{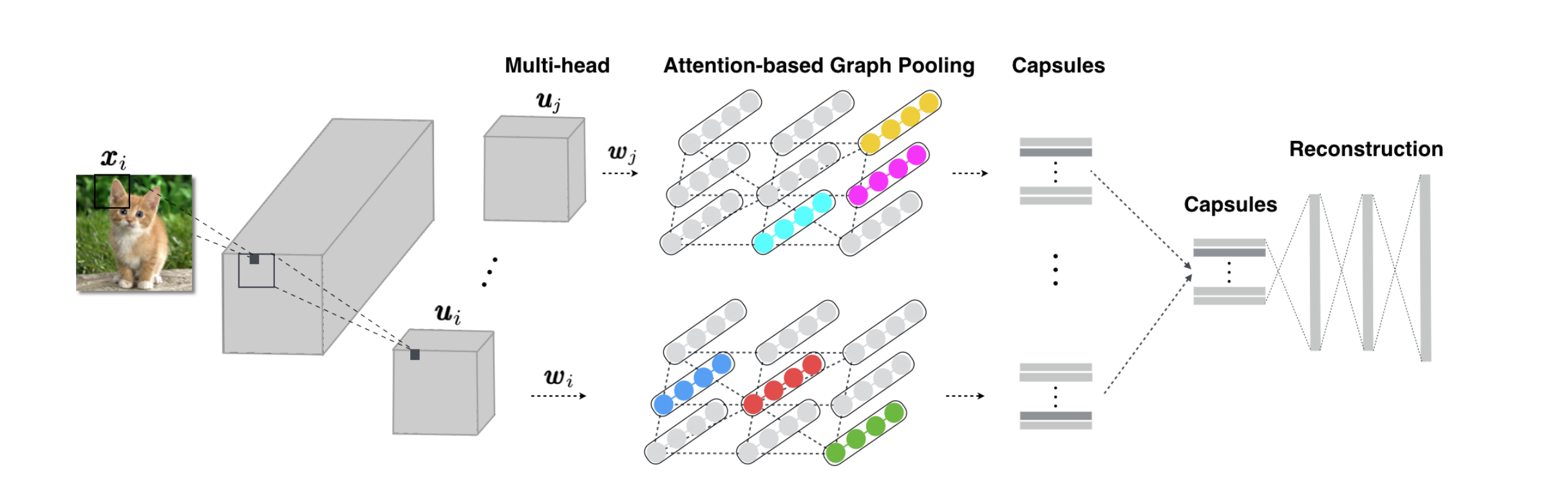}
        \caption{GraCapsNets architecture, showing how the extracted primary capsules are transformed and modeled as multiple graphs through attention-based graph pooling. Figure from ~\cite{gu2020interpretable}.}
        \label{fig:GraCapsNets}
\end{figure}

Aiming at making CapsNets more interpretable, in a manner  similar to Grad-CAM~\cite{selvaraju2017grad}  that has been proposed for explaining CNN-based classifications, the method termed \textit{GraCapsNets} proposed by Gu \& Volker~\cite{gu2020interpretable}, modifies CapsNets to have built-in explanations. As shown in Figure~\ref{fig:GraCapsNets}, the part-part relationship, i.e. the relationship between primary capsules, is modeled with graphs, followed by graph pooling operations that pool relevant object parts from the graphs to make a classification vote. The idea is that since the graph pooling operation reveals which input features are pooled as relevant ones one can create explanations to explain the classification decisions. In addition to interpretability, the proposed model improves object recognition via integrating graph modeling into CapsNets, hence treating capsules as node feature vectors and representing them as graphs so that one can leverage graph structure information.

\begin{figure}[t]
        \centering
        \includegraphics[trim={15 0 0 40},clip,width=\columnwidth]{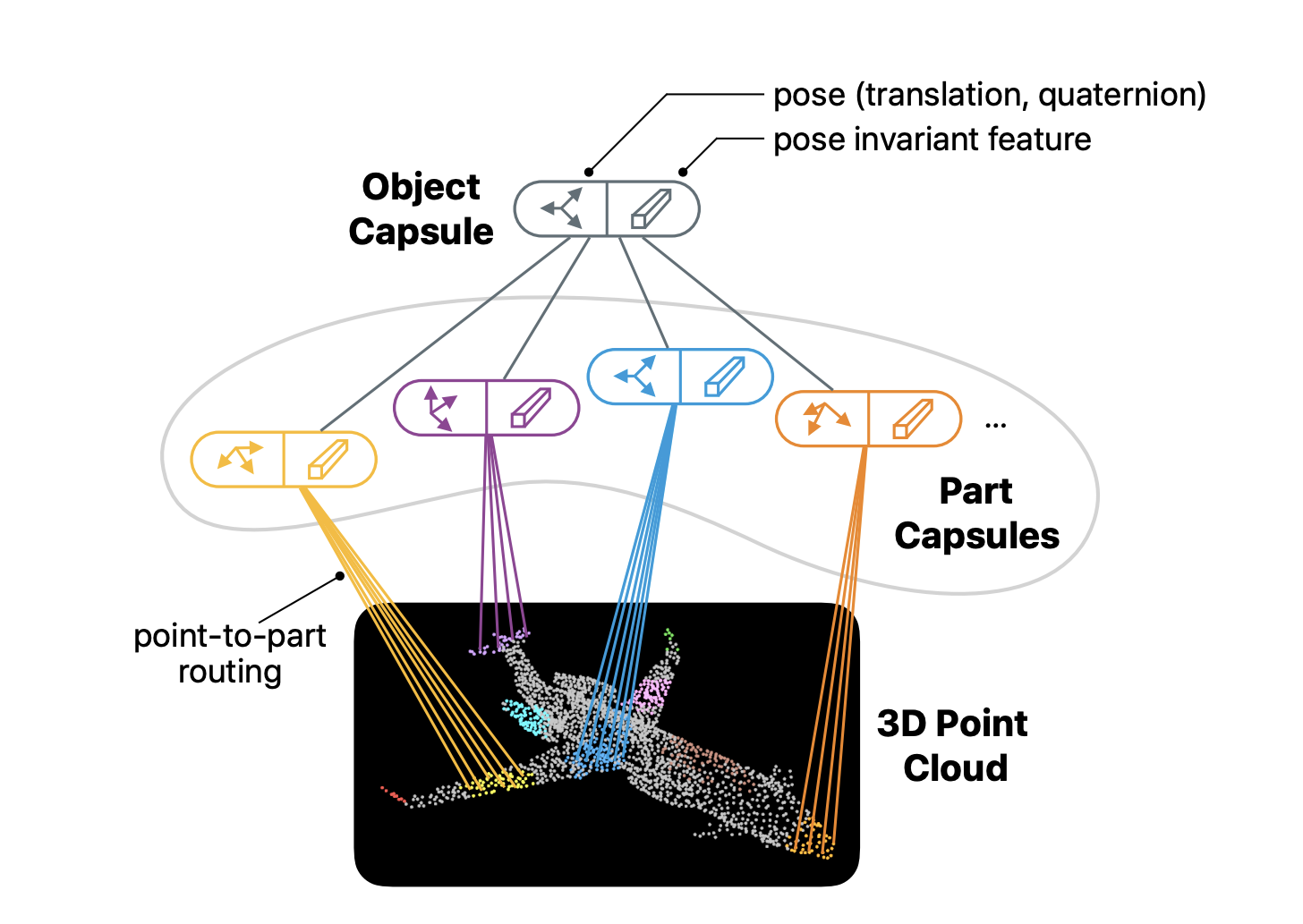}
        \caption{Geometric Capsule model overview showing insight into how different part Capsules represent different areas of the image along with details of the parts pose. These part capsules then activate the object Capsule to describe the entire object and its pose. Figure from ~\cite{srivastava2019geometric}. }
        \label{fig:geometriccapsule}
\end{figure}

Srivastava \etal ~\cite{srivastava2019geometric} proposed a geometric capsule design (Figure ~\ref{fig:geometriccapsule}), in which every visual entity -- part or whole object -- is encoded using two components: a pose and a feature. The \textit{pose} represents the transformation between a global frame and the entity's canonical frame in a geometrically interpretable manner, as a six-degree-of-freedom coordinate transformation. Conversely, the \textit{feature} is represented as a real-valued vector which encodes all non-pose attributes and is invariant to the object's pose w.r.t the viewer. The proposed \textit{Geometric Capsule Autoencoder} is constructed to group 3D points into parts and these parts into objects in an unsupervised manner. 

\begin{figure}[t]
        \centering
        \includegraphics[width=\columnwidth]{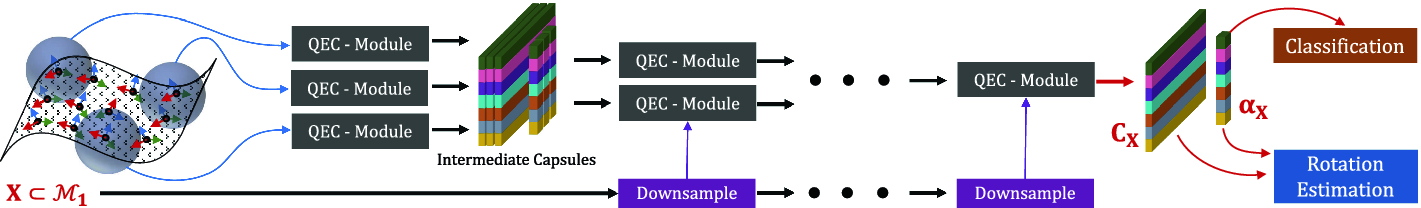}
        \caption{Entire capsule network architecutre based on a hierarchy of quaternion equivariant capsule modules. The input is a 3D point set $X$, where all the local patches are sent to the quaternion equivariant capsule (QEC) modules. At each level the points are pooled in order to increase the receptive field, gradually reducing the local reference frames into a single capsule per class. The classification and rotation estimation are used as supervision cues to train the transform kernels. Figure from ~\cite{zhao2020quaternion}.}
        \label{fig:quaternion}
\end{figure}

As highlighted by Zhao \etal ~\cite{zhao2020quaternion}, processing 3D point clouds is a challenging problem due to two main reasons: a) point clouds are irregular and unorganized, and b) the group of transformations that one has to deal with is more complex given that 3D data are often observed under arbitrary non-communicative $SO(3)$ rotations. Consequently, extracting and learning relevant embeddings requires 3D point networks to be equivariant to these transformations, while maintaining invariance properties to point permutations. The quaternion equivariant capsule module presented in~\cite{zhao2020quaternion} (Figure~\ref{fig:quaternion}) extends the work presented in~\cite{lenssen2018group} and is able to process point clouds while maintaining equivariance to $SO(3)$ rotations and preserving translation and permutation invariance. This work achieves $SO(3)$ by restricting the model to a sparse set of local reference frames (LRFs) that collectively determine the object orientation. In addition, the authors proposed a variation of dynamic routing, termed \textit{Weiszfeld dynamic routing} that uses inlier scores as activations, and which together with LRFs, form part of the quaternion equivariant capsule module. This process allows for equivariant latent representations to be extracted that point to local orientations and activations, while also disentangling orientation from evidence of object existence.

\begin{figure}[t]
        \centering
        \includegraphics[width=\columnwidth]{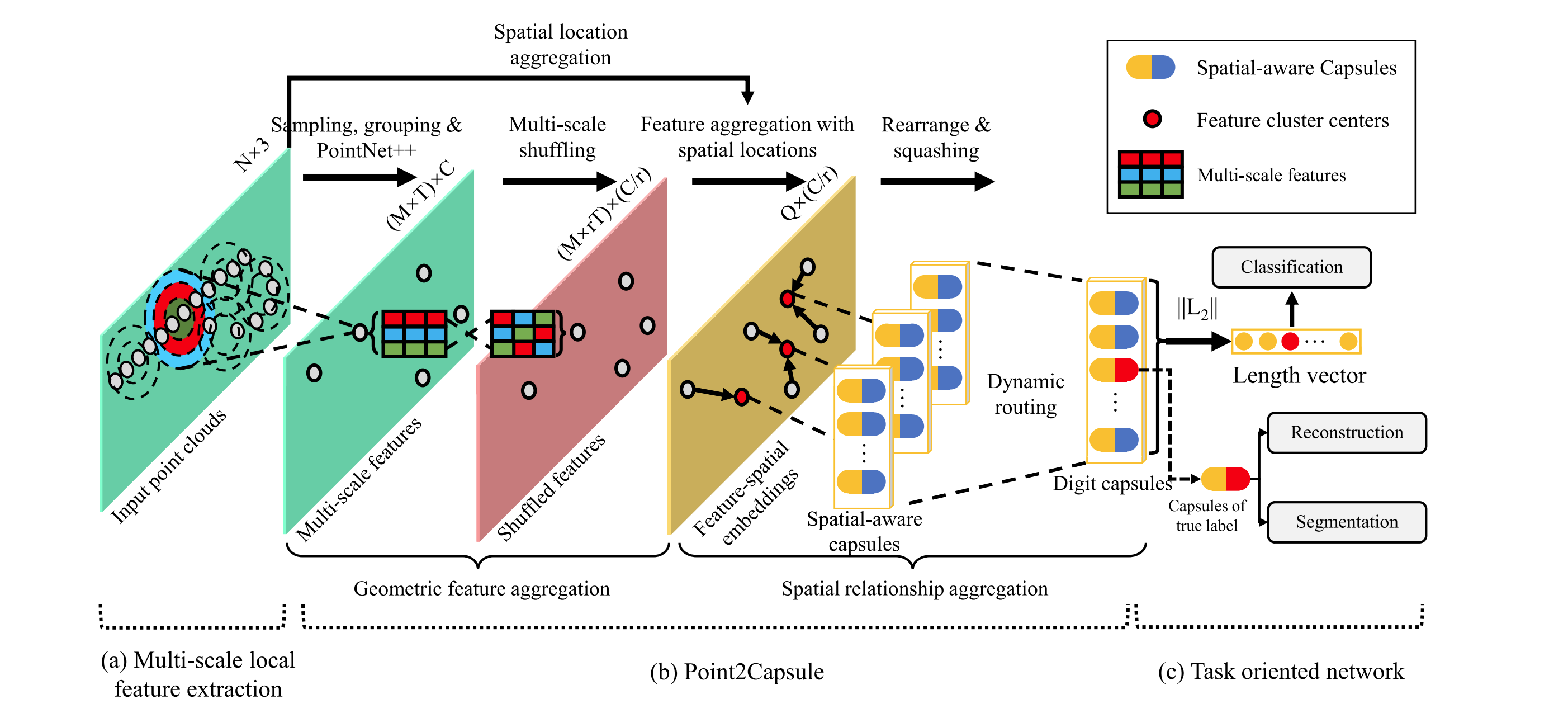}
        \caption{ Point2SpatialCapsule architecture. Assuming point clouds as input, the three steps involved are as follows: a) multi-scale features are extracted from multi-scale areas; b) feature-spatial embeddings are created and are aggregated via a spatial relationship aggregation step that considers both the embeddings and their spatial relationship; c) task-specific network to perform the downstream task. Figure from ~\cite{wen2020point2spatialcapsule}. }
        \label{fig:point2spatial}
\end{figure}

Another challenge when dealing with 3D point clouds concern adequately capturing spatial relationships between local regions, e.g. the relative locations to other regions in order to learn discriminative shape representation. Pooling-based feature aggregation methods struggle to achieve this satisfactorily. A way to overcome is presented in \cite{wen2020point2spatialcapsule} whereby a new architecture called "Point2SpatialCapsule" is proposed that consists of two parts: a) a module named \textit{geometric feature aggregation} is designed to aggregate the local region features into learnable cluster centers, which manages to encode the spatial locations from the original 3D space, and b) a module named \textit{spatial relationship aggregation} is also proposed that further aggregates the clustered features and the spatial relationships among them in the feature space using a new capsule layer termed \textit{spatial-aware capsules}. The complete architecture can be seen in Figure~\ref{fig:point2spatial}.

\begin{figure}[t]
        \centering
        \includegraphics[trim={10 0 25 70},clip,width=\columnwidth]{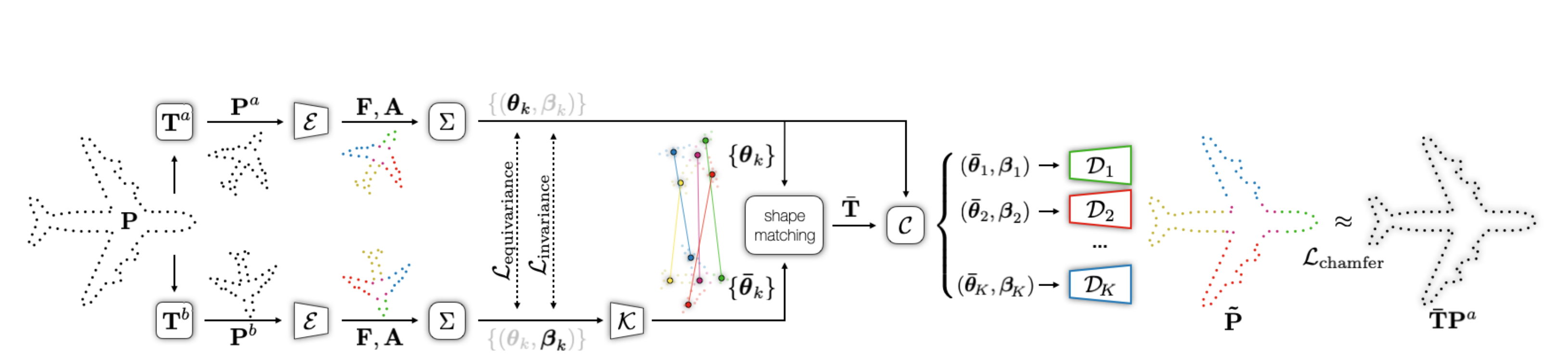}
        \caption{Canonical Capsules framework: the model learns a capsule encoder for 3D point clouds by relating the decomposition result of two random rigid transformations $T^a$ and $T^b$, of a given point cloud. Figure from ~\cite{sun2021canonical}. }
        \label{fig:canonical}
\end{figure}

Recently, a self-supervised capsule architecture for 3D point clouds was proposed by Sun \etal \cite{sun2021canonical} termed \textit{canonical capsules}. The algorithm computes K-part capsule decompositions of 3D point-cloud objects through permutation-equivariant attention while self-supervising the process by training with pairs of randomly rotated objects (i.e. siamese training). This process removes the need to pre-align training datasets. The decomposition of the point cloud takes place by assigning each point into one of the K parts via attention, which is then integrated into K keypoints. To ascertain equivariance, the two keypoint sets are set to differ only by the known  -- relative -- transformation; regarding invariance, this  takes place naturally by asking the descriptors of each keypoint of the two instances to match. The whole framework can be seen in detail in Figure ~\ref{fig:canonical}.

\begin{figure}[t]
        \centering
        \includegraphics[width=\columnwidth]{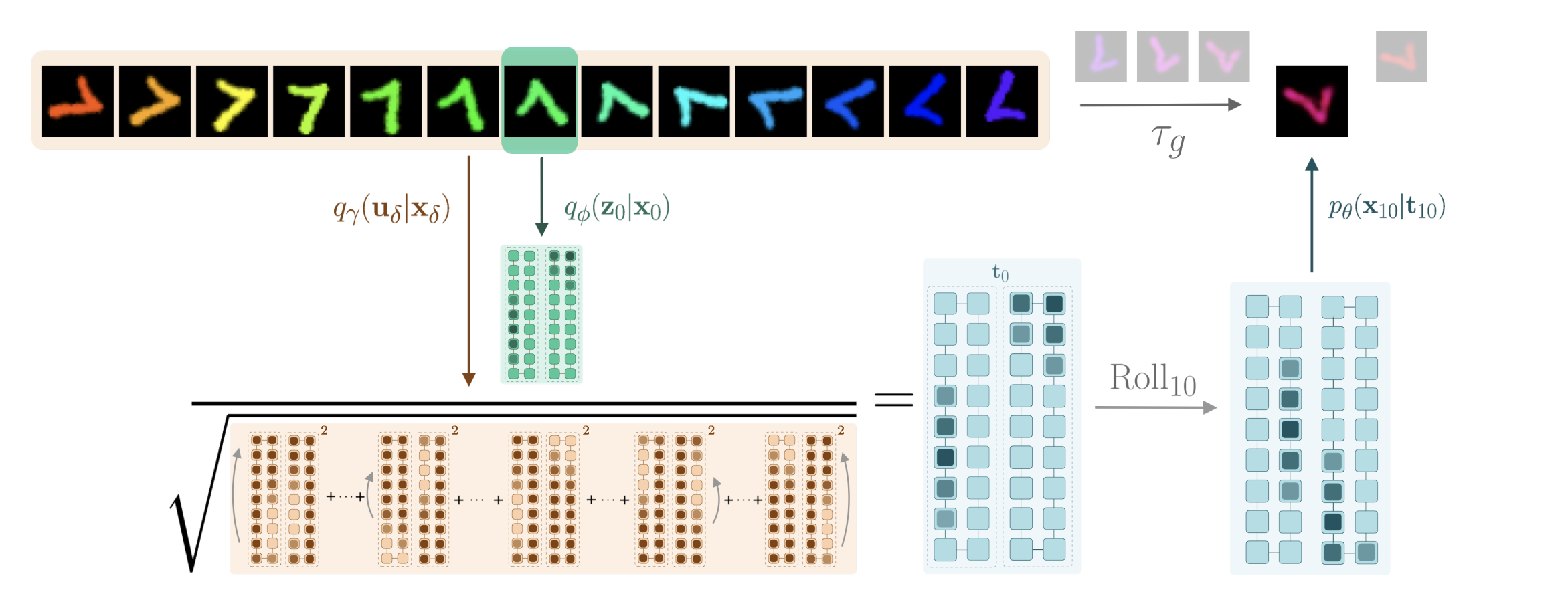}
        \caption{ Overview of the Topographic Variational Autoencoder with shifting temporal coherence. The combined color/rotation transformation in the input space $\tau_g$ becomes encoded as a Roll within the capsule dimension. Figure from~\cite{keller2021topographic}. }
        \label{fig:topographic}
\end{figure}

It is well established that developing better inductive biases can lead to better deep neural network architectures. Many parts of our brain are oranized topographically, such as the ocular dominance maps. Keller and Welling~\cite{keller2021topographic} built upon this concept and proposed the topographic variational autoencoder: a novel method for efficiently training deep generative models with topographically
organized latent variables (Figure ~\ref{fig:topographic}). In their paper they refer to capsules as "learning sets of approximately equivariant features or subspaces", and in fact the model they proposed tries to bridge two different classes of models, i.e., topographic generative models and equivariance neural networks. The proposed model is built upon the notion that inducing topographic organization can be leveraged to learn a basis of approximately equivariant capsules for observed transformation sequences. The resulting representation consists of a large set of "capsules" where the dimensions inside the capsule are topographically
structured, but between the capsules there is independence. The algorithm allows for sequences of input to be introduced to the model via encouraging topographic structure over time between sequentially permuted activations within a capsule, a property that the authors refer to as shifting temporal coherence. The mathematical background and further details can be found in ~\cite{keller2021topographic}.
\section{Generative Adversarial Capsule Networks}
Since the inception of CapsNets, there have been a few studies that proposed implementations and variations of Generative Adversarial Networks (GANs) with CapsNets. One of the earliest work was proposed by Jaiswal \etal~\cite{jaiswal2018capsulegan}, where a CapsNets was used as the discriminator and a deep CNN as the generator. However, the authors did not propose a new routing mechanism but instead the model used was the one proposed in~\cite{sabour2017dynamic}. 

\begin{figure}[t]
        \centering
        \includegraphics[trim={0 0 0 40},clip,width=\columnwidth]{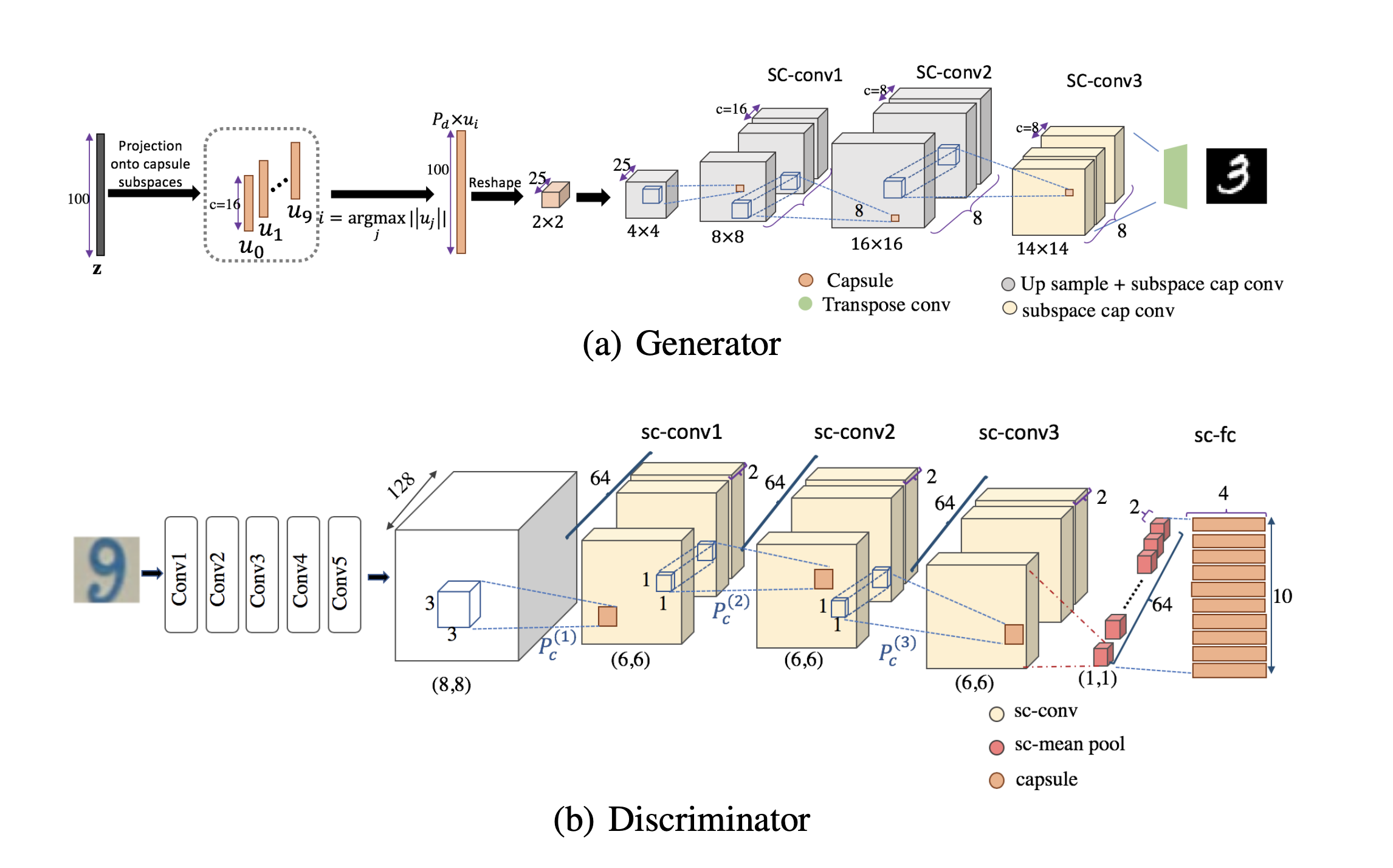}
        \caption{Overview of the Subspace Capsule Network (SCN) architecture. At the top (a), we can see the building blocks of SCN for the generative part of the GAN model, whereas at the bottom (b), a similar architecture is used for the discriminator component. Figure from~\cite{edraki2020subspace}. }
        \label{fig:subspace}
\end{figure}

On the other hand, in the work presented in~\cite{edraki2020subspace}, a new CapsNet model (Figure~\ref{fig:subspace}) was proposed, called Subspace Capsule Network (SCN), which are built upon the idea of modeling the properties of an entity through a group of capsule subspaces instead of simply grouping neurons to create capsules. Using a learnable transformation, a capsule is then created by projecting an
input feature vector from a lower layer onto the capsule subspace. This transformation finds the degree of alignment of the input with the properties modeled by the capsule subspace.

\begin{figure}[t!]
        \centering
        \includegraphics[width=\columnwidth]{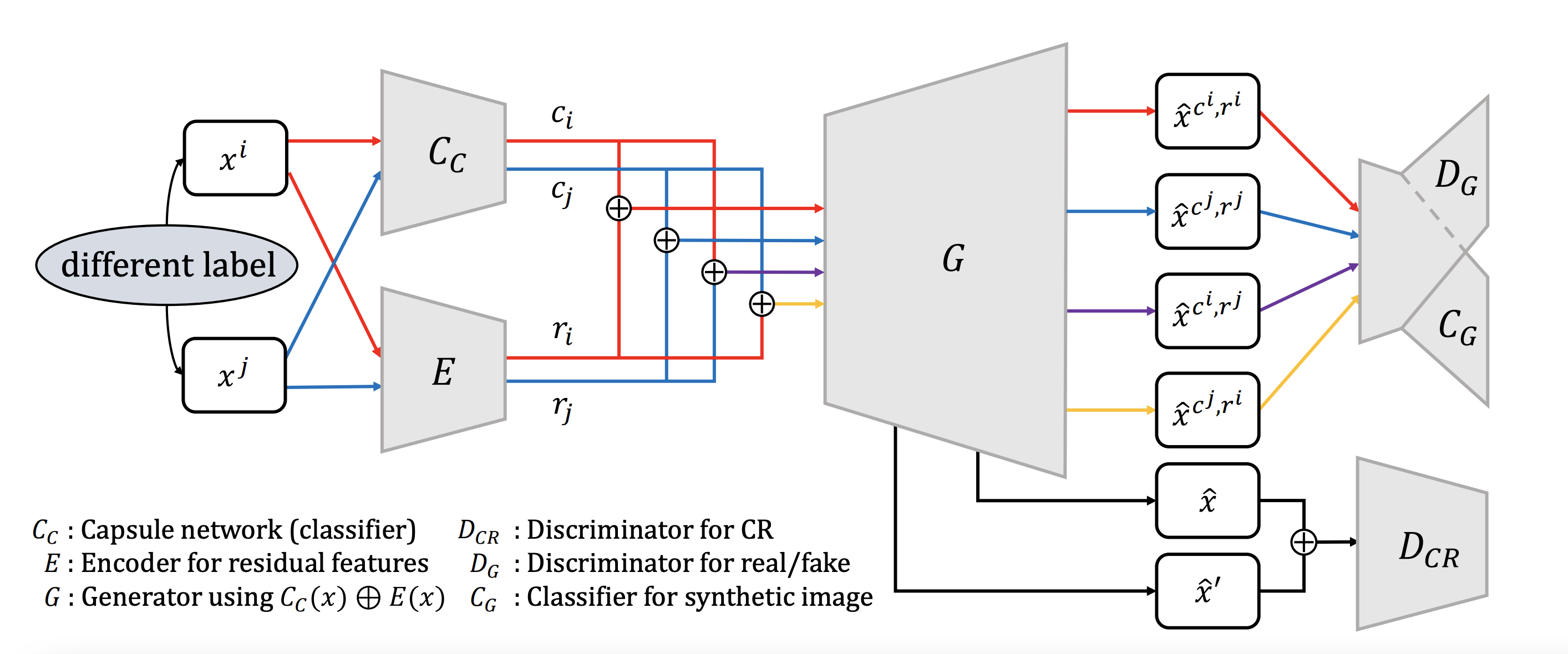}
        \caption{iCaps architecture, showing all the building blocks that correspond to the Capsule Network, the Encoder, the Generator, the Discriminator for contrastive regularization, the Discriminator for real/fake, and the classifier for synthetic image. Figure from ~\cite{jung2020icaps}.}
        \label{fig:icaps}
\end{figure}

An interpretable variation of CapsNets termed iCaps was proposed in ~\cite{jung2020icaps}, using class-supervised disentanglement learning. This approach aims at disetangling the latent
feature of $x$ into two complementary subspaces, i.e. class-relevant and class-irrelevant subspaces, in a setting where the class label for images in the training set is provided. The iCaps architecture, as can be seen in Figure ~\ref{fig:icaps}, consists of six different parts:
\begin{itemize}
    \item $C_c$: a capsule network (classifier) that represents the class-relevant latent space.
    \item $E$: an encoder that represents the class-irrelevant (residual) latent space.
    \item $G$: a generator that creates synthetic images using $C_c(x) \bigoplus E(x)$.
    \item $D_G$: a discriminator for image generation, that distinguishes whether an observation is from the dataset or from $G$.
    \item $C_G$: a classifier for image generation, that estimates class labels.
    \item $D_{CR}$: a discriminator for contrastive regularization that maximizes the distance between the concepts represented by $C_C$.
\end{itemize}

\begin{figure}[t!]
        \centering
        \includegraphics[trim={0 30 0 25},clip,width=.9\columnwidth]{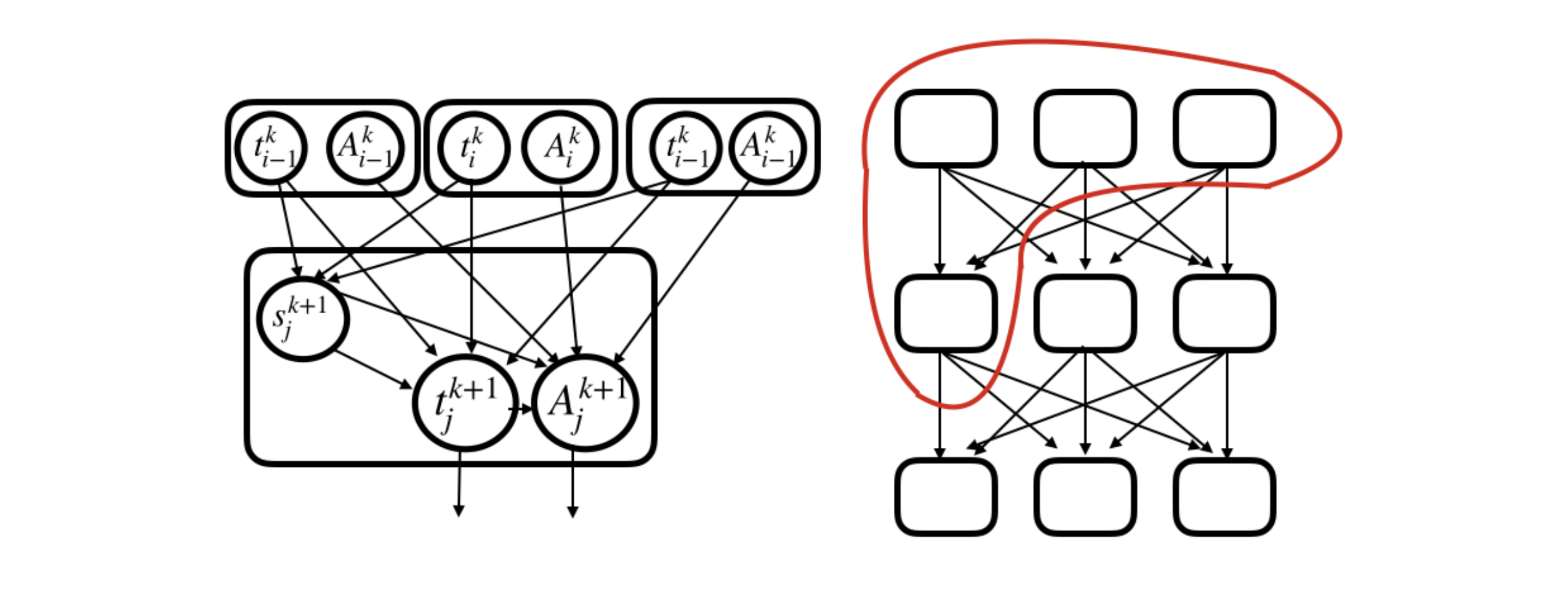}
        \caption{A diagrammatic overview of the generative model for a capsule network. On the left sketch one can see the detailed connectivity between the random variables that correspond to the red circled region of the overall graph that can be seen on the right. Figure from~\cite{smith2020capsule}. }
        \label{fig:probabilistic}
\end{figure}

A probabilistic generative version of capsule networks (Figure ~\ref{fig:probabilistic}) was proposed by Smith \etal in~\cite{smith2020capsule}, which aims to encode the assumptions under which capsules are built. This work is similar in spirit to the more recent work on inference in generative capsule models by~\cite{nazabal2021inference} as discussed previously in Section 4. Smith \etal introduced a variational bound which allowed them to explore the properties of their generative capsule model independently of the approximate inference scheme. In doing so, the authors gained insights into failures of the capsule assumptions and inference amortisation. Concretely, the authors expressed the modeling assumptions of capsules as a probabilistic model with joint distribution over all latent and observed random variables. They then derived a routing algorithm directly from variational inference principles, leading to an amortised method similar to variational autoencoders~\cite{kingma2013auto}. The approach they introduced for routing phrases the problem as approximate inference in a graphical model, hence allowing for further future improvements by leveraging advancements on inference in graphical models. Their model performs comparably with previous works on capsules, showing that their probabilistic interpretation is a close approximation to capsule network assumptions.

Their results also suggest that generative capsule formulations such as the one proposed may be helpful for enforcing desirable
equivariance properties, but that this is far from sufficient, as these models typically come without theoretical guarantees. They elaborate that, while promising, this type of formulation is still somewhat underdetermined. Specifically, there are issues relating to the identifiability of \textit{objects}, which suggests changes to the generative model may be necessary going forward.



\section{Capsules for Natural Language Processing}

Capsule networks have gained popularity in the field of natural language processing due to their ability to model part-whole relationships. Here, the sentence parts are individual words and the routing procedure learns the spatial and semantic relationships between the sentences components. Capsule-based architectures have been applied to a variety of natural language tasks including text classification, relation extraction, search personalization, and recommender systems.

\textbf{Text Classification.} Text classification spans several tasks including sentiment classification, question categorisation, news categorisation, and intent detection. Yang \etal \cite{yang2018investigating} first explored the use of capsule networks in natural language processing for the problem of text classification. Here, the dynamic routing algorithm \cite{sabour2017dynamic} is augmented to deal with noisy capsules in three ways: the addition of orphan (i.e. background) categories, the use of leaky-softmax instead of the standard softmax operation to obtain the routing coefficients, and the multiplication of routing coefficients by the probability of existence of child capsules (denoted coefficient amendment). This work shows that capsule network can achieve strong performance across 6 datasets when compared with standard neural network methods like CNNs and LSTMs. Kim \etal \cite{kim2020text} also propose a capsule network for text classification. To circumvent the need for max-pooling the text sequence, this work makes use of an ELU-gate unit that does not lose spatial information. Following the gate unit, the primary capsules are generated and passed through a ``static routing" procedure which consists of a single forward pass of the dynamic routing algorithm. Another work \cite{zhao2019towards} attempts to adapt capsule networks to be more successfully applied to NLP applications. First, a capsule compression operation is performed which merges similar capsules to reduce the number of primary capsules. Then, for routing an adaptive optimizer is introduced which allows for a variable number of routing iterations for a given sample. Lastly, for final classification a partial routing procedure allows for a reduced number of output capsules to be produced and leading to a large reduction on computational cost.

\begin{figure}[t]
        \centering
        \includegraphics[width=\columnwidth]{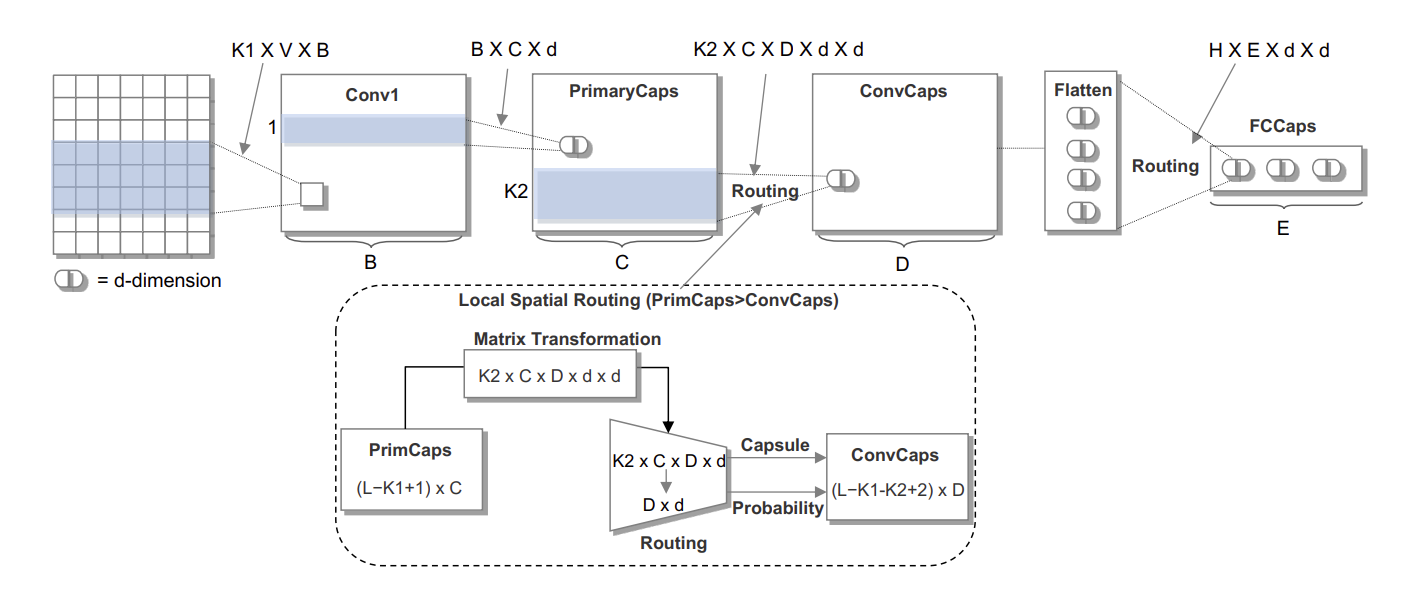}
        \caption{Architecture of a Capsule Network for text classification in \cite{yang2018investigating} The adjusted local spatial routing between the Primary Capsules and the Convolutional Capsules is shown in detail.}
        \label{fig:nlpcaps}
\end{figure}

Another work \cite{xia2018zero} utilizes capsule networks for the task of zero-shot user intent detection. The proposed capsule network can detect intents unseen at training time by modifying the dynamic routing algorithm with self-attention and allowing for the generation of capsules for emerging intents (\ie intents not used during training). Furthermore, analysis of the routing coefficients illustrate the capsule networks' ability to model the relationships between parts (words) and their corresponding whole (intents). Chen \etal \cite{chen2019transfer} propose a Transfer Capsule Network (TransCap) for the problem of aspect-level sentiment classification (\ie  classifying the sentiment of a specific aspect occurring in a sentence). Given a sentence and the given aspect, TransCap generates a set of feature capsules from the input words and then performs ``aspect routing" which gates the sentence (context) capsules using the aspect features to generate semantic capsules (SemanCaps). These SemanCaps are then passed through the dynamic routing algorithm to obtain the final classification capsule layer. Capsules have also been applied to the task of slot filling and intent detection \cite{zhang2018joint}. For a given sentence, this task involves a two-step classification problem: first assigning words to a specific slot class (\eg \textit{artist}, \textit{playlist}, \textit{movie\_type}) and then classifying the intent of the overall query (\eg \textit{change\_playlist}, \textit{play\_music}). Zhang \etal \cite{zhang2018joint} propose a capsule network which generate a capsule for each word in the sentence and applies dynamic routing to create a set of slot capsules. The routing coefficients represent the assignment of each word in the sentence to a slot class. Then, another routing operation is used to obtain the final intent capsules and the final intent classification.

\textbf{Relation Extraction.} Relation extraction is a problem involving finding the relationships between different entities (\ie words) within a sentence. Zhang \etal \cite{zhang2018attention} illustrate capsule networks' ability to learn these relationships. The proposed capsule network makes use of a bi-directional LSTM to generate the initial capsule layer. Then, dynamic routing with coefficient amendment is used to generate a set of parent capsules, whose activations are the probabilities of different relations. The entities by which these relations are represented are determined through another pretrained method \cite{han2018openke}. Another work \cite{zhang2019multi} proposes the architecture Att-CapNet which also uses a bi-LSTM to generate the initial capsule layer. Here, however, the hidden states of the recurrent model are also used in the generation of attention coefficients for improved capsule routing. 

\textbf{Search Personalization and Recommendation Systems.} Capsule networks have been applied to knowledge graph completion for the task of user search personalization \cite{vu2019capsule}. For this problem, the goal is to generate a rating for a tuple (\textit{query}, \textit{user}, \textit{document}), such that the output document is personalized for a given user's query. Vu \etal use a capsule-based approach which first encodes all elements of the tuple into vectors, and use convolution operations to generate a set of primary capsules. Then, the dynamic routing algorithm is used to generate a single two-dimensional capsule whose magnitude determines the score for the given tuple. A higher score denotes that there was agreement between the input capsules, leading to a higher ranking for the document when given the user's query. Capsules have also been used in recommendation systems. Given a set of reviews generated by a user and reviews for various items, recommendation systems attempt to suggest which items the user would like the most. Li \etal \cite{li2019capsule} present a capsule-based architecture, which takes embeddings for user and item reviews, and generates a set of positive and negative capsules to represent the sentiment of various aspects of the reviews. Then, a routing by bi-agreement algorithm is proposed which attempts to find agreement not only between different capsules, but also within dimensions of the same capsules (both inter-capsule and intra-capsule agreement). From the output of the routing alogorithm, a recommandation score (\ie rating) is generated for the given item.

\textbf{Language and Vision.} Several applications require the use of both textual and visual data. Capsules networks have shown promise in these multi-modal tasks. One such problem is that of visual question answering where a natural language question is given for an image, and the goal is to select a multiple choice answer or generate natural language answer. One of the first capsule-based approaches for this task is \cite{zhou2019dynamic}. Here, routing-by-agreement is used as an attention mechanism between the visual and textual features to improved learned representations.  Urooj \etal \cite{urooj2021found} propose a capsule framework for visual question answering grounding systems. This work shows that including capsules with EM-routing \cite{hinton2018matrix} in the generation of visual features leads to drastic improvement in grounding accuracy; the capsule architecture uses relevant visual information in intermediate reasoning steps. Recently, Cao \etal \cite{cao2021linguistically} present a routing algorithm that adjusts the capsule routing weights based on the parse tree generated from the given question. The proposed linguistically routed capsule network is shown to achieve strong visual questioning answering performance, even on out-of-distribution data. Capsule networks have also been applied to multimodal machine translation \cite{lin2020dynamic}. The goal of this problem is to improve the translation a natural language sentence with visual features. Lin \etal \cite{lin2020dynamic} propose a context-guided dynamic routing procedure to update the routing coefficients between visual capsules using cross-modal correlations. They show that this context-guided routing outperforms the standard attention and dynamic routing mechanisms when applied to their approach.

\section{Capsules for Medical Image Analysis}

Another field that Capsules have been applied fairly extensively to is that of medical imaging \cite{afshar2019capsule, zhang2019cervical, wang2020multikernel, gaddipati2019glaucoma, liu2019wbcaps}. This is largely in part due to a Capsule Networks ability to generalise to new variations of the learnt classes in unseen data which were not captured in the training data. Additionally this property is achieved after being trained on the small amount of data which is typical for medical datasets due to the expertise required in labelling.

SegCaps \cite{lalonde2018capsules} replaces the convolutional blocks of a U-Net \cite{ronneberger2015unet} with convolutional capsule blocks and modifies dynamic routing to be locally connected by only routing capsules in layer L to parent capsules in layer L+1 within a $k_h \times k_w$ kernel to semantically segment 2D slices of computed tomography (CT) scans showing irregular lesions and nodules. The modified Dynamic Routing alogrithm is shown below. This method achieved state of the art Dice Coefficient on the LUNA16 dataset with 98.479\% accuracy while also reducing the parameters by 95.4\%. 
Again, applied to detection of nodules in CT scans, Fast CapsNet \cite{mobiny2018fast} modified dynamic routing CapsNet \cite{sabour2017dynamic} in order to scale to 3D data. Their contributions are to add a constraint which allows only one Capsule per pixel location in the Primary Capsules. By doing this they reduce the computational overhead of routing by agreement by 32x resulting in a 3x overall increase of network training per epoch while retaining approximately the same accuracy as base CapsNets on 2D images which were slices of a 3D volume of the full CT scan. While this change may be small for 2D data, when they attempted to apply base CapsNets to 3D scans the network was unable to be trained stably whilst their modified CapsNet was able to achieve better accuracy on 3D than 2D data. Additionally they replace the fully connected layers in the decoder section of the CapsNet with deconvolutional layers. Overall they achieve 91.84\% accuracy which is greater than the 91.05\% accuracy provided by the best CNN based architectures and greatly improved upon the 73.65\% provided by non deep learning approaches. 

In the realm of Brain Tumour classification, CNNs have been shown to fail to utilise the spatial relationships between brain perturbations which causes misdiagnosis of tumours. As a result of this, the authors in \cite{afshar2019capsule} propose using a locally routed Capsule Network similar to \cite{mobiny2018fast}. Additionally, rough boundaries of where the tumours are located are given to the network to ensure that it ignores irrelevant areas of the image. When trained with the entire image without rough tumour boundaries, the network achieves 78\% accuracy on their dataset of MRI scans. This is then improved to 90.89\% when additionally given the rough tumour boundaries, slightly improving upon the state of the art CNNs which are able to achieve 88.33\% accuracy when given the rough tumour boundaries.

Building upon their previous work in \cite{lalonde2018capsules}, \cite{lalonde2021capsules} extends their work to five new datasets as well as introducing the condition that child capsules are only routed within a spatially located window. This change allows their network to scale to images of 512x512 size while remaining at around 1.5 million parameters. For reference a standard dynamically routed capsule network would require approximately 2 quadrillion parameters to scale to images of this size. Again, their method is able to outperform state of the art networks in five datasets DICE coefficient and outperforms in four of the five datasets in Hausdorff Distance score. Additionally it should be noted that their method has a significantly lower standard deviation between runs on different random seeds.

\begin{figure}[t]
    \centering
    \includegraphics[width=1\columnwidth]{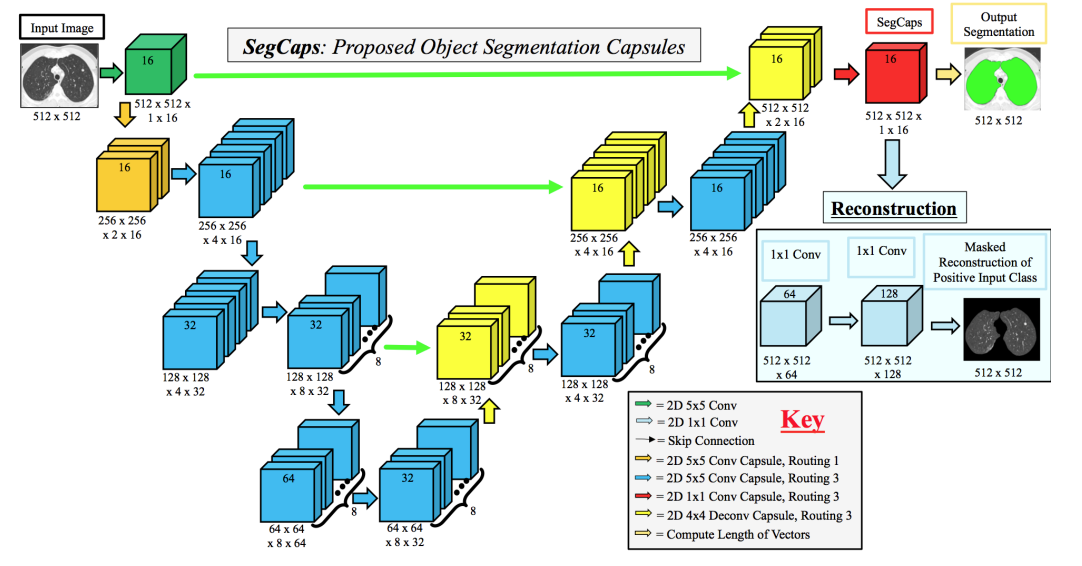}
    \caption{The architecture for SegCaps \cite{lalonde2018capsules} which uses Capsule Networks in a U-Net like architecture combining the information learnt at different levels of downscaling via Capsule Network convolutional layers. 
    }
    \label{fig:covid-caps}
\end{figure}

Covid-19 created a demand for fast and accurate diagnosis of patients. As a result multiple deep learning techniques were tested upon CT scan and x-rays of lungs. Motivated by Capsules Networks ability to achieve strong results with low amounts of data, COVID-CAPS \cite{AFSHAR2020638} was devised. Utilising a standard dynamic routing \cite{sabour2017dynamic} architecture to classify either positive or negative for Covid-19, COVID-CAPS is able to achieve an accuracy of 95.7\%. When pre-trained using a large dataset of other x-ray images, COVID-CAPS is able to achieve 98.3\% accuracy.

Similarly, \cite{li2021mha} propose MHA-CoroCapsule a novel capsule architecture which builds upon the dynamic routing capsules \cite{sabour2017dynamic} by replacing the dynamic routing element with a non iterative multi-head attention routing process. By using a dataset of under 300 lung x-rays of Covid-19 patients they were able to achieve nearly state of the art results of 97.28\% accuracy compared to 98.30\% achieved by COVID-CAPS \cite{AFSHAR2020638}, however COVID-CAPS is pre-trained on over 100 thousand images of other lung diseases compared to the 1 thousand total images used in MHA-CoroCapsules.

\section{Capsules in Other Applications}

Capsule Networks have been applied to many different fields. The majority of these applications only use the original Dynamically Routed Capsules in domains where either data is difficult to obtain or label or domains where a network needs to be able to generalise to different conditions than the training data.

\subsection{Fault Diagnosis}

In the field of bearing fault diagnosis it is not possible to easily obtain images to visually inspect whether certain elements of machinery are failing during operation. In \cite{zhu2019convolutional} the authors convert 1D signals of vibration and electrical current to time frequency graphs via Fourier transforms which can then be fed into  for analysis. Additionally these signals often contain a lot of noise and variations between every machine depending on the state of the sensors and the speed and load that the machine is operating at. The only modification which the authors make to standard dynamically routed capsules \cite{sabour2017dynamic} is the replacement of the initial convolutional layers with inception blocks from GoogLeNet \cite{szegedy2015going} to create Inception Capsule Networks (ICN). Over 6 different fault diagnosis tasks, ICNs are consistently either state of the art or comparative to non capsule methods, this is in stark contrast to the other methods which generally perform very well in one task but underperform in others. Overall, ICNs are able to achieve 97.15\% accuracy on average across the 6 tasks, compared to 94.58\% of the best competing method. Additionally, ICNs achieve 82.05\% when tested on data for 6 new tasks from machines operating at different loads and speeds. Additionally, the authors propose that the output of the class capsules can be used for regression, achieving 94.04\% accuracy in determining the severity of the faults detected.

Building upon the work in \cite{zhu2019convolutional}, the authors of \cite{wang2019novel} propose an extension which proposes the Xception module Capsule Network (XCN). This method follows the previous methodology of converting 1D signals to time frequency graphs, but uses wavelet time frequency analysis rather than Fourier transforms. XCNs are able to achieve 98.4\% accuracy across the three tasks of inner ring faults, outer ring faults and ball faults. This is a fairly large improvement compared to the 97.6\% that the best ensemble method is able to achieve, additionally XCNs perform more consistently well with 99.2\%, 99.7\% and 96.3\% compared to 96.4\%, 100\% and 96.4\% from the ensemble approach. Additionally, when compared to live data from a working machine, XCNs achieve 97.2\%, 98.7\% and 94.5\% accuracy, a significantly smaller drop off than all over approaches, showing clearly the strength of Capsule Networks ability to generalise.

\subsection{Hyperspectral Images}

Another application of Capsule Networks is in hyperspectral Image classification. Hyperspectral images are traditionally badly classified by current deep learning approaches due to their lack of ability to exploit the spatial relationships in the spectral spacial domain which is a key factor in dealing with the extremely high dimensional data. Additionally CNNs are known to require a large amount of data, which given the high dimensionality and complexity of hyperspectral imaging is not possible. In \cite{paoletti2018capsule} the authors propose using capsule networks named Spectral-Spatial Capsule Networks (SSCN) to classify these images. The authors achieve a large amount of success in this field with an unmodified dynamically routed Capsule Network. Over 5 different random seeds and trained on only 15\% of the available data, the author achieve state of the art segmentation in every class of two different datasets with 99.45\% and 99.95\% accuracy. In the third dataset SSCNs are able to achieve state of the art in 56 out of 58 classes with an average accuracy of 98.25\% on this most complex dataset. However it should be noted that the SSCN approach was significantly slower in terms of time per training epoch than all but one of the other approaches on all three of the datasets. Additionally in \cite{deng2018hyperspectral} the authors build upon the work in \cite{paoletti2018capsule} showing further how standard dynamically routed Capsule Networks are able to outperform CNNs achieving 96.27\% accuracy on a difficult hyperspectral image dataset compared to the state of the art CNN which achieves 95.63\%.

\subsection{Forgery Detection}

Detection of artificially generated forgery is a task where previously would require specific models for each variation of an attack. However in \cite{nguyen2019capsule} the authors propose a unified Capsule Powered framework named CAPSULE-FORENSICS (CF) where one network is able to detect forgeries of different types in both images and video. CFs use a pipeline of pre-processing where images are normalised and a video is split up into individual frames. These images are then fed through a VGG-19 \cite{simonyan2014very} network to extract the latent features rather than the traditional convolutional layers prior to the Primary Capsules. The vector output from the two class capsules of either real or fake is then used in a traditional forgery detection framework in order to detect forgeries. By applying noise to transformation matrix weights during the routing process, CFs are able to achieve state of the art or comparable results in 6 different datasets. Building upon the work in \cite{nguyen2019capsule}, the authors of \cite{luo2021capsule} improve the forgery detection framework to include the ability to detect AI generated audio forgeries, they achieve 98.93\% and 97.95\% accuracy on the PA and LA subsets of the ASVspoof2019 dataset while the previous best techniques are only able to achieve 98.16\% and 96.22\% accuracy.

\subsection{Adverserial Attacks}

The properties of Capsule Networks are desirable for a number of reasons, including how they are resistant to affine transformation and single pixel attacks which effect CNNs. In \cite{Qin2020Detecting} authors show that the reconstruction element of Capsule Networks improves the robustness of the network against standard CNN attacks.

In \cite{gu2021effective} is  shown that Dynamically Routed Capsule Networks \cite{sabour2017dynamic} are able to maintain a 17.3\% accuracy on the CIFAR-10 dataset \cite{krizhevsky2009learning} against PGD attacks \cite{madry2017towards} compared to 
CNNs with 0\% accuracy.   
Also, since the voting mechanism is so slow, it takes longer to iteratively generate adversarial images. However, they then refocus their attacks to specifically attack the voting mechanism by maximising Eq.~\eqref {equ:vote_attack} below. With this attack inside the PGD framework, the authors are able to decrease the performance of the network to 4.83\%.
\begin{equation}
\pmb{\delta}^* =   \argmax_{\pmb{\delta} \in \mathcal{N}_{\epsilon}}  \mathcal{H}\big[ \log g(\frac{1}{N} \sum_{i=1}^N f^i_v(\pmb{x}+\pmb{\delta})), \pmb{y}\big].
\label{equ:vote_attack}
\end{equation}
Using this adversarial attack, the authors proceed to train a Capsule Network which is specifically able to resist this kind of attack within the \cite{Qin2020Detecting} framework which creates a Capsule Network able to maintain 55.3\% accuracy of it's 94.7\% accuracy when tested against adversarial PGD attacks.
\section{Discussion \& Future Directions}
Despite the many successes of modern DL-based vision systems~\cite{krizhevsky2012imagenet,lecun2015deep,he2016deep}, a general lack of robustness to distributional shifts remains prevalent~\cite{greff2020binding}. Indeed, unlike current systems, humans are able to quickly adapt to distributional changes using very few examples to learn from~\cite{bruce1994recognizing,cohen2017,bengio2021deep}. There is compelling evidence that humans parse visual scenes into part-whole hierarchies, and that we do so by modelling the viewpoint-invariant spatial relationship between a part and a whole, as the coordinate transformation between
the intrinsic coordinate frames assigned to them~\cite{rock1973orientation,hinton1979some,kahneman1992reviewing}. One way to make Neural Networks (NN) more transparent and interpretable, is to try to make them understand images in the same way humans do. However, this is difficult for standard NNs because they cannot dynamically represent a different part-whole hierarchy tree structure for each image. This inability was one the main motivations behind capsule networks~\cite{hinton2021represent}. In this paper, we provided an extensive breakdown of the literature on object-centric learning using capsules and related attention-based methods. In doing so, we remark that Capsule networks do not yet work as well as they might, which can
in part be attributed to their lack of eﬃciency in enforcing the aforementioned
premises (see Section~\ref{subsec:foundations}). The additional complexity induced by vector valued neural activities, along with the high-dimensional coincidence ﬁltering algorithm to detect capsule level features (capsule routing), leads to very ineﬃcient models that are often difficult to train. In the following sections, we provide an in depth discussion on what we believe are the main conceptual considerations for future research in the field.

\subsection{The Hardware Lottery}
In their detailed analysis, \cite{barham2019machine} explained that although their convolutional capsule model required around 4 times fewer floating point operations (FLOPS) with 16 times fewer parameters than their CNN, implementations in both TensorFlow~\cite{abadi2016tensorflow} and PyTorch~\cite{paszke2019pytorch} ran significantly slower and ran out of memory with much smaller models. Although several more efficient versions of capsule routing have since then been proposed~\cite{self-routing,ahmed2019star,Tsai2020Capsules,ribeiro2020capsule}, the underlying problem is not only caused by routing but by the capsule voting procedure as well. In their analysis, \cite{barham2019machine} conclude that current frameworks have been highly optimised for a small subset of computations used by a popular family of models, and that these frameworks have become poorly suited to research since there is a huge discrepancy in performance between standard and non-standard compute workloads. As a result, non-standard workloads like those induced by the routing and voting procedures in capsule networks are a lot slower than they could be. As pointed out by~\cite{hooker2020hardware}, while capsule network's operations can be implemented reasonably well on CPUs, performance drops drastically on accelarators like GPUs and TPUs since they have been heavily optimized for standard workloads using the building blocks found in common architectures. This phenomenon begs the question of how much the tools researchers have readily available can predetermine the success of certain ideas based on how well they can be operationalized. 

To conclude, any conceptual changes to capsule networks which can capture their inherent inductive biases whilst improving their operationalization using current hardware/frameworks would constitute a significant breakthrough. The development of more flexible tools that enable research using non-standard workloads is also of paramount importance going forward, if we are to avoid future hardware lotteries.

\subsection{The Binding Problem}
\begin{figure}[t]
    \centering
    \includegraphics[trim={0 0 0 0}, clip,width=.95\columnwidth]{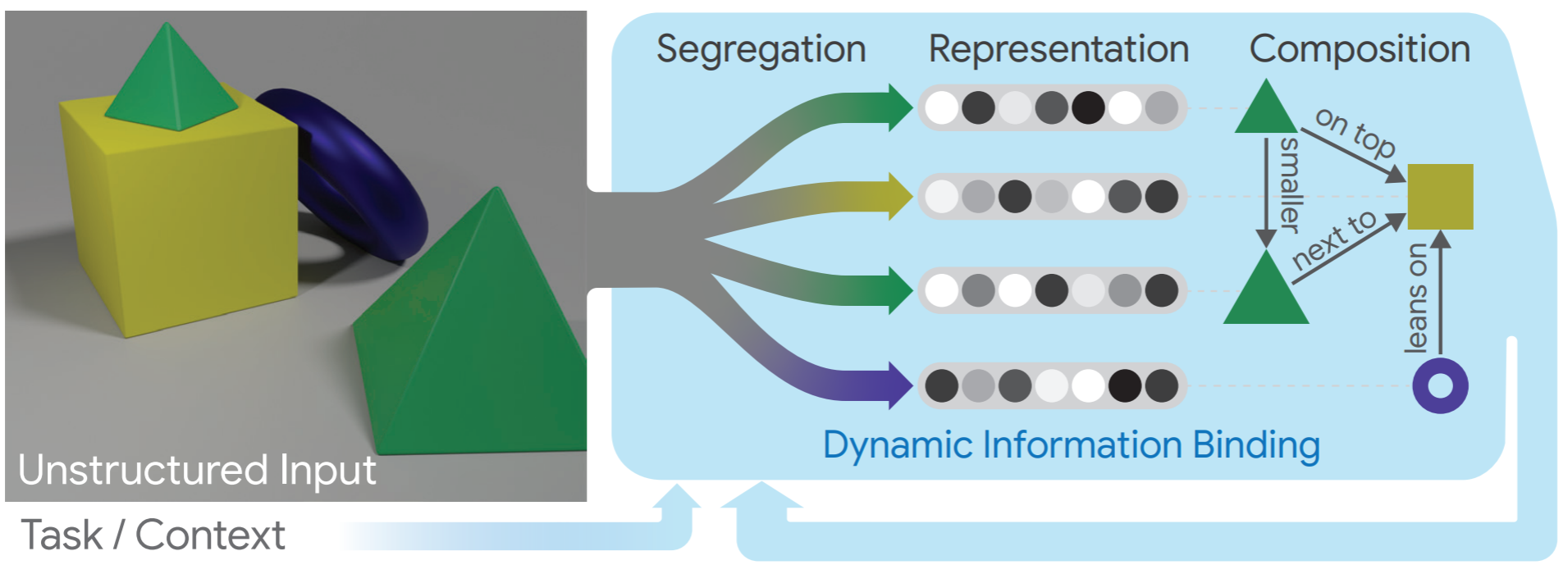}
    \caption{Depiction of the binding problem in neural networks. It can be understood as a collection of three subproblems, namely: \textit{segregation}, \textit{representation} and \textit{composition}. Figure from ~\cite{greff2020binding}.
    }
    \label{fig:binding}
\end{figure}
The motivation behind capsule networks is part of an overarching narrative around addressing the shortcomings of neural networks at human-level generalization. In most works, this problem is tackled from a computer vision perspective, where the goal is to be able to extract sensible object-centric representations from raw visual input in order to endow neural networks with reasoning and compositionality capabilities. The main assumption behind this is that \textit{objects} play a fundamental role in systematic generalization. Greff \etal~\cite{greff2020binding} argue that current limitations of neural networks are due to the \textit{binding problem}, which prevents them from incorporating sensible object-centric representations. As quoted in ~\cite{feldman2013neural} "In its most general form, The Binding Problem concerns how items that are encoded by distinct brain circuits can be combined for perception, decision, and action". Biological neural networks in human brains are said to overcome the binding problem by enabling flexible and dynamic binding of information belonging to separate entities~\cite{lowe2022complex}. On the other hand, even the most advanced DL systems today struggle with compositionality~\cite{nichol2021glide,ramesh2021zero}, so there is a need for neural network based systems that attempt to tackle the binding problem more explicitly like capsule networks. 

With that in mind, Greff \etal~\cite{greff2020binding} propose a functional division of the binding problem into three subproblems as depicted in Figure~\ref{fig:binding}. The \textit{segregation} problem refers to the ability to form modular object representations from raw inputs. The \textit{representation} problem relates to separately representing multiple object representations in a
common format, without interference between them. Lastly, the \textit{composition} problem involves the capacity to dynamically relate and compose object representations to build structured object-centric models for predictive tasks. Indeed, we believe that overcoming each of these open subproblems may give rise to more robust neural systems which can learn to dynamically represent structured models that generalize more like humans do. Capsule networks are one potential approach for tackling some aspects of the binding problem from a neurosymbolic perspective, but thus far they have mostly been applied in constrained supervised settings, and lack the capabilities to integrate segregation, representation and composition into a single system. Since most previous work on capsule networks have only considered the category slot format, there is an opportunity to extend them to different types of slot representations as shown in Figure~\ref{fig:slots}.

%
\subsection{Limitations \& Open Challenges}
It remains to be seen whether capsule networks will become the next big thing. They have the potential to be a disruptive technology, but until the framework and hardware limitations are overcome, it will be quite difficult to truly test them at scale. On that note, we believe that developing more flexible frameworks that are better suited to research is a really important avenue for future work. As stressed by~\cite{barham2019machine}, although the current machine learning tool chains are extremely powerful and useful to many, there are still concerns regarding the lack of flexibility of languages and backends putting a brake on innovative research in our fast developing field. 

From a technical perspective, we believe the main takeaways of capsule network models are the crucial ideas of high-dimensional coincidence filtering and agreement. In fact, many parallels can be drawn between capsules and the very successful attention mechanism in Transformers~\cite{vaswani2017attention}, as both methods measure high dimensional coincidences using the \textit{agreement} between neural activity vectors as a feature detection mechanism. This is generally a good idea since random vectors tend towards orthogonality as the number of dimensions increases. Embedding vectors in Transformers can also be seen as capsules with a much greater number of dimensions. It would be interesting to see if a Transformer-like visual model with the correct part-whole inductive wiring can deal with viewpoint-changes like capsule networks attempt to do, whilst being more efficient~\cite{kosiorek2019stacked}. There is also an opportunity to develop Vision Transformer (ViT)~\cite{dosovitskiy2020image} style capsule architectures which do away with the expensive convolutional capsules formulation in favour of patch based processing. This setup would likely entail using a fewer number of much higher dimensional capsules, and a potential relaxation of the inductive biases incurred by the computationally costly capsule voting procedure. 

The extraction of more faithful primary capsules is also a promising research direction, since capsule networks remain hindered by the inability to learn effective low level part descriptions (i.e. inverse rendering). Initial steps in this direction have recently been taken in~\cite{sabour2021unsupervised} where visual motion is used a a cue for part definition. One of the central ideas behind this is that a \textit{part} or an \textit{object} can be thought of as an entity that is perceptually consistent across time. There is an opportunity to extend these ideas to larger video datasets and using self-supervised learning to uncover 3D parts in much more complex scenes as done in~\cite{kipf2021conditional}. Improved primary capsule representations could also lead to more effective iterative refinement routing algorithms, since the latter are predicated upon adequate low level part descriptions to work as intended. 

To conclude, another aspect of interest for future work is carving out the role of approximately equivariant models like capsule networks in geometric deep learning~\cite{bronstein2017geometric,cohen2016group,keller2021topographic}. Specifically, further study and comparison of approximately equivariant models to their counterparts in terms of equivariance metrics, sample/runtime complexity, generative modelling and semi-supervised learning performance in 2D/3D complex tasks could be particularly important going forward. Moreover, given that capsule networks are known to be only approximately equivariant, there is an opportunity to develop/uncover formal equivariance guarantees which may encourage the usage of these types of models, and mitigate risks of over-reliance on approximate equivariance properties.

\ifCLASSOPTIONcompsoc
  \section*{Acknowledgments}
\else
  \section*{Acknowledgment}
\fi

The authors would like to thank all reviewers, and especially Professor Chris Williams from the School of Informatics of the University of Edinburgh, who provided constructive feedback and ideas on how to improve this work.

\ifCLASSOPTIONcaptionsoff
  \newpage
\fi



%



\bibliographystyle{IEEEtran}
\bibliography{bibfile.bib}

%








\end{document}